\documentclass{article} 
\usepackage{iclr2027_conference,times}
\iclrfinalcopy

\usepackage{amsmath,amsfonts,bm}

\def\eqref#1{equation~\ref{#1}}

\def\1{\bm{1}}

\DeclareMathAlphabet{\mathsfit}{\encodingdefault}{\sfdefault}{m}{sl}
\SetMathAlphabet{\mathsfit}{bold}{\encodingdefault}{\sfdefault}{bx}{n}

\usepackage{subcaption}
\usepackage{hyperref}
\usepackage{url}
\usepackage{booktabs}   
\usepackage{multirow}   
\usepackage{appendix}
\usepackage{titletoc}
\usepackage{graphicx}   
\usepackage[T1]{fontenc}
\usepackage{graphicx}
\usepackage{wrapfig}
\usepackage[most]{tcolorbox}
\usepackage{enumitem}
\usepackage[table]{xcolor}
\usepackage{makecell}
\usepackage{nicematrix}
\usepackage{amsmath}
\usepackage{amssymb}
\title{\alg: Search-Efficient Data Agent via Calibrated Memory}

\author{%
\parbox[t]{\dimexpr\textwidth-2\tabcolsep}{\centering
\textbf{Hanze Li$^{1}$}\thanks{Corresponding author: \texttt{hanzeli@vt.edu}.} \quad
\textbf{Athan Z. Li$^{2}$} \quad
\textbf{Feiyang Kang$^{1}$} \quad
\textbf{Jiachen T. Wang$^{3}$} \\[0.3em]
\textbf{Yuhan Chen$^{1}$} \quad
\textbf{Ruishan Liu$^{2}$} \quad
\textbf{Ruoxi Jia$^{1}$} \\[0.5em]
\normalfont
$^{1}$Virginia Tech \quad
$^{2}$University of Southern California \quad
$^{3}$Princeton University 
}}

\providecommand{\alg}{\textsc{CaliMem}}
\begin{document}

\maketitle

\begin{abstract}
Data curation is a first-order design choice in machine learning, yet tailoring training data to a downstream model and task remains costly in labor, expertise, and compute. Datasets are therefore often processed through fixed pipelines and reused across applications. We show that this assumption is fragile: across 55 dataset--model tasks spanning biological domains, the same curation recipe can help one model while hurting another. Effective curation must adapt to the downstream setting, motivating automated search over data recipes. We first build an agentic system that searches for curation policies at deployment, improving data utility via \textit{test-time curation (TTC)}.
Although achieving performance gains, its search is driven largely by recent feedback and remains local. Adding persistent memory helps the agent retain evidence across iterations and reveals a deeper problem: it forms claims that extend beyond the conditions in which they were observed, sometimes abandoning an entire strategy after a single unsuccessful trial.

To address this problem, we introduce \alg{}, a calibrated-memory agentic loop that requires the agent to predict the outcome of each proposed policy before executing it. Comparing this prediction with the observed result gives the agent a \textit{reality check} on the claims behind its proposal. It uses the prediction error to retain supported claims, revise mistaken ones, and specify the conditions under which a conclusion holds. This makes the agent's beliefs testable, reduces overgeneralization, and supports more effective long-horizon exploration. Across biological and LLM post-training tasks, \alg{} consistently outperforms fixed curation baselines and published methods, and exceeds full-data training by up to 19.4\%. 
Code is open-sourced~\footnote{\small{Code repository: \url{https://anonymous.4open.science/r/CaliMem-450E}}\vspace{-1em}}.
\end{abstract}


\textit{"However beautiful the strategy, you should occasionally look at the results" -- Winston Churchill}

\section{Introduction}
\label{sec:introduction}

How a training set is constructed determines what a model learns and how efficiently it learns it. Data curation has accordingly grown beyond cleaning into a broad design space: removing redundant or low-value examples, selecting data for targeted capabilities, and setting the mixture of sources~\citep{lee2022dedup,xie2023doremi,xia2024less,wu2024icons,chen2026olmix}. Tailoring these choices to a downstream model and task, however, demands expertise and repeated training runs. In practice, datasets are often processed through fixed pipelines and reused across applications. Such pipelines support consistency and reproducibility, but \emph{a curation choice that benefits one downstream setting need not benefit another}.

\begin{figure}[t]
  \centering
  \vspace{-0.5em}
  \includegraphics[width=0.75\linewidth]{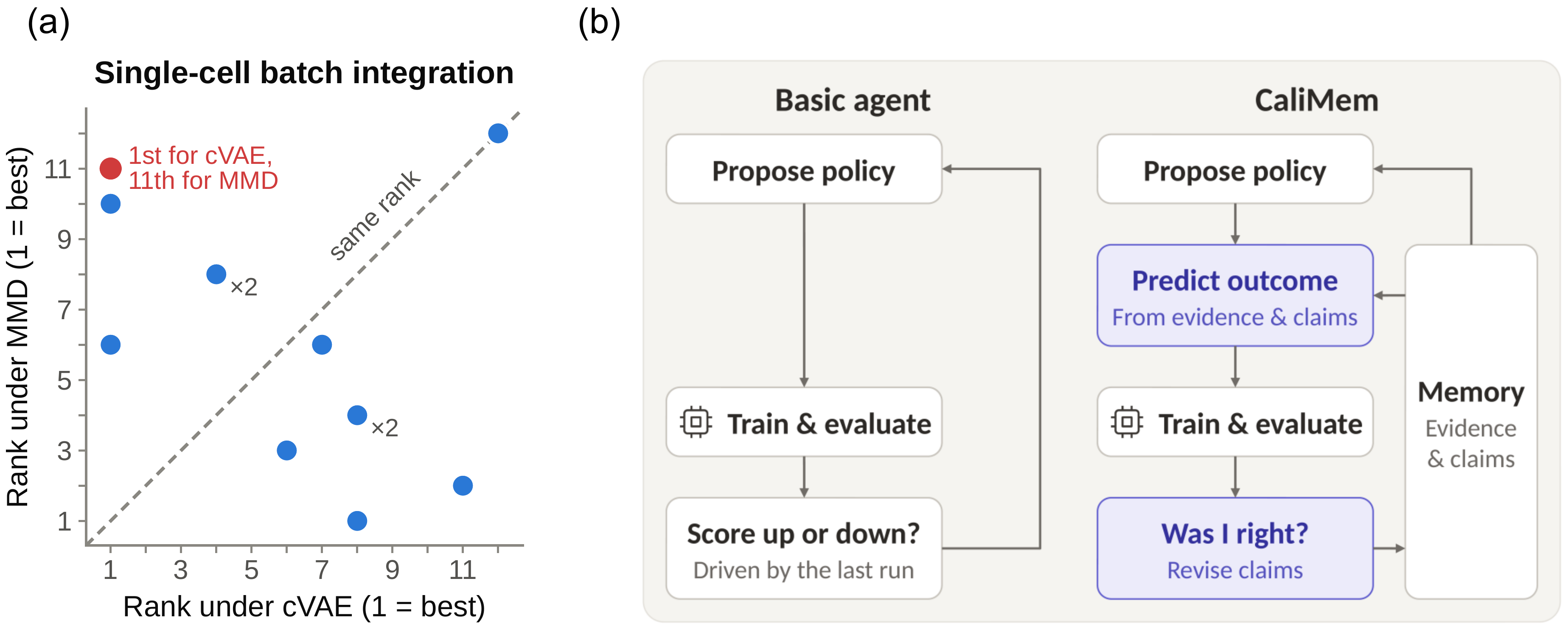}\vspace{-0.5em}
  \caption{\small{\textbf{Curation depends on the downstream model, and calibrated memory guides the search for a suitable policy.}
  \textbf{(a)} 12 standard recipes on a single-cell batch-integration dataset, ranked by the performance of two models trained on their half-size subsets. The recipe that ranks 1st for cVAE ranks 11th for MMD. 
  \textbf{(b)} A basic agent proposes a policy, trains and evaluates the model, and responds to the latest score change. \alg~records evidence and claims in memory, predicts the outcome of each proposed policy, and compares its prediction with the result. The comparison guides revisions to the claims used in subsequent iterations. }}
  
  \label{fig:intro}\vspace{-1.5em}
\end{figure}

This limitation is particularly evident in biological data, where data preparation commonly follows established workflows for single-cell transcriptomics~\citep{luecken2019practices,heumos2023practices,luecken2022scib}, protein variant-effect prediction~\citep{notin2023proteingym}, and non-coding variant-effect prediction~\citep{marin2024bend}. These workflows establish useful standards, but they do not establish which curation choices best serve a particular downstream application or model. We examine 55 dataset--model tasks across these 3 biological domains, comparing 12 standard curation recipes at a fixed data budget. The same recipe can help one model while hurting another. On one batch-integration dataset, for example, a recipe ranks 1st for cVAE (a variational autoencoder that conditions its decoder on the batch label) but 11th for MMD-AE (an autoencoder trained with a distribution discrepancy penalty between batches (Figure~\ref{fig:intro}a). Averaged across recipes, standard curation performs no better than selecting a random subset of the same size. \emph{Effective curation therefore calls for searching for a policy suited to the downstream setting.} The potential reach of such search extends beyond biology: data selection and mixing also shape language-model pretraining, instruction tuning, and vision-language training~\citep{xie2023doremi,xia2024less,wu2024icons}.

Generalist coding agents can carry out machine-learning experiments end to end~\citep{chan2024mlebench,karpathy2026autoresearch,rank2026posttrainbench}, making data curation a natural target for automated search. Following the propose--evaluate--revise setting of Curation-Bench~\citep{kang2026curationbench}, we build an agent that searches for curation policies at deployment, improving data utility via \emph{test-time curation (TTC)}.
Given a candidate data pool, a selection budget, and a fixed training and evaluation pipeline, it proposes a policy, trains on the selected subset, observes the evaluation score, and decides what to try next. The agent improves data utility, but its trajectories are strongly driven by recent feedback. It often changes one policy setting at a time and uses the latest score change to choose its next move (e.g., Table~\ref{tab:basic-trace}). This local search can miss interactions among choices and underuse evidence accumulated earlier in the trajectory. 

To help the agent learn from its full trajectory, we add persistent memory~\citep{shinn2023reflexion}. The memory records executed policies and their outcomes as evidence, alongside the agent's claims about what that evidence means. \emph{Retaining these claims makes a deeper failure mode visible: the agent frequently states conclusions substantially beyond what its experiments warrant.} For example, adding class rebalancing to a confidence-filtered policy lowers performance, and a memory-only agent concludes that rebalancing does not help (Figure~\ref{fig:memory}). The experiment establishes only that this particular combination failed under the tested conditions. It does not explicitly rule out rebalancing with a different filter, setting, or policy; though, the broad claim persists in the agent's belief
essentially 
prevents the agent from revisiting the strategy.

We introduce \alg{}, a calibrated-memory loop that subjects such claims to an empirical \emph{reality check}. Before executing each proposed policy, the agent predicts its outcome from the evidence and claims in memory. After evaluation, it compares the prediction with the observed score. This gap reveals when the agent's understanding of a policy was mistaken, even if the policy itself improved performance. The agent then retains supported claims, revises mistaken ones, or limits them to the conditions under which they hold. In the rebalancing example, an unexpected loss directs attention to its interaction with the confidence filter rather than ruling out rebalancing altogether. The resulting claims are more carefully qualified and provide a stronger basis for long-horizon exploration.

We evaluate \alg{} on the 55 biological dataset--model tasks and on LLM post-training tasks. It improves agentic search, outperforms fixed curation baselines and published methods, and, in the strongest setting, exceeds full-data training by up to 19.4\%. \noindent\textbf{Our contributions are threefold:} (i) we show that standard curation recipes have sharply different effects across downstream models and tasks; (ii) we identify local feedback-driven search and overgeneralized memory claims as limitations of agentic curation; and (iii) we develop \alg{}, which uses predictions and their errors to keep the agent's claims grounded in observed evidence.

\section{Related Work}
\label{sec:related}

\textbf{Data curation.} 
Filtering and deduplication~\citep{lee2022dedup},
target-aware sample selection~\citep{wang2025selecting,xia2024less,wu2024icons,ivison2025large},
and optimization of data-source mixtures~\citep{xie2023doremi,ye2025mixinglaws,
liu2025regmix,chen2025aioli,chen2026olmix} all improve training data; see
\citet{albalak2024survey} for a survey.
A recurring lesson is that there is no context-free ``best'' data: selection rankings change with pool, budget~\citep{ivison2025large}, and compute~\citep{muennighoff2023scaling,goyal2024scaling}. The same dependence appears in scientific workflows. Preprocessing choices in single-cell pipelines (e.g., normalization, feature selection) interact and shift downstream results~\citep{vieth2019pipelines,germain2020pipecomp,luecken2022scib}. This dependence motivates \emph{test-time curation (TTC)}: the agent searches the policy space against the downstream pipeline at deployment rather than committing to one recipe in advance.

\textbf{Agents for machine-learning research and data work.}
Large language model (LLM) agents increasingly carry out iterative machine-learning
experimentation, from bounded machine learning tasks to open-ended
research engineering~\citep{huang2025dca,
chan2024mlebench,wijk2025rebench,rank2026posttrainbench}.
Agents have also been applied to discovering dataset problems, data
standardization, web collection, and general data-science
workflows~\citep{huang2025dca,thirukovalluru2025indata, zhang2026datascibench}. Closest to us, \citet{kang2026curationbench} place a generalist agent in a fixed train--evaluate loop for data refinement; \citet{du2026datamaster} broaden autonomous data engineering
with tree-structured search and global memory. We take the curation loop itself as given and focus on a complementary question:
\emph{what does an agent learn about the response of one downstream pipeline over
a long search trajectory, and when should it trust what it has learned?}

\textbf{Memory, self-improvement, and predictive agent state.} Persistent-memory methods allow agents to accumulate reflections, experiences, and
reusable strategies without changing model weights~\citep{shinn2023reflexion,
zhao2024expel,suzgun2026dynamic,zhang2025ace}. Harness optimizers instead evolve prompts or workflows from execution feedback~\citep{khattab2024dspy,
yuksekgonul2024textgrad,zhang2025aflow}. 
Our scaffold is fixed; what evolves is the agent's beliefs about the target task. Predicting before acting connects to model-based agents~\citep{gu2025webdreamer,gkountouras2025language,richens2025general} and verbalized belief states ~\citep{singh2026agent}. Unlike confidence calibration~\citep{geng2024survey}, the calibrated object is the set of empirical claims used to choose the agent's next intervention. Appendix~\ref{sec:extended_related} extends this discussion.

%

\section{\alg{}: Calibrated Memory for Agentic Data Curation}\label{sec:calimem}

We build \alg{} in three steps, each motivated by a failure observed in the
previous one: a basic agent that implements the curation loop
(Section~\ref{sec:basic}), a persistent memory that lets it use its whole
trajectory (Section~\ref{sec:memory}), and a calibration step that checks the
claims in that memory against measured outcomes (Section~\ref{sec:calib}).

\subsection{Problem formulation: the data curation loop}
\label{sec:loop}

An automated data-curation agent operates through an iterative
propose--evaluate--revise loop~\citep{kang2026curationbench}. The agent is given
a candidate data pool $\mathcal{D}$, a selection budget $B$, and a fixed training
and evaluation pipeline. At iteration $t$, it proposes a \emph{data policy}
$\pi_t$, a program that selects a subset of exactly $B$ examples from
$\mathcal{D}$ subject to task-specific validity constraints, such as a minimum
number of examples per class or per batch (Appendix~\ref{app:domains}). The
downstream model is trained on this subset under the fixed pipeline and
measured on the evaluation split, which produces the score
$
y_t = S(\pi_t).
$ 
The agent observes $y_t$ and proposes the next policy $\pi_{t+1}$. After $t$
iterations, the trajectory
\begin{equation}
\mathcal{H}_t = \{(\pi_1,y_1),\ldots,(\pi_t,y_t)\}
\label{eq:traj}
\end{equation}
holds all the empirical experience available to guide subsequent search. We
write $\pi^\star_t$ for the \emph{incumbent}, the best policy in
$\mathcal{H}_t$, and $y^\star_t=\max_{i\le t}y_i$ for its score. In the default setup, the agent
builds on the incumbent: a change is kept if it improves on
$y^\star$ and reverted otherwise (as shown in Table~\ref{tab:basic-trace}). Every run starts from the same seeded random
subset as $\pi_1$ and is allotted $T$ iterations, each costing one full
training run; the quantity of interest is $y^\star_T$.

\vspace{-0.5em}\paragraph{Test-time curation.}
We refer to this search as \emph{test-time curation} (TTC): rather than preparing the
data once with a fixed recipe, compute is spent at deployment, by retraining
the fixed downstream model, to find the subset that serves that model best.
The trade is particularly favorable in scientific domains such as biology, where new data
come from costly experiments while many downstream models are relatively convenient to train
(Appendix~\ref{app:domains}). In this context, extracting more utility from an existing pool is
then far cheaper than enlarging it. Because the pipeline is fixed, a
difference in score between two iterations is attributable to the change in
the data policy.

\subsection{Basic agent: local, reactive search}
\label{sec:basic}

We first implement the loop with a basic agent. It receives the goal, the
budget, and the commands to submit a subset and to run training and
evaluation, together with a minimal protocol that asks it to read its previous
results before each move; which policies to try,
and how to revise them, is left to the agent. Applied to test-time curation in
both the biological and the LLM post-training domains, the basic agent already
improves on the random starting subset and, in the biological domains, matches
or exceeds the best human-designed recipe chosen separately for each subtask
(Basic agent rows of Tables~\ref{tab:main-bio} and~\ref{tab:main-llm}).

Its trajectories, however, are dominated by recent feedback. A curation policy
consists of a selection \emph{strategy}, such as allocating the budget across
perturbations in proportion to their size, and the \emph{parameters}
that implement it, such as thresholds, counts, and weights. Once the agent has
found a plausible strategy, it typically holds it fixed and spends its
iterations adjusting one setting at a time. 


\providecolor{baselineRow}{HTML}{F2F2F2}
\providecolor{discardRow}{HTML}{FBEAEA}
\providecolor{bestRow}{HTML}{E6F4EA}
\begin{table}[htbp]
\centering
\small
\caption{\small{
Excerpt from the trajectory of the basic agent on single-cell
perturbation-response prediction (Norman19, GEARS); the complete log of all 15
iterations is in Table~\ref{tab:basic-trace-full}. Scores are evaluation cosine
similarity between predicted and observed log fold changes ($\times$100). A
perturbation is the set of cells that received the same genetic perturbation,
and control cells are unperturbed. \emph{Keep} means the policy became the
incumbent; \emph{discard} means the agent reverted to the incumbent. Each row
lists what the agent changed and a paraphrase of the note it wrote after seeing
the score.}}
\label{tab:basic-trace}
\vspace{-0.5em}
\resizebox{0.9\linewidth}{!}{%
\renewcommand{\arraystretch}{1.15}
\begin{tabular}{>{\centering\arraybackslash}p{1.3cm}>{\centering\arraybackslash}p{1.0cm}>{\centering\arraybackslash}p{1.1cm}p{5.0cm}p{6.6cm}}
\toprule
\textbf{Iteration} & \textbf{Score} & \textbf{Status} & \textbf{What changed}
& \textbf{Agent's note} \\
\midrule
\rowcolor{baselineRow}
1 & 61.30 & keep & Random 50\% subset (baseline) & The score to beat. \\
4 & 63.63 & keep & Give each perturbation at least 100 cells (all of its cells
if it has fewer); split the rest by perturbation size
& New best: small perturbations now keep all their cells, and large ones keep
their share. \\
5 & 63.83 & keep & Control cells: $\sim$4{,}900 $\to$ 2{,}000
& New best: cutting controls frees cells for perturbations at little cost. \\
\rowcolor{bestRow}
6 & \textbf{67.53} & keep & Control cells: 2{,}000 $\to$ 1{,}000
& Large gain: more perturbation cells matter more than more controls. \\
\rowcolor{discardRow}
7 & 66.60 & discard & Control cells: 1{,}000 $\to$ 500
& Worse: 500 controls are too few, so 1{,}000 is the lower limit. Next, keep
1{,}000 controls and try a higher per-perturbation minimum. \\
\rowcolor{discardRow}
8 & 65.32 & discard & Control cells back to 1{,}000; per-perturbation minimum:
100 $\to$ 150
& Worse: a higher minimum takes cells from large perturbations, so 100 is the
right minimum. \\
\bottomrule
\end{tabular}%
}
\vspace{-1em}
\end{table}

\paragraph{Local refinement.} Table~\ref{tab:basic-trace} illustrates a sequence of local changes within a curation
strategy. The agent adjusts the number of control cells while
retaining the allocation rule for perturbed cells, then changes
the per-perturbation minimum after further reductions in controls
stop improving the score.

To describe this type of move, write a policy as
$\pi=\pi_{m,\theta}$, where $m$ specifies the strategy and
$\theta$ its numerical settings. A local proposal changes one
coordinate of the incumbent:
$
    \widetilde{\theta}_t
    =
    \theta^\star_{t-1}
    + \alpha_t s_t e_{j_t},
    \,\,
    \widetilde{\pi}_t
    =
    \pi_{m^\star_{t-1},\widetilde{\theta}_t},
$
where \(e_{j_t}\) is the basis vector for the selected \textit{coordinate}, $j_t$ is the selected setting, $s_t\in\{-1,+1\}$ its
direction, and $\alpha_t>0$ is a feasible step size. The resulting
score is compared with the incumbent to determine whether the
proposal is retained.
This resembles directional derivative-free direct search ("zeroth-order coordinate optimization"), which makes decisions
solely based on evaluations rather than an explicitly or implicitly modeled derivative (e.g., "gradient")~\citep{conn2009introduction, jordana2025introduction}. 

Local refinement is not itself a failure. The concern is that
the agent may inaccurately interpret a result obtained at one configuration
as a \emph{conclusion} about settings or strategy variants that have
not been tested. Such a conclusion can restrict later search
even when the underlying mechanism remains useful under
different conditions. (For example, after reducing the number of control cells from 1,000 to 500 hurts performance in Table~\ref{tab:basic-trace}, the agent concludes that 1,000 is a lower limit, even though the result only establishes this under the current perturbation-allocation rule.)
This has two consequences. First, the search refines the current strategy
rather than questioning it, and \emph{settles into a local optimum of that strategy.}
Second, a sign describes a move at one point of the policy
space, yet the agent records what it learns as a general conclusion;
yet whether an effect
depends on the other settings, or reverses outside the range tested, cannot be
read from a sign.

\vspace{-0.5em}\subsection{Holistic search via persistent memory}
\label{sec:memory}

The basic agent's exploration is driven by its most recent result and overlooks
patterns that its earlier iterations already reveal. Following work on
persistent memory for LLM agents~\citep{shinn2023reflexion,zhang2025ace},
we give the agent an external memory to explicitly record what it has
tried, what it has measured, what it has concluded, and which it consults
before choosing each new policy.

\vspace{-0.5em}\paragraph{Evidence and claims.}
The agent keeps a persistent memory $\mathcal{M}_t=(\mathcal{E}_t,\mathcal{C}_t)$,
updated after every iteration and read in full before each decision. The
\emph{evidence} $\mathcal{E}_t$ records facts: the policies executed so far with
their scores, $\{(\pi_i,y_i)\}_{i<t}$, and facts measured about the pool and the
pipeline, such as class counts or batch sizes.
The
\emph{claims} $\mathcal{C}_t$ record the conclusions the agent draws from this
evidence, such as which settings interact, where an effect reverses, and which
directions are spent, and are rewritten after each result,
$
\mathcal{C}_t\ \xrightarrow{\;(\pi_t,\,y_t)\;}\ \mathcal{C}_{t+1}.
$
The next policy is thus informed by the whole memory rather than only local feedback. 
Compared to the basic agent's log that records what was tried, the
claims record what the agent currently believes it means. For example, in one cell type annotation task, the agent revisited the number of input genes across non-adjacent iterations, reducing it from the default 2{,}000 to 1{,}684 at iteration 5 and later increasing it to 2{,}191 at iteration 10. Because both deviations lowered the score, the memory consolidated these observations into the claim that 2{,}000 genes was already the best (Appendix \ref{app:memfiles}). 
Writing the claims down also makes the agent's beliefs inspectable, and reading
them reveals a second problem. In memory, a conclusion
becomes a claim that is read before every later decision, so an overbroad claim
steers the rest of the run.

\begin{figure}[t]
  \centering
\includegraphics[width=0.85\linewidth]{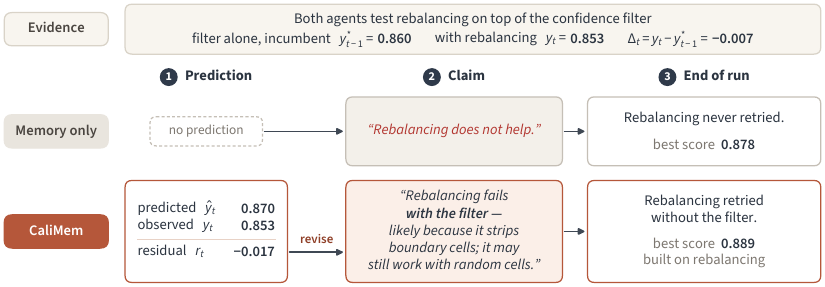}
  \caption{\small{\textbf{Without a prediction, one failure closes a whole direction;
with a prediction, the failure is tied to its condition.} Cell-type annotation
(\texttt{pancreas\_crosstech}, SVM; evaluation macro-F1). Both agents see
rebalancing fail on top of a confidence filter (0.860~$\to$~0.853). Top: the
failure becomes a claim about rebalancing as a whole. Bottom: the gap between
the predicted $+0.01$ and the observed $-0.007$ ties the failure to the
filter, and rebalancing stays open. The top row follows the memory-only
run; the bottom row adapts the memory + prediction run on the same task to
the same move, and its best score is from that run.}}\vspace{-1.5em}
  \label{fig:memory}
\end{figure}

Figure~\ref{fig:memory} (left) illustrates how an \textbf{overgeneralized conclusion} can restrict subsequent exploration: the memory accurately records the performance decrease, but the agent interprets this result as evidence that rebalancing is generally ineffective. This conclusion extends beyond the tested configuration. Once retained in memory, it guides subsequent decisions and prevents the agent from revisiting rebalancing, potentially excluding configurations that could improve performance.

\vspace{-0.5em}\subsection{Calibrated memory: predict, then calibrate}
\label{sec:calib}

To reduce overgeneralization, \alg{} requires the agent to predict each proposed policy's score from the evidence and claims in memory before evaluation. After evaluation, it compares the predicted and observed scores and uses the discrepancy to reassess the supporting claims. The intuition is that a performance change indicates whether a policy helped, while a prediction error indicates whether its effect matched the agent's expectations. The agent revises the claims accordingly and uses the updated memory to guide subsequent search.

\vspace{-0.5em}\paragraph{1. Predict before evaluation.}
Before executing policy $\pi_t$, the agent records a point
prediction $\widehat y_t$, the evidence and claims supporting
it, and an outcome that would challenge those claims. This
record is fixed before evaluation:
$
    \widehat y_t
    =
    \widehat S(\pi_t\mid E_t,C_t).
$

\vspace{-0.5em}\paragraph{2. Two diagnostics from one evaluation.} After observing $y_t$, the agent records \emph{(i) the signed prediction
residual (i.e., prediction error)}: 
$
    r_t = y_t-\widehat y_t.
$
A positive residual indicates that the outcome exceeded the
prediction; a negative residual indicates that the prediction
was too optimistic.  We also record \emph{(ii) the observed
changes} relative to the incumbent:
$
    \Delta_t
    =
    y_t-y^\star_{t-1}.
$
The diagnostics answer different questions.
$\Delta_t$ determines the keep-or-revert decision, subject to
the validity checks. The residual $r_t$ measures the discrepancy
between the expected and observed score. Their incumbent-relative
directions show whether the proposal crossed the improvement
boundary as expected. A policy can therefore become the new
incumbent while still producing a substantial prediction miss.

\vspace{-0.5em}\paragraph{3. Revising the claims.}
The agent uses the recorded prediction, outcome, diagnostics,
and supporting evidence to rewrite its claims in natural language. 
The interpretation of residual magnitude is left to the agent: Agreement with a prediction is evidence about the tested
configuration, not independent validation of every claim used
to justify it. Conversely, a discrepancy prompts reassessment
but does not by itself identify the cause: the agent must
consider the executed policy, the available observations, and
possible variability in the measured outcome. 

As a result, with \alg{}, the claims in memory carry the conditions under which they were observed, as in Figure~\ref{fig:memory} (right), and a failed variant narrows a mechanism instead of closing it.

\section{Empirical Studies}\label{sec:exp}
\textit{How does \alg{} compare with fixed curation methods and basic agents in real tasks?} In this section, we investigate the empirical effectiveness of agentic test-time curation with \alg, covering diverse scenarios from computational biology tasks (science domains) to prevalent machine learning tasks, such as post-training for large language models (LLMs). 

\vspace{-0.5em}\paragraph{Experiment Setup.}
\label{sec:setup}
The \textbf{science experiments} cover 3 biological domains, each with its own data type: protein sequence and functional data, genomics, and single-cell transcriptomics. We tested 5 tasks common in these domains, including protein variant effect prediction, non-coding variant effect prediction, cell type annotation, batch integration and perturbation response prediction, giving \textbf{55 dataset--model tasks} in total. In each task, the agent selects 50\% of the pool. \textbf{LLM post-training} experiments include \textit{general instruction tuning}, selecting 10{,}000 rows from the T\"ulu~3 dataset \citep{lambert2024tulu} or LESS mixture ~\citep{xia2024less} to fine-tune OLMo-2-1B or Llama-2-7B, and \textit{code instruction tuning}, selecting 10{,}000 rows from OpenCodeInstruct \citep{ahmad2025opencodeinstruct} to fine-tune Qwen2.5-Coder-1.5B. Every task is run under two agentic pipelines, \textit{Basic agent} and \alg. We test 3 agents with both agentic pipelines: Claude Code driven by Opus~4.8, Sonnet~5, or Haiku~4.5; Codex driven by GPT-5.3-Codex or GPT-5.6~Sol; and OpenHands driven by Qwen3.5-397B-A17B, an open-weight model. The agents are initiated from the same-seeded random subset, counted as the first iteration, and run 15 iterations in biology and 10 in LLM post-training track. Each task is scored by several metrics following major benchmarks. Performance is measured by a fixed scorer on a \textbf{held-out} test split \textit{disjoint} from training/validation splits; the agent observes only the resulting metrics. Additional details are provided in Appendix~\ref{app:setup}.

\subsection{Science Experiments: Biology Tasks  }
\label{sec:exp_bio}

\begin{wrapfigure}{r}{0.65\linewidth}
  \centering
\includegraphics[width=\linewidth]{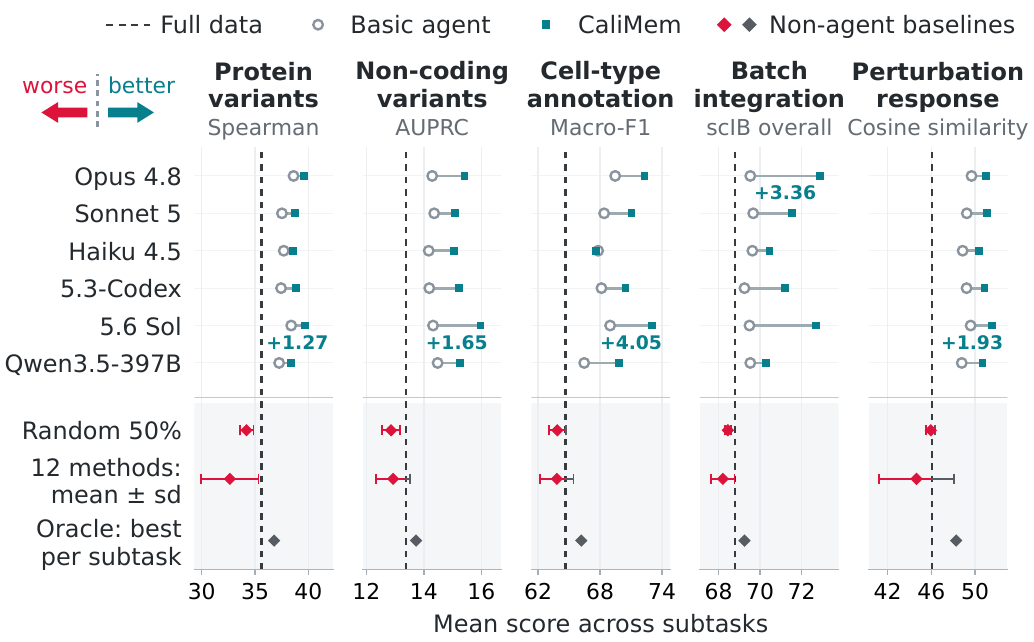}
\vspace{-1.5em}
    \caption{\small{\textbf{Main results on the biology domains.}
Each entry is the mean, over the dataset--model pairs of the domain,
of the best headline score reached within the 15-iteration budget, scaled to 0--100.
Random is the seeded subset every run starts from,
Full data trains on the entire pool without curation,
and the fixed-pipeline rows are standard curation recipes of each domain:
the mean $\pm$ sd over the 12 recipes, and the best recipe chosen separately for every subtask. Detailed results can be found in Appendix~\ref{app:res}.}}
  \label{fig:main-bio}
  \vspace{-2em}
\end{wrapfigure}

Figure~\ref{fig:main-bio} reports results across the 5 biological tasks.
\textbf{Fixed curation baselines:} For each biological domain, we instantiate 12 literature-derived curation recipes. The recipes span the major families of interventions commonly used for that data type—including quality filtering, balancing and stratification, coverage-based selection, and difficulty-based selection, etc. Appendix~\ref{app:bio-baselines}
 gives its source and exact implementation. These baselines represent a diverse set of reasonable human-designed choices available a priori, before observing how a particular downstream model responds. Figure~\ref{fig:5} shows the detailed comparison task by
task.

\vspace{-0.5em}\paragraph{Fixed curation is not reliably beneficial.}
Averaged over the 12 recipes, fixed curation underperforms full-data training in five biological tasks with a full-data reference, by 0.8--8.3\%, and even an oracle choosing the best recipe per dataset–model task improves on full data only modestly. The dependence is one of sign, not just magnitude: class balancing is redundant or harmful when the classifier already uses inverse-frequency weights, and coverage-based selection over-samples sparse, noisy regions unless unreliable examples are filtered first. A heuristic that corrects one model's weakness can remove useful signal for another.

\vspace{-0.5em}\paragraph{Test-time curation exploits this dependence.}
The Basic agent searches directly against the target pipeline and exceeds both Random and the fixed-recipe alternatives across agent families. In all 24 reported Basic-agent results with a full-data reference, the selected 50\% subset also outperforms training on the entire pool.
\vspace{-0.5em}\paragraph{\alg~further improves agentic search.} Across the 30 Basic-agent/\alg~pairs in Table~\ref{tab:main-bio},
\alg~improves 29, spanning protein, genomic, and single-cell tasks and both proprietary and open-weight agents. All 30 \alg~results with full-data references remain above full-data training despite using only half of the pool. By keeping conclusions tied to the conditions that produced them, CaliMem spends the fixed budget exploring alternative policies rather than committing early to a local one.

\definecolor{tabHeader}{RGB}{232,240,248}
\definecolor{tabRef}{RGB}{247,247,247}
\definecolor{tabClaude}{RGB}{243,248,253}
\definecolor{tabCodex}{RGB}{244,250,244}
\definecolor{tabOpen}{RGB}{250,246,240}
\definecolor{tabGain}{RGB}{0,128,60}
    \definecolor{tabLoss}{RGB}{220,20,60}   

\providecommand{\tabthresh}{5}
\providecommand{\tabentry}[4]{%
  \def\tabentrycol{.}%
  \ifdim #4pt>\tabthresh pt%
    \ifx#3-\def\tabentrycol{tabLoss}\else\def\tabentrycol{tabGain}\fi
  \fi
  \textcolor{\tabentrycol}{#1%
    \ifx\relax#2\relax\else\,{\tiny$\pm$\,#2}\fi
    \ (\ifx#3-$-$\else+\fi#4\%)}}

\subsection{LLM Post-Training Experiments: General and Code Instruction Tuning }
\label{sec:exp_llm}

\begin{wrapfigure}[21]{r}{0.6\linewidth}
  \vspace{-1.0em}
  \centering
\includegraphics[width=\linewidth]{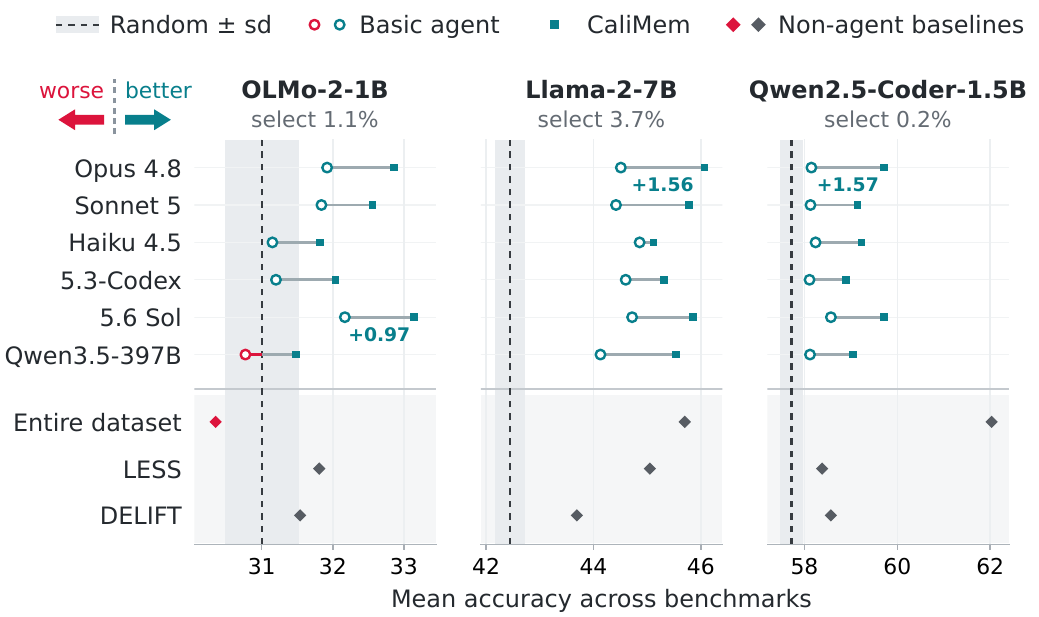}\vspace{-0.5em}
    \caption{\small{\textbf{Main results on LLM post-training tasks.}
Each entry is the best mean accuracy (\%) over the setup's benchmark suite
reached within the 10-iteration budget.
Random, LESS~\citep{xia2024less}, DELIFT~\citep{agarwal2025delift}, and the agents all select a subset of the same size
from the same pool (the selection ratio is given under each backbone):
the dashed line and band mark Random as the mean $\pm$ sd over seeds,
LESS and DELIFT are human-designed selection methods,
and Entire dataset trains on the entire pool without curation.
Points and segments left of the dashed line fall below Random.
Detailed results can be found in Appendix~\ref{app:res}.}}\vspace{-2.5em}
  \label{fig:main-llm}
\end{wrapfigure}
We next evaluate substantially more aggressive curation regimes in LLM post-training, retaining only 1.1\%, 3.7\%, or 0.2\% of the respective pools. Results are shown in Figure~\ref{fig:main-llm}.

\vspace{-0.5em}\paragraph{Agentic search remains effective at very small data budgets.}
The Basic agent improves over Random in 17 of 18 agent–task combinations, and \alg~improves over the corresponding Basic agent in all 18. On the two general instruction-tuning setups, the strongest curated subsets even exceed full-data training while using only 1.1\% or 3.7\% of the data.

\alg~is also competitive with the task-specific selection method such as LESS, matching or outperforming it in 17 of 18 comparisons. Notably, LESS needs warm-up training and per-example gradients over the pool while \alg~uses only the downstream train-and-evaluate interface. The discovered policies go beyond filtering heuristics: on Llama-2-7B one agent adapted RDS+ so that source allocation and within-source similarity ranking are decided separately (Appendix~\ref{app:casellm}).

\definecolor{tabHeader}{RGB}{232,240,248}
\definecolor{tabRef}{RGB}{247,247,247}
\definecolor{tabClaude}{RGB}{243,248,253}
\definecolor{tabCodex}{RGB}{244,250,244}
\definecolor{tabOpen}{RGB}{250,246,240}
\definecolor{tabGain}{RGB}{0,128,60}
\definecolor{tabLoss}{RGB}{220,20,60}   

\providecommand{\tabthresh}{5}
\providecommand{\tabentry}[4]{%
  \def\tabentrycol{.}%
  \ifdim #4pt>\tabthresh pt%
    \ifx#3-\def\tabentrycol{tabLoss}\else\def\tabentrycol{tabGain}\fi
  \fi
  \textcolor{\tabentrycol}{#1%
    \ifx\relax#2\relax\else\,{\tiny$\pm$\,#2}\fi
    \ (\ifx#3-$-$\else+\fi#4\%)}}

\begin{figure*}[htbp]
  \centering
  \includegraphics[width=\textwidth]{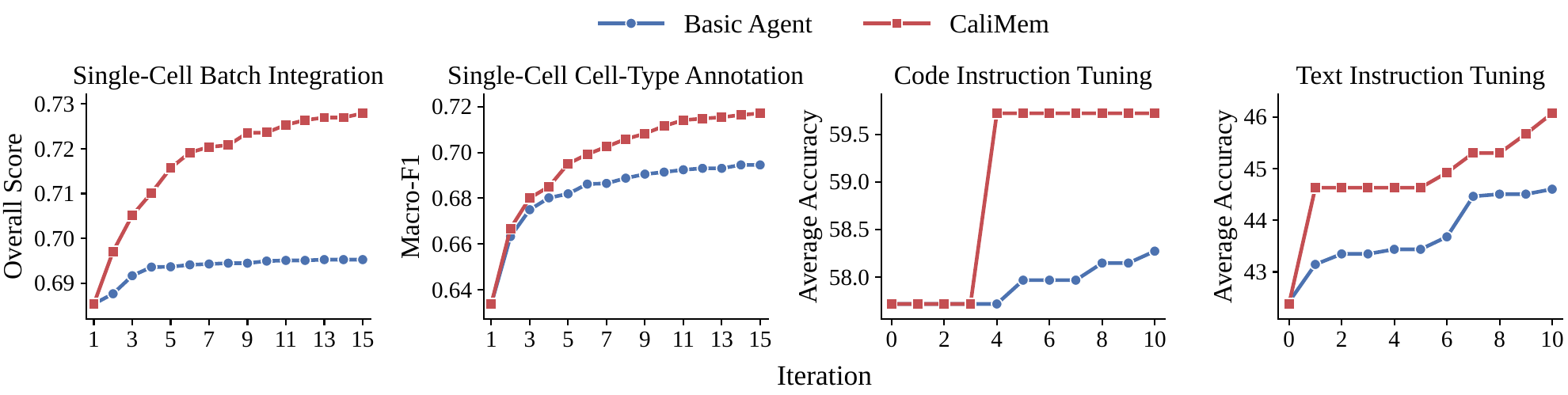}\vspace{-0.5em}
    \caption{\small{\textbf{Best score so far per iteration under the basic agent and
  \alg{}} (Claude Opus 4.8). Left two panels: single-cell batch integration and
  cell-type annotation, averaged over the 15 dataset--model tasks of each
  domain, 15 iterations. Right two panels: code instruction tuning
  (Qwen2.5-Coder-1.5B) and instruction tuning (Llama-2-7B on the LESS mixture),
  10 iterations. In every panel \alg{} passes the basic agent's final score
  within the first 5 iterations.}}
  \label{fig:4}
  \vspace{-1em}
\end{figure*}

Figure~\ref{fig:4} shows the best score reached so far as the agent spends its iteration budget. Across the various settings shown, \alg~consistently establishes
stronger incumbents earlier and finishes with a higher score. Calibrated memory improves the efficiency of the search itself, allowing evidence from earlier interventions to shape subsequent experiments more effectively.

\begin{tcolorbox}
\textbf{\textcolor{teal}{Takeaway: \alg{} is more search-efficient as well
as more accurate.}} It reaches the basic agent's final score in a third or fewer
of the iterations, and on the biology tasks it keeps improving after the basic
agent has plateaued.
\end{tcolorbox}

\section{Understanding \alg's Search Behavior}
The main results show that \alg~finds stronger policies with the same experimental budget. We next investigate why. First, we test whether its calibrated memory generalizes beyond summarizing past experiments by asking it to predict the outcomes of curation interventions it has not yet seen (Section~\ref{sec:forecast}). Second, we ablate persistent memory and prospective prediction to isolate their contributions (Section~\ref{sec:ablation}). Finally, we compare the adaptive policies found by the agent with fixed biological curation recipes to understand where adaptation provides its advantage (Section~\ref{sec:bio_baselines}).

\subsection{Does the Memory Predict Unseen Curation Interventions?}
\label{sec:forecast}

We test whether \alg’s memory captures reusable structure rather than merely summarizing past experiments. After iteration \(k\), we freeze the memory and give it to a fresh agent, which predicts the scores of policies executed later in the original trajectory without seeing their outcomes. For each later policy \(\pi_j\),
$ \hat y_{j|k}=\hat S(\pi_j\mid M_k),\,\, e_{j|k}=|y_j-\hat y_{j|k}|. $
We vary \(k\) across the trajectory and average forecasting error across dataset–model tasks (prompt in Appendix~\ref{app:exam-prompt}).
Figure~\ref{fig:matrix} shows the results for cell-type annotation. Along a row, error grows with extrapolation distance; down a column, the same target iteration becomes easier to predict as the memory is fitted on a longer prefix. The other domains show the same two patterns (Appendix~\ref{app:exam}).

\begin{tcolorbox}
\textbf{\textcolor{olive}{Finding: \alg~remembers beyond which policies succeeded.}} As the trajectory grows, its memory becomes increasingly predictive of unseen
interventions, suggesting that the calibrated claims encode reusable structure about the
task's response to data curation.
\end{tcolorbox}

\subsection{Are Memory and Prediction Both Necessary?}
\label{sec:ablation}

We separately ablate persistent memory and prospective prediction on protein variant-effect prediction and single-cell batch integration. Memory alone improves the Basic agent by 0.44 and 1.71 points, while prediction alone changes performance by \(-0.09\) and \(+0.54\). Combining them performs best in both settings, improving a further 0.54 and 1.65 points over memory alone. 
We hypothesize that comparing predicted and observed scores helps the agent reassess the claims used to justify a policy, while persistent memory preserves the resulting revisions for subsequent decisions. Without explicit memory updates, the agent may recognize an unexpected outcome without consistently revising the assumptions used in later iterations.

\noindent
\begin{minipage}[c]{0.46\linewidth}
\begin{tcolorbox}[before skip=0pt, after skip=0pt]
\textbf{\textcolor{olive}{Finding: Persistent memory provides a compact, reusable summary of evidence and claims across iterations; prediction provides a prospective check on those claims.}} Memory alone improves over the Basic agent in both ablation settings, whereas prediction alone has a smaller and inconsistent effect. Combining the two performs best in both settings, consistent with prediction being most useful when its outcome can revise a persistent state that guides subsequent decisions.
\end{tcolorbox}
\end{minipage}\hfill
\begin{minipage}[c]{0.5\linewidth}
  \centering
  \includegraphics[width=\linewidth]{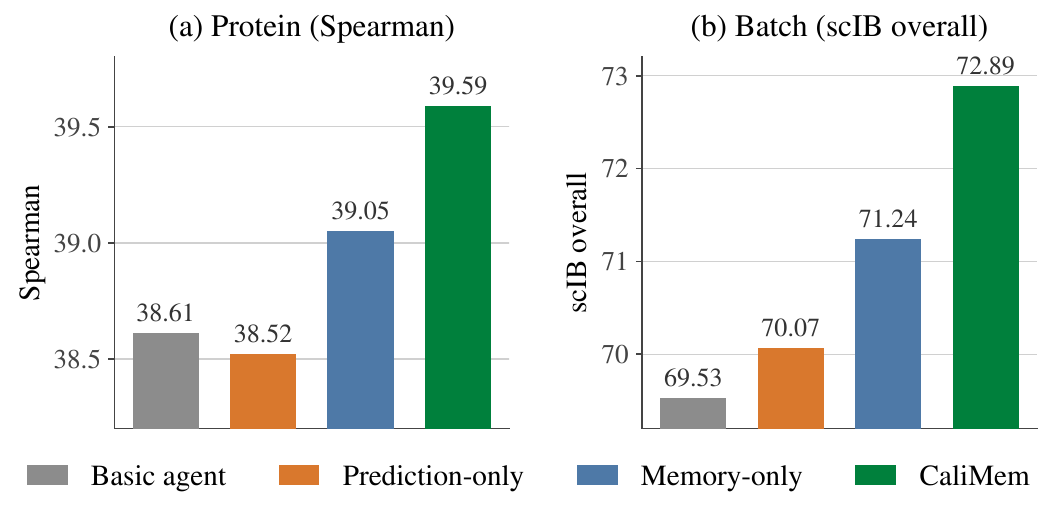}
  \vspace{-1.5em}
  \captionof{figure}{\small{\textbf{Ablation of calibrated memory.}
Results for Claude Opus~4.8 on protein variant-effect prediction and single-cell batch
integration.
Prediction-only adds a prospective prediction and comparison without persistent claims;
Memory-only maintains persistent evidence and claims but does not require a committed
prediction before each experiment.
The complete \alg~combines both.}}
  \label{fig:ablation}
\end{minipage}
\vspace{0.5em}



\subsection{Comparison with Fixed Biological Curation Pipelines}
\label{sec:bio_baselines}
Figure~\ref{fig:5} breaks the fixed-recipe comparison down to individual recipes. The same classes of intervention (quality filtering, balancing, stratification, coverage, difficulty-based selection) help in one setting and hurt in another, so no recipe in our set transfers uniformly across datasets and models. In aggregate (Table~\ref{tab:main-bio}), the mean recipe is below full data in most domain, and even a per-task oracle closes only a small part of the gap.
Test-time curation searches beyond this fixed menu: the agent alters thresholds and allocations, combines mechanisms, changes representations, or writes new policies in response to the measured pipeline. Notably, many of its strongest policies build on the same heuristics (Appendix~\ref{app:bio-baselines}). Their value is conditional on the downstream model and on interactions with the rest of the policy--the search supplies the adaptation a fixed recipe cannot.

\begin{tcolorbox}
\textbf{\textcolor{olive}{Finding: The limitation of fixed curation is not a lack of useful heuristics, but a lack of
adaptation.}} Standard biological pipelines contain valuable prior knowledge, yet their effects vary across
downstream models.
Test-time curation turns those heuristics into a searchable policy space, and \alg~uses calibrated
memory to explore that space more reliably.
\end{tcolorbox}

\section{Conclusion}
\label{sec:conclusion}
 Data curation is usually done once, with a fixed recipe, and reused. Across 55 dataset--model tasks in biological domains, we found this practice non-ideal: the same recipe helps one frozen model and hurts another, and averaged over recipes it is no better than a random subset of the same size. We therefore treated curation as agentic search carried out at deployment against the fixed downstream pipeline, test-time curation. A basic agent already improves on every fixed baseline. Adding persistent memory lets it use its whole trajectory and exposes a deeper failure: claims that reach beyond their evidence and steer the rest of the run. \alg{} makes every claim answerable by requiring a prediction before each experiment and writing the prediction error back into the claims. The resulting agent substantially improves on the basic agent in 47 of
48 models, exceeds full-data training by up to 19.4\%, and reaches the basic agent's final score in a third of the iterations or fewer.

 \vspace{-0.5em}\paragraph{Limitations and future work.}\emph{1. Noise in single-run feedback.} Each iteration is a single training run, so the score the agent reacts to, and the prediction error it learns from, carry run-to-run noise. \emph{2. Cost of test-time curation.} Test-time curation exchanges computation for data utility. Our budgets are practical in the settings studied here, but they would be expensive for substantially larger models or training recipes. Cheaper proxy models, early stopping, learned surrogates, and adaptive allocation of the budget could all reduce the cost. A natural extension is an agent that predicts and decides \emph{which experiments are worth paying for}. \emph{3. From curation policies to scientific hypotheses.} Connecting the agent's empirical claims to explicit hypothesis testing, uncertainty-aware experimental design, and review by domain experts is future work, and it matters most where curation decisions guide the acquisition of new experimental data.

\clearpage

\section*{AI Use Statement}
 
Generative AI is both the object of this study and a tool used in producing
it, and we separate the two. As the object of study, LLM-driven coding agents
(Claude Code, Codex, and OpenHands, driven by the models listed in
Section~\ref{sec:setup}) proposed, implemented, and revised every curation
policy reported in this paper, under the fixed harness of
Appendix~\ref{app:domains}. We have not used generative AI
to formulate or prove mathematical
claims or to propose the research hypotheses. We have reviewed all AI-assisted work. We take responsibility for the final content of this work,
including text, claims, or artifacts produced with the aid of generative AI.
 
\section*{Ethics Statement}
 
All datasets used in this work are public research benchmarks used under
their released terms: the deep mutational scanning assays of ProteinGym, the
non-coding variant sets distributed by BEND (ClinVar, an IGVF reporter assay,
and GRASP against 1000 Genomes), the scIB single-cell atlases, the PerturBench
screens, and the T\"ulu~3, LESS, and OpenCodeInstruct instruction-tuning
pools. The human-derived biological data are de-identified aggregate or
cell-level measurements released for research; we collected no new data and
conducted no human-subject studies. Automated data curation is a
general-purpose capability that, like any tool for improving training data,
could be applied to models built for harmful ends; we see no risk specific to
this work beyond that of the underlying models and datasets.
 
\section*{Reproducibility Statement}
We release the code, the evaluation harness, agent protocol files (Basic
and \alg{}), implementation, and example trajectory logs of the runs reported in the paper. Appendix~\ref{app:domains} specifies each domain's
datasets, held-out splits, composition floors, frozen models, metrics, and
validity guards; Appendix~\ref{app:llm-recipe} the fine-tuning
hyperparameters of the three instruction-tuning setups; and Appendix~\ref{app:exam-prompt} the forecasting prompt.
Every run starts from the same seeded random subset, the frozen models train
with fixed seeds, the code setup uses deterministic kernels, and each
human-designed recipe is materialised with 5 fill seeds. The agent runs
themselves are stochastic and several of the driving models are proprietary
and may change; we report the model identifiers.
 
\subsubsection*{Acknowledgments}
We acknowledge support from the National Science Foundation through grants IIS-2312794, IIS-2313130, OAC-2239622, CNS-2424127, OAC-2613761, and the Amazon--Virginia Tech Initiative for Efficient and Robust Machine Learning. This research is in part supported by the NVIDIA Academic Grant Program using NVIDIA Brev GPU instances.
\clearpage

\bibliography{iclr2027_conference}

@book{conn2009introduction,
  title={Introduction to derivative-free optimization},
  author={Conn, Andrew R and Scheinberg, Katya and Vicente, Luis N},
  year={2009},
  publisher={SIAM}
}

@article{jordana2025introduction,
  title={An introduction to zero-order optimization techniques for robotics},
  author={Jordana, Armand and Zhang, Jianghan and Amigo, Joseph and Righetti, Ludovic},
  journal={arXiv preprint arXiv:2506.22087},
  year={2025}
}

@article{kaplan2020scaling,
  title={Scaling Laws for Neural Language Models},
  author={Kaplan, Jared and McCandlish, Sam and Henighan, Tom and Brown, Tom B. and Chess, Benjamin and Child, Rewon and Gray, Scott and Radford, Alec and Wu, Jeffrey and Amodei, Dario},
  journal={arXiv preprint arXiv:2001.08361},
  year={2020}
}

@article{sorscher2022beyond,
  title={Beyond Neural Scaling Laws: Beating Power Law Scaling via Data Pruning},
  author={Sorscher, Ben and Geirhos, Robert and Shekhar, Shashank and Ganguli, Surya and Morcos, Ari},
  journal={Advances in Neural Information Processing Systems},
  volume={35},
  pages={19523--19536},
  year={2022}
}

@article{muennighoff2023scaling,
  title={Scaling Data-Constrained Language Models},
  author={Muennighoff, Niklas and Rush, Alexander and Barak, Boaz and Le Scao, Teven and Tazi, Nouamane and Piktus, Aleksandra and Pyysalo, Sampo and Wolf, Thomas and Raffel, Colin A.},
  journal={Advances in Neural Information Processing Systems},
  volume={36},
  pages={50358--50376},
  year={2023}
}

@article{wang2025selecting,
  title={Selecting Auxiliary Data via Neural Tangent Kernels for Low-Resource Domains},
  author={Wang, Pingjie and Liu, Hongcheng and Liao, Yusheng and Fan, Ziqing and Du, Yaxin and Tang, Shuo and Wang, Yanfeng and Wang, Yu},
  year={2025}
}

@article{gadre2023datacomp,
  title={{DataComp}: In Search of the Next Generation of Multimodal Datasets},
  author={Gadre, Samir Yitzhak and Ilharco, Gabriel and Fang, Alex and Hayase, Jonathan and Smyrnis, Georgios and Nguyen, Thao and Marten, Ryan and Wortsman, Mitchell and Ghosh, Dhruba and Zhang, Jieyu and Orgad, Eyal and Entezari, Rahim and Daras, Giannis and Pratt, Sarah and Ramanujan, Vivek and Bitton, Yonatan and others},
  journal={Advances in Neural Information Processing Systems},
  volume={36},
  pages={27092--27112},
  year={2023}
}

@inproceedings{lee2022dedup,
  title={Deduplicating Training Data Makes Language Models Better},
  author={Lee, Katherine and Ippolito, Daphne and Nystrom, Andrew and Zhang, Chiyuan and Eck, Douglas and Callison-Burch, Chris and Carlini, Nicholas},
  booktitle={Proceedings of the 60th Annual Meeting of the Association for Computational Linguistics (Volume 1: Long Papers)},
  pages={8424--8445},
  year={2022}
}

@article{xie2023doremi,
  title={{DoReMi}: Optimizing Data Mixtures Speeds Up Language Model Pretraining},
  author={Xie, Sang Michael and Pham, Hieu and Dong, Xuanyi and Du, Nan and Liu, Hanxiao and Lu, Yifeng and Liang, Percy S. and Le, Quoc V. and Ma, Tengyu and Yu, Adams Wei},
  journal={Advances in Neural Information Processing Systems},
  volume={36},
  pages={69798--69818},
  year={2023}
}

@article{xia2024less,
  title={{LESS}: Selecting Influential Data for Targeted Instruction Tuning},
  author={Xia, Mengzhou and Malladi, Sadhika and Gururangan, Suchin and Arora, Sanjeev and Chen, Danqi},
  journal={arXiv preprint arXiv:2402.04333},
  year={2024}
}

@article{wu2024icons,
  title={{ICONS}: Influence Consensus for Vision-Language Data Selection},
  author={Wu, Xindi and Xia, Mengzhou and Shao, Rulin and Deng, Zhiwei and Koh, Pang Wei and Russakovsky, Olga},
  journal={arXiv preprint arXiv:2501.00654},
  year={2024}
}

@article{chen2026olmix,
  title={Olmix: A Framework for Data Mixing Throughout {LM} Development},
  author={Chen, Mayee F. and Murray, Tyler and Heineman, David and Jordan, Matt and Hajishirzi, Hannaneh and R{\'e}, Christopher and Soldaini, Luca and Lo, Kyle},
  journal={arXiv preprint arXiv:2602.12237},
  year={2026}
}

@article{kang2026curationbench,
  title={Can Generalist Agents Automate Data Curation?},
  author={Kang, Feiyang and Li, Hanze and Nguyen, Adam and Dabas, Mahavir and Ma, Jiaqi W. and Sala, Frederic and Song, Dawn and Jia, Ruoxi},
  journal={arXiv preprint arXiv:2606.04261},
  year={2026}
}

@article{chan2024mlebench,
  title={{MLE}-bench: Evaluating Machine Learning Agents on Machine Learning Engineering},
  author={Chan, Jun Shern and Chowdhury, Neil and Jaffe, Oliver and Aung, James and Sherburn, Dane and Mays, Evan and Starace, Giulio and Liu, Kevin and Maksin, Leon and Patwardhan, Tejal and others},
  journal={arXiv preprint arXiv:2410.07095},
  year={2024}
}

@misc{karpathy2026autoresearch,
  title={autoresearch: {AI} agents running research on single-{GPU} nanochat training automatically},
  author={Karpathy, Andrej},
  year={2026},
  howpublished={\url{https://github.com/karpathy/autoresearch}},
  note={Accessed 2026-09-08}
}

@article{rank2026posttrainbench,
  title={{PostTrainBench}: Can {LLM} Agents Automate {LLM} Post-Training?},
  author={Rank, Ben and Bhatnagar, Hardik and Prabhu, Ameya and Eisenberg, Shira and Nguyen, Karina and Bethge, Matthias and Andriushchenko, Maksym},
  journal={arXiv preprint arXiv:2603.08640},
  year={2026}
}

@article{zhang2025ace,
  title={Agentic Context Engineering: Evolving Contexts for Self-Improving Language Models},
  author={Zhang, Qizheng and Hu, Changran and Upasani, Shubhangi and Ma, Boyuan and Hong, Fenglu and Kamanuru, Vamsidhar and Rainton, Jay and Wu, Chen and Ji, Mengmeng and Li, Hanchen and Thakker, Urmish and Zou, James and Olukotun, Kunle},
  journal={arXiv preprint arXiv:2510.04618},
  year={2025}
}

@article{shinn2023reflexion,
  title={Reflexion: Language Agents with Verbal Reinforcement Learning},
  author={Shinn, Noah and Cassano, Federico and Berman, Edward and Gopinath, Ashwin and Narasimhan, Karthik and Yao, Shunyu},
  journal={arXiv preprint arXiv:2303.11366},
  year={2023},
  note={NeurIPS 2023}
}

@article{luecken2022scib,
  title={Benchmarking Atlas-Level Data Integration in Single-Cell Genomics},
  author={Luecken, Malte D. and B{\"u}ttner, Maren and Chaichoompu, Kridsadakorn and Danese, Anna and Interlandi, Marta and Mueller, Michaela F. and Strobl, Daniel C. and Zappia, Luke and Dugas, Martin and Colom{\'e}-Tatch{\'e}, Maria and Theis, Fabian J.},
  journal={Nature Methods},
  volume={19},
  number={1},
  pages={41--50},
  year={2022}
}

@article{luecken2019practices,
  author  = {Luecken, Malte D. and Theis, Fabian J.},
  title   = {Current best practices in single-cell {RNA}-seq analysis: a tutorial},
  journal = {Molecular Systems Biology},
  volume  = {15},
  number  = {6},
  pages   = {e8746},
  year    = {2019},
  doi     = {10.15252/msb.20188746}
}

@article{heumos2023practices,
  author  = {Heumos, Lukas and Schaar, Anna C. and Lance, Christopher and Litinetskaya, Anastasia and Drost, Felix and Zappia, Luke and L{\"u}cken, Malte D. and Strobl, Daniel C. and Henao, Juan and Curion, Fabiola and {Single-cell Best Practices Consortium} and Schiller, Herbert B. and Theis, Fabian J.},
  title   = {Best practices for single-cell analysis across modalities},
  journal = {Nature Reviews Genetics},
  volume  = {24},
  number  = {8},
  pages   = {550--572},
  year    = {2023},
  doi     = {10.1038/s41576-023-00586-w}
}

@inproceedings{notin2023proteingym,
  author    = {Notin, Pascal and Kollasch, Aaron and Ritter, Daniel and van Niekerk, Lood and Paul, Steffanie and Spinner, Han and Rollins, Nathan and Shaw, Ada and Orenbuch, Rose and Weitzman, Ruben and Frazer, Jonathan and Dias, Mafalda and Franceschi, Dinko and Gal, Yarin and Marks, Debora},
  title     = {{ProteinGym}: Large-Scale Benchmarks for Protein Fitness Prediction and Design},
  booktitle = {Advances in Neural Information Processing Systems},
  volume    = {36},
  pages     = {64331--64379},
  year      = {2023},
  doi       = {10.52202/075280-2810}
}

@inproceedings{marin2024bend,
  author    = {Marin, Frederikke Isa and Teufel, Felix and Horlacher, Marc and Madsen, Dennis and Pultz, Dennis and Winther, Ole and Boomsma, Wouter},
  title     = {{BEND}: Benchmarking {DNA} Language Models on Biologically Meaningful Tasks},
  booktitle = {International Conference on Learning Representations (ICLR)},
  pages     = {15246--15281},
  year      = {2024},
  url       = {https://proceedings.iclr.cc/paper_files/paper/2024/hash/429e7b31625a8b7839f9e4d6e2aa9bb9-Abstract-Conference.html}
}

@article{peidli2024scperturb,
  author  = {Peidli, Stefan and Green, Tessa D. and Shen, Ciyue and Gross, Torsten and Min, Joseph and Garda, Samuele and Yuan, Bo and Schumacher, Linus J. and Taylor-King, Jake P. and Marks, Debora S. and Luna, Augustin and Bl{\"u}thgen, Nils and Sander, Chris},
  title   = {{scPerturb}: harmonized single-cell perturbation data},
  journal = {Nature Methods},
  volume  = {21},
  number  = {3},
  pages   = {531--540},
  year    = {2024},
  doi     = {10.1038/s41592-023-02144-y}
}

@article{vieth2019pipelines,
  author  = {Vieth, Beate and Parekh, Swati and Ziegenhain, Christoph and Enard, Wolfgang and Hellmann, Ines},
  title   = {A systematic evaluation of single cell {RNA}-seq analysis pipelines},
  journal = {Nature Communications},
  volume  = {10},
  number  = {1},
  pages   = {4667},
  year    = {2019},
  doi     = {10.1038/s41467-019-12266-7}
}

@article{germain2020pipecomp,
  author  = {Germain, Pierre-Luc and Sonrel, Anthony and Robinson, Mark D.},
  title   = {{pipeComp}, a general framework for the evaluation of computational pipelines, reveals performant single cell {RNA}-seq preprocessing tools},
  journal = {Genome Biology},
  volume  = {21},
  number  = {1},
  pages   = {227},
  year    = {2020},
  doi     = {10.1186/s13059-020-02136-7}
}

@article{rubin2017enrich2,
  author  = {Rubin, Alan F. and Gelman, Hannah and Lucas, Nathan and Bajjalieh, Sandra M. and Papenfuss, Anthony T. and Speed, Terence P. and Fowler, Douglas M.},
  title   = {A statistical framework for analyzing deep mutational scanning data},
  journal = {Genome Biology},
  volume  = {18},
  number  = {1},
  pages   = {150},
  year    = {2017},
  doi     = {10.1186/s13059-017-1272-5}
}

@article{faure2020dimsum,
  author  = {Faure, Andre J. and Schmiedel, J{\"o}rn M. and Baeza-Centurion, Pablo and Lehner, Ben},
  title   = {{DiMSum}: an error model and pipeline for analyzing deep mutational scanning data and diagnosing common experimental pathologies},
  journal = {Genome Biology},
  volume  = {21},
  number  = {1},
  pages   = {207},
  year    = {2020},
  doi     = {10.1186/s13059-020-02091-3}
}

@article{gelman2021neural,
  author  = {Gelman, Sam and Fahlberg, Sarah A. and Heinzelman, Pete and Romero, Philip A. and Gitter, Anthony},
  title   = {Neural networks to learn protein sequence--function relationships from deep mutational scanning data},
  journal = {Proceedings of the National Academy of Sciences},
  volume  = {118},
  number  = {48},
  pages   = {e2104878118},
  year    = {2021},
  doi     = {10.1073/pnas.2104878118}
}

@article{hsu2022learning,
  author  = {Hsu, Chloe and Nisonoff, Hunter and Fannjiang, Clara and Listgarten, Jennifer},
  title   = {Learning protein fitness models from evolutionary and assay-labeled data},
  journal = {Nature Biotechnology},
  volume  = {40},
  number  = {7},
  pages   = {1114--1122},
  year    = {2022},
  doi     = {10.1038/s41587-021-01146-5}
}

@article{wittmann2021informed,
  author  = {Wittmann, Bruce J. and Yue, Yisong and Arnold, Frances H.},
  title   = {Informed training set design enables efficient machine learning-assisted directed protein evolution},
  journal = {Cell Systems},
  volume  = {12},
  number  = {11},
  pages   = {1026--1045.e7},
  year    = {2021},
  doi     = {10.1016/j.cels.2021.07.008}
}

@article{sruthi2020deep2full,
  author  = {Sruthi, C. K. and Prakash, Meher},
  title   = {{Deep2Full}: Evaluating strategies for selecting the minimal mutational experiments for optimal computational predictions of deep mutational scan outcomes},
  journal = {PLOS ONE},
  volume  = {15},
  number  = {1},
  pages   = {e0227621},
  year    = {2020},
  doi     = {10.1371/journal.pone.0227621}
}

@inproceedings{dallago2021flip,
  author    = {Dallago, Christian and Mou, Jody and Johnston, Kadina E. and Wittmann, Bruce J. and Bhattacharya, Nicholas and Goldman, Samuel and Madani, Ali and Yang, Kevin K.},
  title     = {{FLIP}: Benchmark tasks in fitness landscape inference for proteins},
  booktitle = {Proceedings of the Neural Information Processing Systems Track on Datasets and Benchmarks},
  volume    = {1},
  year      = {2021},
  url       = {https://datasets-benchmarks-proceedings.neurips.cc/paper/2021/hash/2b44928ae11fb9384c4cf38708677c48-Abstract-round2.html}
}

@inproceedings{sener2018coreset,
  author    = {Sener, Ozan and Savarese, Silvio},
  title     = {Active Learning for Convolutional Neural Networks: A Core-Set Approach},
  booktitle = {International Conference on Learning Representations (ICLR)},
  year      = {2018},
  url       = {https://arxiv.org/abs/1708.00489}
}

@article{gonzalez1985clustering,
  author  = {Gonzalez, Teofilo F.},
  title   = {Clustering to minimize the maximum intercluster distance},
  journal = {Theoretical Computer Science},
  volume  = {38},
  pages   = {293--306},
  year    = {1985},
  doi     = {10.1016/0304-3975(85)90224-5}
}

@article{kircher2014cadd,
  author  = {Kircher, Martin and Witten, Daniela M. and Jain, Preti and O'Roak, Brian J. and Cooper, Gregory M. and Shendure, Jay},
  title   = {A general framework for estimating the relative pathogenicity of human genetic variants},
  journal = {Nature Genetics},
  volume  = {46},
  number  = {3},
  pages   = {310--315},
  year    = {2014},
  doi     = {10.1038/ng.2892}
}

@article{whalen2022pitfalls,
  author  = {Whalen, Sean and Schreiber, Jacob and Noble, William S. and Pollard, Katherine S.},
  title   = {Navigating the pitfalls of applying machine learning in genomics},
  journal = {Nature Reviews Genetics},
  volume  = {23},
  number  = {3},
  pages   = {169--181},
  year    = {2022},
  doi     = {10.1038/s41576-021-00434-9}
}

@article{ghandi2014gkmsvm,
  author  = {Ghandi, Mahmoud and Lee, Dongwon and Mohammad-Noori, Morteza and Beer, Michael A.},
  title   = {Enhanced Regulatory Sequence Prediction Using Gapped k-mer Features},
  journal = {PLOS Computational Biology},
  volume  = {10},
  number  = {7},
  pages   = {e1003711},
  year    = {2014},
  doi     = {10.1371/journal.pcbi.1003711}
}

@article{lee2015deltasvm,
  author  = {Lee, Dongwon and Gorkin, David U. and Baker, Maggie and Strober, Benjamin J. and Asoni, Alessandro L. and McCallion, Andrew S. and Beer, Michael A.},
  title   = {A method to predict the impact of regulatory variants from {DNA} sequence},
  journal = {Nature Genetics},
  volume  = {47},
  number  = {8},
  pages   = {955--961},
  year    = {2015},
  doi     = {10.1038/ng.3331}
}

@article{purcell2007plink,
  author  = {Purcell, Shaun and Neale, Benjamin and Todd-Brown, Kathe and Thomas, Lori and Ferreira, Manuel A. R. and Bender, David and Maller, Julian and Sklar, Pamela and de Bakker, Paul I. W. and Daly, Mark J. and Sham, Pak C.},
  title   = {{PLINK}: A Tool Set for Whole-Genome Association and Population-Based Linkage Analyses},
  journal = {The American Journal of Human Genetics},
  volume  = {81},
  number  = {3},
  pages   = {559--575},
  year    = {2007},
  doi     = {10.1086/519795}
}

@article{landrum2018clinvar,
  author  = {Landrum, Melissa J. and Lee, Jennifer M. and Benson, Mark and Brown, Garth R. and Chao, Chen and Chitipiralla, Shanmuga and Gu, Baoshan and Hart, Jennifer and Hoffman, Douglas and Jang, Wonhee and Karapetyan, Karen and Katz, Kenneth and Liu, Chunlei and Maddipatla, Zenith and Malheiro, Adriana and McDaniel, Kurt and Ovetsky, Michael and Riley, George and Zhou, George and Holmes, J. Bradley and Kattman, Brandi L. and Maglott, Donna R.},
  title   = {{ClinVar}: improving access to variant interpretations and supporting evidence},
  journal = {Nucleic Acids Research},
  volume  = {46},
  number  = {D1},
  pages   = {D1062--D1067},
  year    = {2018},
  doi     = {10.1093/nar/gkx1153}
}

@inproceedings{shrivastava2016ohem,
  author    = {Shrivastava, Abhinav and Gupta, Abhinav and Girshick, Ross},
  title     = {Training Region-Based Object Detectors with Online Hard Example Mining},
  booktitle = {Proceedings of the IEEE Conference on Computer Vision and Pattern Recognition (CVPR)},
  pages     = {761--769},
  year      = {2016},
  doi       = {10.1109/CVPR.2016.89}
}

@article{zhou2015deepsea,
  author  = {Zhou, Jian and Troyanskaya, Olga G.},
  title   = {Predicting effects of noncoding variants with deep learning--based sequence model},
  journal = {Nature Methods},
  volume  = {12},
  number  = {10},
  pages   = {931--934},
  year    = {2015},
  doi     = {10.1038/nmeth.3547}
}

@article{wolf2018scanpy,
  author  = {Wolf, F. Alexander and Angerer, Philipp and Theis, Fabian J.},
  title   = {{SCANPY}: large-scale single-cell gene expression data analysis},
  journal = {Genome Biology},
  volume  = {19},
  number  = {1},
  pages   = {15},
  year    = {2018},
  doi     = {10.1186/s13059-017-1382-0}
}

@article{hao2021seurat,
  author  = {Hao, Yuhan and Hao, Stephanie and Andersen-Nissen, Erica and Mauck, III, William M. and Zheng, Shiwei and Butler, Andrew and Lee, Maddie J. and Wilk, Aaron J. and Darby, Charlotte and Zager, Michael and Hoffman, Paul and Stoeckius, Marlon and Papalexi, Efthymia and Mimitou, Eleni P. and Jain, Jaison and Srivastava, Avi and Stuart, Tim and Fleming, Lamar M. and Yeung, Bertrand and Rogers, Angela J. and McElrath, Juliana M. and Blish, Catherine A. and Gottardo, Raphael and Smibert, Peter and Satija, Rahul},
  title   = {Integrated analysis of multimodal single-cell data},
  journal = {Cell},
  volume  = {184},
  number  = {13},
  pages   = {3573--3587.e29},
  year    = {2021},
  doi     = {10.1016/j.cell.2021.04.048}
}

@article{lun2016stepbystep,
  author  = {Lun, Aaron T. L. and McCarthy, Davis J. and Marioni, John C.},
  title   = {A step-by-step workflow for low-level analysis of single-cell {RNA}-seq data with {Bioconductor}},
  journal = {F1000Research},
  volume  = {5},
  pages   = {2122},
  year    = {2016},
  doi     = {10.12688/f1000research.9501.2}
}

@article{germain2021scdblfinder,
  author  = {Germain, Pierre-Luc and Lun, Aaron and Garcia Meixide, Carlos and Macnair, Will and Robinson, Mark D.},
  title   = {Doublet identification in single-cell sequencing data using {scDblFinder}},
  journal = {F1000Research},
  volume  = {10},
  pages   = {979},
  year    = {2022},
  doi     = {10.12688/f1000research.73600.2}
}

@article{wolock2019scrublet,
  author  = {Wolock, Samuel L. and Lopez, Romain and Klein, Allon M.},
  title   = {{Scrublet}: Computational Identification of Cell Doublets in Single-Cell Transcriptomic Data},
  journal = {Cell Systems},
  volume  = {8},
  number  = {4},
  pages   = {281--291.e9},
  year    = {2019},
  doi     = {10.1016/j.cels.2018.11.005}
}

@article{xi2021doublet,
  author  = {Xi, Nan Miles and Li, Jingyi Jessica},
  title   = {Benchmarking Computational Doublet-Detection Methods for Single-Cell {RNA} Sequencing Data},
  journal = {Cell Systems},
  volume  = {12},
  number  = {2},
  pages   = {176--194.e6},
  year    = {2021},
  doi     = {10.1016/j.cels.2020.11.008}
}

@article{dominguezconde2022crosstissue,
  author  = {Dom{\'i}nguez Conde, C. and Xu, C. and Jarvis, L. B. and Rainbow, D. B. and Wells, S. B. and Gomes, T. and Howlett, S. K. and Suchanek, O. and Polanski, K. and King, H. W. and Mamanova, L. and Huang, N. and Szabo, P. A. and Richardson, L. and Bolt, L. and Fasouli, E. S. and Mahbubani, K. T. and Prete, M. and Tuck, L. and Richoz, N. and Tuong, Z. K. and Campos, L. and Mousa, H. S. and Needham, E. J. and Pritchard, S. and Li, T. and Elmentaite, R. and Park, J. and Rahmani, E. and Chen, D. and Menon, D. K. and Bayraktar, O. A. and James, L. K. and Meyer, K. B. and Yosef, N. and Clatworthy, M. R. and Sims, P. A. and Farber, D. L. and Saeb-Parsy, K. and Jones, J. L. and Teichmann, S. A.},
  title   = {Cross-tissue immune cell analysis reveals tissue-specific features in humans},
  journal = {Science},
  volume  = {376},
  number  = {6594},
  pages   = {eabl5197},
  year    = {2022},
  doi     = {10.1126/science.abl5197}
}

@article{abdelaal2019comparison,
  author  = {Abdelaal, Tamim and Michielsen, Lieke and Cats, Davy and Hoogduin, Dylan and Mei, Hailiang and Reinders, Marcel J. T. and Mahfouz, Ahmed},
  title   = {A comparison of automatic cell identification methods for single-cell {RNA} sequencing data},
  journal = {Genome Biology},
  volume  = {20},
  number  = {1},
  pages   = {194},
  year    = {2019},
  doi     = {10.1186/s13059-019-1795-z}
}

@article{lotfollahi2022scarches,
  author  = {Lotfollahi, Mohammad and Naghipourfar, Mohsen and Luecken, Malte D. and Khajavi, Matin and B{\"u}ttner, Maren and Wagenstetter, Marco and Avsec, {\v{Z}}iga and Gayoso, Adam and Yosef, Nir and Interlandi, Marta and Rybakov, Sergei and Misharin, Alexander V. and Theis, Fabian J.},
  title   = {Mapping single-cell data to reference atlases by transfer learning},
  journal = {Nature Biotechnology},
  volume  = {40},
  number  = {1},
  pages   = {121--130},
  year    = {2022},
  doi     = {10.1038/s41587-021-01001-7}
}

@article{kang2021symphony,
  author  = {Kang, Joyce B. and Nathan, Aparna and Weinand, Kathryn and Zhang, Fan and Millard, Nghia and Rumker, Laurie and Moody, D. Branch and Korsunsky, Ilya and Raychaudhuri, Soumya},
  title   = {Efficient and precise single-cell reference atlas mapping with {Symphony}},
  journal = {Nature Communications},
  volume  = {12},
  number  = {1},
  pages   = {5890},
  year    = {2021},
  doi     = {10.1038/s41467-021-25957-x}
}

@article{hie2019geosketch,
  author  = {Hie, Brian and Cho, Hyunghoon and DeMeo, Benjamin and Bryson, Bryan and Berger, Bonnie},
  title   = {Geometric Sketching Compactly Summarizes the Single-Cell Transcriptomic Landscape},
  journal = {Cell Systems},
  volume  = {8},
  number  = {6},
  pages   = {483--493.e7},
  year    = {2019},
  doi     = {10.1016/j.cels.2019.05.003}
}

@inproceedings{goyal2024scaling,
  title={Scaling laws for data filtering—data curation cannot be compute agnostic},
  author={Goyal, Sachin and Maini, Pratyush and Lipton, Zachary C and Raghunathan, Aditi and Kolter, J Zico},
  booktitle={2024 IEEE/CVF Conference on Computer Vision and Pattern Recognition (CVPR)},
  pages={22702--22711},
  year={2024},
  organization={IEEE}
}

@article{ahmad2025opencodeinstruct,
  title={Opencodeinstruct: A large-scale instruction tuning dataset for code llms},
  author={Ahmad, Wasi Uddin and Ficek, Aleksander and Samadi, Mehrzad and Huang, Jocelyn and Noroozi, Vahid and Majumdar, Somshubra and Ginsburg, Boris},
  journal={arXiv preprint arXiv:2504.04030},
  year={2025}
}

@article{lambert2024tulu,
  title={Tulu 3: Pushing frontiers in open language model post-training},
  author={Lambert, Nathan and Morrison, Jacob and Pyatkin, Valentina and Huang, Shengyi and Ivison, Hamish and Brahman, Faeze and Miranda, Lester James V and Liu, Alisa and Dziri, Nouha and Lyu, Shane and others},
  journal={arXiv preprint arXiv:2411.15124},
  year={2024}
}

@article{hendrycks2020measuring,
  title={Measuring massive multitask language understanding},
  author={Hendrycks, Dan and Burns, Collin and Basart, Steven and Zou, Andy and Mazeika, Mantas and Song, Dawn and Steinhardt, Jacob},
  journal={arXiv preprint arXiv:2009.03300},
  year={2020}
}

@article{ivison2025large,
  title={Large-scale data selection for instruction tuning},
  author={Ivison, Hamish and Zhang, Muru and Brahman, Faeze and Koh, Pang Wei and Dasigi, Pradeep},
  journal={arXiv preprint arXiv:2503.01807},
  year={2025}
}

@article{albalak2024survey,
  title={A survey on data selection for language models},
  author={Albalak, Alon and Elazar, Yanai and Xie, Sang Michael and Longpre, Shayne and Lambert, Nathan and Wang, Xinyi and Muennighoff, Niklas and Hou, Bairu and Pan, Liangming and Jeong, Haewon and others},
  journal={arXiv preprint arXiv:2402.16827},
  year={2024}
}

@inproceedings{agarwal2025delift,
  title={DELIFT: Data efficient language model instruction fine-tuning},
  author={Agarwal, Ishika and Killamsetty, Krishnateja and Popa, Lucian and Danilevsky, Marina},
  booktitle={International Conference on Learning Representations},
  volume={2025},
  pages={100488--100509},
  year={2025}
}

@inproceedings{huang2025dca,
  title={DCA-bench: a benchmark for dataset curation agents},
  author={Huang, Benhao and Yu, Yingzhuo and Huang, Jin and Zhang, Xingjian and Ma, Jiaqi W},
  booktitle={Proceedings of the 31st ACM SIGKDD Conference on Knowledge Discovery and Data Mining V. 2},
  pages={5482--5492},
  year={2025}
}

@inproceedings{zhang2026datascibench,
  title={Datascibench: An llm agent benchmark for data science},
  author={Zhang, Dan and Zhoubian, Sining and Cai, Min and Li, Fengzu and Yang, Lekang and Wang, Wei and Dong, Tianjiao and Hu, Ziniu and Tang, Jie and Yue, Yisong},
  booktitle={Findings of the Association for Computational Linguistics: ACL 2026},
  pages={3685--3728},
  year={2026}
}

@inproceedings{geng2024survey,
  title={A survey of confidence estimation and calibration in large language models},
  author={Geng, Jiahui and Cai, Fengyu and Wang, Yuxia and Koeppl, Heinz and Nakov, Preslav and Gurevych, Iryna},
  booktitle={Proceedings of the 2024 conference of the north American chapter of the association for computational linguistics: human language technologies (volume 1: long papers)},
  pages={6577--6595},
  year={2024}
}

@article{singh2026agent,
  title={Agent-brace: Decoupling beliefs from actions in long-horizon tasks via verbalized state uncertainty},
  author={Singh, Joykirat and Khan, Zaid and Prasad, Archiki and Chen, Justin Chih-Yao and Nambi, Akshay and Lee, Hyunji and Stengel-Eskin, Elias and Bansal, Mohit},
  journal={arXiv preprint arXiv:2605.11436},
  year={2026}
}

@article{richens2025general,
  title={General agents contain world models},
  author={Richens, Jonathan and Abel, David and Bellot, Alexis and Everitt, Tom},
  journal={arXiv preprint arXiv:2506.01622},
  year={2025}
}

@inproceedings{gkountouras2025language,
  title={Language Agents Meet Causality--Bridging LLMs and Causal World Models},
  author={Gkountouras, John and Lindemann, Matthias and Lippe, Phillip and Gavves, Efstratios and Titov, Ivan},
  booktitle={International Conference on Learning Representations},
  volume={2025},
  pages={37311--37336},
  year={2025}
}

@inproceedings{zhang2025aflow,
  title={Aflow: Automating agentic workflow generation},
  author={Zhang, Jiayi and Xiang, Jinyu and Yu, Zhaoyang and Teng, Fengwei and Chen, Xionghui and Chen, Jiaqi and Zhuge, Mingchen and Cheng, Xin and Hong, Sirui and Wang, Jinlin and others},
  booktitle={International Conference on Learning Representations},
  volume={2025},
  pages={34040--34077},
  year={2025}
}

@article{yuksekgonul2024textgrad,
  title={Textgrad: Automatic" differentiation" via text},
  author={Yuksekgonul, Mert and Bianchi, Federico and Boen, Joseph and Liu, Sheng and Huang, Zhi and Guestrin, Carlos and Zou, James},
  journal={arXiv preprint arXiv:2406.07496},
  year={2024}
}

@inproceedings{khattab2024dspy,
  title={Dspy: Compiling declarative language model calls into state-of-the-art pipelines},
  author={Khattab, Omar and Singhvi, Arnav and Maheshwari, Paridhi and Zhang, Zhiyuan and Santhanam, Keshav and Haq, Saiful and Sharma, Ashutosh and Joshi, Thomas and Moazam, Hanna and Miller, Heather and others},
  booktitle={International Conference on Learning Representations},
  volume={2024},
  pages={54928--54958},
  year={2024}
}

@article{liu2023your,
  title={Is your code generated by chatgpt really correct? rigorous evaluation of large language models for code generation},
  author={Liu, Jiawei and Xia, Chunqiu Steven and Wang, Yuyao and Zhang, Lingming},
  journal={Advances in neural information processing systems},
  volume={36},
  pages={21558--21572},
  year={2023}
}

@article{zhou2023instruction,
  title={Instruction-following evaluation for large language models},
  author={Zhou, Jeffrey and Lu, Tianjian and Mishra, Swaroop and Brahma, Siddhartha and Basu, Sujoy and Luan, Yi and Zhou, Denny and Hou, Le},
  journal={arXiv preprint arXiv:2311.07911},
  year={2023}
}

@inproceedings{lin2022truthfulqa,
  title={Truthfulqa: Measuring how models mimic human falsehoods},
  author={Lin, Stephanie and Hilton, Jacob and Evans, Owain},
  booktitle={Proceedings of the 60th annual meeting of the association for computational linguistics (volume 1: long papers)},
  pages={3214--3252},
  year={2022}
}

@inproceedings{suzgun2023challenging,
  title={Challenging big-bench tasks and whether chain-of-thought can solve them},
  author={Suzgun, Mirac and Scales, Nathan and Sch{\"a}rli, Nathanael and Gehrmann, Sebastian and Tay, Yi and Chung, Hyung Won and Chowdhery, Aakanksha and Le, Quoc and Chi, Ed H and Zhou, Denny and others},
  booktitle={Findings of the Association for Computational Linguistics: ACL 2023},
  pages={13003--13051},
  year={2023}
}

@article{cobbe2021training,
  title={Training verifiers to solve math word problems},
  author={Cobbe, Karl and Kosaraju, Vineet and Bavarian, Mohammad and Chen, Mark and Jun, Heewoo and Kaiser, Lukasz and Plappert, Matthias and Tworek, Jerry and Hilton, Jacob and Nakano, Reiichiro and others},
  journal={arXiv preprint arXiv:2110.14168},
  year={2021}
}

@inproceedings{suzgun2026dynamic,
  title={Dynamic cheatsheet: Test-time learning with adaptive memory},
  author={Suzgun, Mirac and Yuksekgonul, Mert and Bianchi, Federico and Jurafsky, Dan and Zou, James},
  booktitle={Proceedings of the 19th Conference of the European Chapter of the Association for Computational Linguistics (Volume 1: Long Papers)},
  pages={7080--7106},
  year={2026}
}

@inproceedings{zhao2024expel,
  title={Expel: Llm agents are experiential learners},
  author={Zhao, Andrew and Huang, Daniel and Xu, Quentin and Lin, Matthieu and Liu, Yong-Jin and Huang, Gao},
  booktitle={Proceedings of the AAAI Conference on Artificial Intelligence},
  volume={38},
  number={17},
  pages={19632--19642},
  year={2024}
}

@article{du2026datamaster,
  title={DataMaster: Data-Centric Autonomous AI Research},
  author={Du, Yaxin and Yang, Xiyuan and Zhou, Zhifan and Liu, Wanxu and Lei, Zixing and Chen, Zimeng and Liu, Fenyi and Wu, Haotian and Cai, Yuzhu and Liu, Zexi and others},
  journal={arXiv preprint arXiv:2605.10906},
  year={2026}
}

@article{thirukovalluru2025indata,
  title={InData: Towards Secure Multi-Step, Tool-Based Data Analysis},
  author={Thirukovalluru, Raghuveer and Dhingra, Bhuwan and Carlson, David Edwin and others},
  journal={arXiv preprint arXiv:2511.11933},
  year={2025}
}

@inproceedings{chen2025aioli,
  title={Aioli: A unified optimization framework for language model data mixing},
  author={Chen, Mayee and Hu, Michael and Lourie, Nicholas and Cho, Kyunghyun and R{\'e}, Christopher},
  booktitle={International Conference on Learning Representations},
  volume={2025},
  pages={46089--46132},
  year={2025}
}

@inproceedings{liu2025regmix,
  title={Regmix: Data mixture as regression for language model pre-training},
  author={Liu, Qian and Zheng, Xiaosen and Muennighoff, Niklas and Zeng, Guangtao and Dou, Longxu and Pang, Tianyu and Jiang, Jing and Lin, Min},
  booktitle={International Conference on Learning Representations},
  volume={2025},
  pages={38305--38339},
  year={2025}
}

@article{ilicic2016lowquality,
  author  = {Ilicic, Tomislav and Kim, Jong Kyoung and Kolodziejczyk, Aleksandra A. and Bagger, Frederik Otzen and McCarthy, Davis James and Marioni, John C. and Teichmann, Sarah A.},
  title   = {Classification of low quality cells from single-cell {RNA}-seq data},
  journal = {Genome Biology},
  volume  = {17},
  number  = {1},
  pages   = {29},
  year    = {2016},
  doi     = {10.1186/s13059-016-0888-1}
}

@article{ding2020systematic,
  author  = {Ding, Jiarui and Adiconis, Xian and Simmons, Sean K. and Kowalczyk, Monika S. and Hession, Cynthia C. and Marjanovic, Nemanja D. and Hughes, Travis K. and Wadsworth, Marc H. and Burks, Tyler and Nguyen, Lan T. and Kwon, John Y. H. and Barak, Boaz and Ge, William and Kedaigle, Amanda J. and Carroll, Shaina and Li, Shuqiang and Hacohen, Nir and Rozenblatt-Rosen, Orit and Shalek, Alex K. and Villani, Alexandra-Chlo{\'e} and Regev, Aviv and Levin, Joshua Z.},
  title   = {Systematic comparison of single-cell and single-nucleus {RNA}-sequencing methods},
  journal = {Nature Biotechnology},
  volume  = {38},
  number  = {6},
  pages   = {737--746},
  year    = {2020},
  doi     = {10.1038/s41587-020-0465-8}
}

@article{johnson2007combat,
  author  = {Johnson, W. Evan and Li, Cheng and Rabinovic, Ariel},
  title   = {Adjusting batch effects in microarray expression data using empirical {Bayes} methods},
  journal = {Biostatistics},
  volume  = {8},
  number  = {1},
  pages   = {118--127},
  year    = {2007},
  doi     = {10.1093/biostatistics/kxj037}
}

@article{korsunsky2019harmony,
  author  = {Korsunsky, Ilya and Millard, Nghia and Fan, Jean and Slowikowski, Kamil and Zhang, Fan and Wei, Kevin and Baglaenko, Yuriy and Brenner, Michael and Loh, Po-ru and Raychaudhuri, Soumya},
  title   = {Fast, sensitive and accurate integration of single-cell data with {Harmony}},
  journal = {Nature Methods},
  volume  = {16},
  number  = {12},
  pages   = {1289--1296},
  year    = {2019},
  doi     = {10.1038/s41592-019-0619-0}
}

@article{tran2020benchmark,
  author  = {Tran, Hoa Thi Nhu and Ang, Kok Siong and Chevrier, Marion and Zhang, Xiaomeng and Lee, Nicole Yee Shin and Goh, Michelle and Chen, Jinmiao},
  title   = {A benchmark of batch-effect correction methods for single-cell {RNA} sequencing data},
  journal = {Genome Biology},
  volume  = {21},
  number  = {1},
  pages   = {12},
  year    = {2020},
  doi     = {10.1186/s13059-019-1850-9}
}

@article{demeo2020hopper,
  author  = {DeMeo, Benjamin and Berger, Bonnie},
  title   = {{Hopper}: a mathematically optimal algorithm for sketching biological data},
  journal = {Bioinformatics},
  volume  = {36},
  number  = {Supplement\_1},
  pages   = {i236--i241},
  year    = {2020},
  doi     = {10.1093/bioinformatics/btaa408}
}

@article{haghverdi2018mnn,
  author  = {Haghverdi, Laleh and Lun, Aaron T. L. and Morgan, Michael D. and Marioni, John C.},
  title   = {Batch effects in single-cell {RNA}-sequencing data are corrected by matching mutual nearest neighbors},
  journal = {Nature Biotechnology},
  volume  = {36},
  number  = {5},
  pages   = {421--427},
  year    = {2018},
  doi     = {10.1038/nbt.4091}
}

@article{stuart2019comprehensive,
  author  = {Stuart, Tim and Butler, Andrew and Hoffman, Paul and Hafemeister, Christoph and Papalexi, Efthymia and Mauck, III, William M. and Hao, Yuhan and Stoeckius, Marlon and Smibert, Peter and Satija, Rahul},
  title   = {Comprehensive Integration of Single-Cell Data},
  journal = {Cell},
  volume  = {177},
  number  = {7},
  pages   = {1888--1902.e21},
  year    = {2019},
  doi     = {10.1016/j.cell.2019.05.031}
}

@article{welch2019liger,
  author  = {Welch, Joshua D. and Kozareva, Velina and Ferreira, Ashley and Vanderburg, Charles and Martin, Carly and Macosko, Evan Z.},
  title   = {Single-Cell Multi-omic Integration Compares and Contrasts Features of Brain Cell Identity},
  journal = {Cell},
  volume  = {177},
  number  = {7},
  pages   = {1873--1887.e17},
  year    = {2019},
  doi     = {10.1016/j.cell.2019.05.006}
}

@article{grantham1974amino,
  author  = {Grantham, R.},
  title   = {Amino Acid Difference Formula to Help Explain Protein Evolution},
  journal = {Science},
  volume  = {185},
  number  = {4154},
  pages   = {862--864},
  year    = {1974},
  doi     = {10.1126/science.185.4154.862}
}

@article{miyata1979two,
  author  = {Miyata, Takashi and Miyazawa, Sanzo and Yasunaga, Teruo},
  title   = {Two types of amino acid substitutions in protein evolution},
  journal = {Journal of Molecular Evolution},
  volume  = {12},
  number  = {3},
  pages   = {219--236},
  year    = {1979},
  doi     = {10.1007/BF01732340}
}

@article{henikoff1992blosum,
  author  = {Henikoff, S. and Henikoff, J. G.},
  title   = {Amino acid substitution matrices from protein blocks},
  journal = {Proceedings of the National Academy of Sciences},
  volume  = {89},
  number  = {22},
  pages   = {10915--10919},
  year    = {1992},
  doi     = {10.1073/pnas.89.22.10915}
}

@article{brodley1999identifying,
  author  = {Brodley, Carla E. and Friedl, Mark A.},
  title   = {Identifying Mislabeled Training Data},
  journal = {Journal of Artificial Intelligence Research},
  volume  = {11},
  pages   = {131--167},
  year    = {1999},
  doi     = {10.1613/jair.606}
}

@article{fowler2014dms,
  author  = {Fowler, Douglas M. and Fields, Stanley},
  title   = {Deep mutational scanning: a new style of protein science},
  journal = {Nature Methods},
  volume  = {11},
  number  = {8},
  pages   = {801--807},
  year    = {2014},
  doi     = {10.1038/nmeth.3027}
}

@inproceedings{wu2025perturbench,
  author    = {Wu, Yan and Wershof, Esther and Schmon, Sebastian M. and Nassar, Marcel and Osi{\'n}ski, B{\l}a{\.z}ej and Eksi, Ridvan and Yan, Zichao and Stark, Rory and Zhang, Kun and Graepel, Thore},
  title     = {{PerturBench}: Benchmarking Machine Learning Models for Cellular Perturbation Analysis},
  booktitle = {Advances in Neural Information Processing Systems},
  year      = {2025},
  eprint    = {2408.10609},
  archivePrefix = {arXiv}
}

@inproceedings{agrawal2025gepa,
  title={{GEPA}: Reflective Prompt Evolution Can Outperform Reinforcement Learning},
  author={Agrawal, Lakshya A and Tan, Shangyin and Soylu, Dilara and Ziems, Noah and Khare, Rishi and Opsahl-Ong, Krista and Singhvi, Arnav and Shandilya, Herumb and Ryan, Michael J and Jiang, Meng and Potts, Christopher and Sen, Koushik and Dimakis, Alexandros G. and Stoica, Ion and Klein, Dan and Zaharia, Matei and Khattab, Omar},
  booktitle={International Conference on Learning Representations (ICLR)},
  year={2026},
  note={arXiv:2507.19457}
}

@inproceedings{bjorck2024lr,
  title={Scaling Optimal {LR} Across Token Horizons},
  author={Bjorck, Johan and Benhaim, Alon and Chaudhary, Vishrav and Wei, Furu and Song, Xia},
  booktitle={International Conference on Learning Representations (ICLR)},
  year={2025},
  note={arXiv:2409.19913}
}

@article{cai2026mlautoresearch,
  title   = {{ML-AutoResearch}: Training Machine Learning Research Agents with Automatically Generated Environments},
  author  = {Cai, Ziyang and Saeidi, Amir and Behl, Harkirat},
  journal = {arXiv preprint arXiv:2603.17216},
  year    = {2026}
}

@inproceedings{chae2025wma,
  title={Web Agents with World Models: Learning and Leveraging Environment Dynamics in Web Navigation},
  author={Chae, Hyungjoo and Kim, Namyoung and Ong, Kai Tzu-iunn and Gwak, Minju and Song, Gwanwoo and Kim, Jihoon and Kim, Sunghwan and Lee, Dongha and Yeo, Jinyoung},
  booktitle={The Thirteenth International Conference on Learning Representations},
  year={2025}
}

@article{chang2026forecastcompass,
  title   = {{ForecastCompass}: Guiding Agentic Forecasting with Adaptive Factor Memory},
  author  = {Chang, Yurui and Du, Yongkang and Cao, Yuanpu and Chen, Jinghui and Lin, Lu},
  journal = {arXiv preprint arXiv:2605.30858},
  year    = {2026}
}

@inproceedings{chen2024datajuicer,
  title     = {{Data-Juicer}: A One-Stop Data Processing System for Large Language Models},
  author    = {Chen, Daoyuan and Huang, Yilun and Ma, Zhijian and Chen, Hesen and Pan, Xuchen and Ge, Ce and Gao, Dawei and Xie, Yuexiang and Liu, Zhaoyang and Gao, Jinyang and Li, Yaliang and Ding, Bolin and Zhou, Jingren},
  booktitle = {Companion of the 2024 International Conference on Management of Data (SIGMOD/PODS '24)},
  pages     = {120--134},
  publisher = {ACM},
  year      = {2024},
  doi       = {10.1145/3626246.3653385}
}

@article{gandhi2025boxinggym,
  title   = {{BoxingGym}: Benchmarking Progress in Automated Experimental Design and Model Discovery},
  author  = {Gandhi, Kanishk and Li, Michael Y. and Goodyear, Lyle and Bhatia, Agam and Li, Louise and Bhaskar, Aditi and Zaman, Mohammed and Goodman, Noah D.},
  journal = {arXiv preprint arXiv:2501.01540},
  year    = {2025},
  note    = {Also presented at the NeurIPS 2025 Workshop on Scaling Environments for Agents}
}

@article{gao2025selfevolving,
  title   = {A Survey of Self-Evolving Agents: What, When, How, and Where to Evolve on the Path to Artificial Super Intelligence},
  author  = {Gao, Huan-ang and Geng, Jiayi and Hua, Wenyue and Hu, Mengkang and Juan, Xinzhe and Liu, Hongzhang and Liu, Shilong and Qiu, Jiahao and Qi, Xuan and Ren, Qihan and Wu, Yiran and Wang, Hongru and Xiao, Han and Zhou, Yuhang and Zhang, Shaokun and Zhang, Jiayi and Xiang, Jinyu and Fang, Yixiong and Zhao, Qiwen and Liu, Dongrui and Qian, Cheng and Wang, Zhenhailong and Hu, Minda and Wang, Huazheng and Wu, Qingyun and Ji, Heng and Wang, Mengdi},
  journal = {Transactions on Machine Learning Research},
  year    = {2026},
  note    = {arXiv:2507.21046}
}

@article{gottweis2025coscientist,
  title   = {Accelerating scientific discovery with {Co-Scientist}},
  author  = {Gottweis, Juraj and Weng, Wei-Hung and Daryin, Alexander and Tu, Tao and Sirkovic, Petar and Myaskovsky, Artiom and Glowaty, Grzegorz and Weissenberger, Felix and Orlandi, Alessio and Popovici, Dan and Palepu, Anil and Rong, Keran and Tanno, Ryutaro and Saab, Khaled and Zhang, Fan and Blum, Jacob and Carroll, Andrew and Kulkarni, Kavita and Toma{\v{s}}ev, Nenad and Zverinski, Dina and Rendulic, Ivor and Vedadi, Elahe and Hasler, Florian and Rimanic, Luka and Boia, Marina and Budiselic, Ivan and Feinstein, Ben and Bellaiche, Mathias and Sheffer, Tom and Freyberg, Jan and Ratcliff, Jeremy and Bertolli, Ottavia and Chou, Katherine and Hassidim, Avinatan and Gokturk, Burak and Vahdat, Amin and Guan, Yuan and Dhillon, Vikram and Vaishnav, Eeshit Dhaval and Lee, Byron and Costa, Tiago R. D. and Penad{\'e}s, Jos{\'e} R. and Peltz, Gary and Matias, Yossi and Manyika, James and Hassabis, Demis and Xu, Yunhan and Kohli, Pushmeet and Pawlosky, Annalisa and Karthikesalingam, Alan and Natarajan, Vivek},
  journal = {Nature},
  volume  = {655},
  number  = {8122},
  pages   = {487--496},
  year    = {2026},
  doi     = {10.1038/s41586-026-10644-y},
  note    = {Preprint: arXiv:2502.18864 (2025), originally titled ``Towards an AI co-scientist''}
}

@article{gu2025webdreamer,
  title={Is Your {LLM} Secretly a World Model of the Internet? Model-Based Planning for Web Agents},
  author={Gu, Yu and Zhang, Kai and Ning, Yuting and Zheng, Boyuan and Gou, Boyu and Xue, Tianci and Chang, Cheng and Srivastava, Sanjari and Xie, Yanan and Qi, Peng and Sun, Huan and Su, Yu},
  journal={Transactions on Machine Learning Research},
  year={2025},
  note={arXiv:2411.06559}
}

@inproceedings{guo2024dsagent,
  title     = {{DS-Agent}: Automated Data Science by Empowering Large Language Models with Case-Based Reasoning},
  author    = {Guo, Siyuan and Deng, Cheng and Wen, Ying and Chen, Hechang and Chang, Yi and Wang, Jun},
  booktitle = {Proceedings of the 41st International Conference on Machine Learning},
  series    = {Proceedings of Machine Learning Research},
  volume    = {235},
  pages     = {16813--16848},
  publisher = {PMLR},
  year      = {2024}
}

@article{hoffmann2022chinchilla,
  title={Training Compute-Optimal Large Language Models},
  author={Hoffmann, Jordan and Borgeaud, Sebastian and Mensch, Arthur and Buchatskaya, Elena and Cai, Trevor and Rutherford, Eliza and de Las Casas, Diego and Hendricks, Lisa Anne and Welbl, Johannes and Clark, Aidan and Hennigan, Tom and Noland, Eric and Millican, Katie and van den Driessche, George and Damoc, Bogdan and Guy, Aurelia and Osindero, Simon and Simonyan, Karen and Elsen, Erich and Vinyals, Oriol and Rae, Jack W. and Sifre, Laurent},
  journal={arXiv preprint arXiv:2203.15556},
  year={2022}
}

@inproceedings{hu2025adas,
  title={Automated Design of Agentic Systems},
  author={Hu, Shengran and Lu, Cong and Clune, Jeff},
  booktitle={International Conference on Learning Representations (ICLR)},
  year={2025}
}

@inproceedings{huang2024mlagentbench,
  title     = {{MLAgentBench}: Evaluating Language Agents on Machine Learning Experimentation},
  author    = {Huang, Qian and Vora, Jian and Liang, Percy and Leskovec, Jure},
  booktitle = {Proceedings of the 41st International Conference on Machine Learning},
  series    = {Proceedings of Machine Learning Research},
  volume    = {235},
  pages     = {20271--20309},
  publisher = {PMLR},
  year      = {2024}
}

@inproceedings{huang2025popper,
  title     = {Automated Hypothesis Validation with Agentic Sequential Falsifications},
  author    = {Huang, Kexin and Jin, Ying and Li, Ryan and Li, Michael Y. and Cand{\`e}s, Emmanuel and Leskovec, Jure},
  booktitle = {Proceedings of the 42nd International Conference on Machine Learning},
  series    = {Proceedings of Machine Learning Research},
  volume    = {267},
  pages     = {25372--25437},
  publisher = {PMLR},
  year      = {2025}
}

@article{huang2026memorysurvey,
  title   = {A Survey of Agent Memory in the Second Half: Towards Self-Evolving and Long-Horizon Agents},
  author  = {Huang, Wei-Chieh and Zhang, Weizhi and Liang, Yueqing and Bei, Yuanchen and Chen, Yankai and Feng, Tao and Pan, Xinyu and Tan, Zhen and Wang, Yu and Wei, Tianxin and Wu, Shanglin and Xu, Ruiyao and Yang, Liangwei and Yang, Rui and Yang, Wooseong and Yeh, Chin-Yuan and Zhang, Hanrong and Zhang, Haozhen and Zhu, Siqi and Zou, Henry Peng and Zhao, Wanjia and Wang, Song and Xu, Wujiang and Ke, Zixuan and Hui, Zheng and Li, Dawei and Wu, Yaozu and He, Langzhou and Wang, Chen and Xu, Xiongxiao and Huang, Baixiang and Tan, Juntao and Heinecke, Shelby and Wang, Huan and Xiong, Caiming and Metwally, Ahmed A. and Yan, Jun and Lee, Chen-Yu and Zeng, Hanqing and Xia, Yinglong and Wei, Xiaokai and Payani, Ali and Wang, Yu and Ma, Haitong and Wang, Wenya and Wang, Chenguang and Zhang, Yu and Wang, Xin Eric and Zhang, Yongfeng and You, Jiaxuan and Tong, Hanghang and Luo, Xiao and Liu, Xue and Sun, Yizhou and Wang, Wei and McAuley, Julian and Zou, James and Han, Jiawei and Yu, Philip S. and Shu, Kai},
  journal = {Transactions on Machine Learning Research},
  year    = {2026},
  note    = {arXiv:2602.06052}
}

@inproceedings{jansen2024discoveryworld,
  title     = {{DiscoveryWorld}: A Virtual Environment for Developing and Evaluating Automated Scientific Discovery Agents},
  author    = {Jansen, Peter and C{\^o}t{\'e}, Marc-Alexandre and Khot, Tushar and Bransom, Erin and Dalvi Mishra, Bhavana and Majumder, Bodhisattwa Prasad and Tafjord, Oyvind and Clark, Peter},
  booktitle = {Advances in Neural Information Processing Systems},
  volume    = {37},
  pages     = {10088--10116},
  year      = {2024},
  publisher = {Curran Associates, Inc.},
  note      = {Datasets and Benchmarks Track},
  doi       = {10.52202/079017-0324}
}

@inproceedings{jansen2025codescientist,
  title     = {{CodeScientist}: End-to-End Semi-Automated Scientific Discovery with Code-based Experimentation},
  author    = {Jansen, Peter and Tafjord, Oyvind and Radensky, Marissa and Siangliulue, Pao and Hope, Tom and Dalvi Mishra, Bhavana and Majumder, Bodhisattwa Prasad and Weld, Daniel S. and Clark, Peter},
  booktitle = {Findings of the Association for Computational Linguistics: ACL 2025},
  year      = {2025},
  address   = {Vienna, Austria},
  publisher = {Association for Computational Linguistics},
  pages     = {13370--13467},
  doi       = {10.18653/v1/2025.findings-acl.692}
}

@article{jiang2025aide,
  title   = {{AIDE}: {AI}-Driven Exploration in the Space of Code},
  author  = {Jiang, Zhengyao and Schmidt, Dominik and Srikanth, Dhruv and Xu, Dixing and Kaplan, Ian and Jacenko, Deniss and Wu, Yuxiang},
  journal = {arXiv preprint arXiv:2502.13138},
  year    = {2025}
}

@article{kadavath2022know,
  title={Language Models (Mostly) Know What They Know},
  author={Kadavath, Saurav and Conerly, Tom and Askell, Amanda and Henighan, Tom and Drain, Dawn and Perez, Ethan and Schiefer, Nicholas and Hatfield-Dodds, Zac and DasSarma, Nova and Tran-Johnson, Eli and Johnston, Scott and El-Showk, Sheer and Jones, Andy and Elhage, Nelson and Hume, Tristan and Chen, Anna and Bai, Yuntao and Bowman, Sam and Fort, Stanislav and Ganguli, Deep and Hernandez, Danny and Jacobson, Josh and Kernion, Jackson and Kravec, Shauna and Lovitt, Liane and Ndousse, Kamal and Olsson, Catherine and Ringer, Sam and Amodei, Dario and Brown, Tom and Clark, Jack and Joseph, Nicholas and Mann, Ben and McCandlish, Sam and Olah, Chris and Kaplan, Jared},
  journal={arXiv preprint arXiv:2207.05221},
  year={2022}
}

@article{li2024autokaggle,
  title   = {{AutoKaggle}: A Multi-Agent Framework for Autonomous Data Science Competitions},
  author  = {Li, Ziming and Zang, Qianbo and Ma, David and Guo, Jiawei and Zheng, Tuney and Liu, Minghao and Niu, Xinyao and Wang, Yue and Yang, Jian and Liu, Jiaheng and Zhong, Wanjun and Zhou, Wangchunshu and Huang, Wenhao and Zhang, Ge},
  journal = {arXiv preprint arXiv:2410.20424},
  year    = {2024}
}

@article{li2025steplaw,
  title={Predictable Scale: {Part I}, {Step Law} -- Optimal Hyperparameter Scaling Law in Large Language Model Pretraining},
  author={Li, Houyi and Zheng, Wenzhen and Wang, Qiufeng and Zhang, Hanshan and Wang, Zili and Xuyang, Shijie and Fan, Yuantao and Ding, Zhenyu and Wang, Haoying and Ding, Ning and Zhou, Shuigeng and Zhang, Xiangyu and Jiang, Daxin},
  journal={arXiv preprint arXiv:2503.04715},
  year={2025}
}

@article{lidayan2025abbel,
  title   = {{ABBEL}: Learning Natural-Language Belief States for Memory-Efficient Interaction},
  author  = {Lidayan, Aly and Bjorner, Jakob and Golechha, Satvik and Goyal, Kartik and Suhr, Alane},
  journal = {arXiv preprint arXiv:2512.20111},
  year    = {2025}
}

@article{liu2025mlmaster,
  title   = {{ML-Master}: Towards {AI}-for-{AI} via Integration of Exploration and Reasoning},
  author  = {Liu, Zexi and Cai, Yuzhu and Zhu, Xinyu and Zheng, Yujie and Chen, Runkun and Wen, Ying and Wang, Yanfeng and E, Weinan and Chen, Siheng},
  journal = {arXiv preprint arXiv:2506.16499},
  year    = {2025}
}

@article{lu2024aiscientist,
  title   = {The {AI} {Scientist}: Towards Fully Automated Open-Ended Scientific Discovery},
  author  = {Lu, Chris and Lu, Cong and Lange, Robert Tjarko and Foerster, Jakob and Clune, Jeff and Ha, David},
  journal = {arXiv preprint arXiv:2408.06292},
  year    = {2024}
}

@article{luo2026dataagents,
  title   = {Data Agents: Levels, State of the Art, and Open Problems},
  author  = {Luo, Yuyu and Li, Guoliang and Fan, Ju and Tang, Nan},
  journal = {arXiv preprint arXiv:2602.04261},
  year    = {2026},
  note    = {SIGMOD 2026 Tutorial}
}

@inproceedings{ma2025autodata,
  title     = {{AutoData}: A Multi-Agent System for Open Web Data Collection},
  author    = {Ma, Tianyi and Qian, Yiyue and Zhang, Zheyuan and Wang, Zehong and Qian, Xiaoye and Bai, Feifan and Ding, Yifan and Luo, Xuwei and Zhang, Shinan and Murugesan, Keerthiram and Zhang, Chuxu and Ye, Yanfang},
  booktitle = {Advances in Neural Information Processing Systems (NeurIPS)},
  year      = {2025}
}

@inproceedings{mazumder2023dataperf,
  title     = {{DataPerf}: Benchmarks for Data-Centric {AI} Development},
  author    = {Mazumder, Mark and Banbury, Colby and Yao, Xiaozhe and Karla{\v{s}}, Bojan and Gaviria Rojas, William and Diamos, Sudnya and Diamos, Greg and He, Lynn and Parrish, Alicia and Kirk, Hannah Rose and Quaye, Jessica and Rastogi, Charvi and Kiela, Douwe and Jurado, David and Kanter, David and Mosquera, Rafael and Cukierski, Will and Ciro, Juan and Aroyo, Lora and Acun, Bilge and Chen, Lingjiao and Raje, Mehul and Bartolo, Max and Eyuboglu, Evan Sabri and Ghorbani, Amirata and Goodman, Emmett and Howard, Addison and Inel, Oana and Kane, Tariq and Kirkpatrick, Christine R. and Sculley, D. and Kuo, Tzu-Sheng and Mueller, Jonas W. and Thrush, Tristan and Vanschoren, Joaquin and Warren, Margaret and Williams, Adina and Yeung, Serena and Ardalani, Newsha and Paritosh, Praveen and Zhang, Ce and Zou, James and Wu, Carole-Jean and Coleman, Cody and Ng, Andrew and Mattson, Peter and Janapa Reddi, Vijay},
  booktitle = {Advances in Neural Information Processing Systems},
  volume    = {36},
  pages     = {5320--5347},
  publisher = {Curran Associates, Inc.},
  year      = {2023},
  doi       = {10.52202/075280-0235}
}

@article{mitchener2025kosmos,
  title   = {{Kosmos}: An {AI} Scientist for Autonomous Discovery},
  author  = {Mitchener, Ludovico and Yiu, Angela and Chang, Benjamin and Bourdenx, Mathieu and Nadolski, Tyler and Sulovari, Arvis and Landsness, Eric C. and Barabasi, Daniel L. and Narayanan, Siddharth and Evans, Nicky and Reddy, Shriya and Foiani, Martha and Kamal, Aizad and Shriver, Leah P. and Cao, Fang and Wassie, Asmamaw T. and Laurent, Jon M. and Melville-Green, Edwin and Caldas, Mayk and Bou, Albert and Roberts, Kaleigh F. and Zagorac, Sladjana and Orr, Timothy C. and Orr, Miranda E. and Zwezdaryk, Kevin J. and Ghareeb, Ali E. and McCoy, Laurie and Gomes, Bruna and Ashley, Euan A. and Duff, Karen E. and Buonassisi, Tonio and Rainforth, Tom and Bateman, Randall J. and Skarlinski, Michael and Rodriques, Samuel G. and Hinks, Michaela M. and White, Andrew D.},
  journal = {arXiv preprint arXiv:2511.02824},
  year    = {2025}
}

@article{novikov2025alphaevolve,
  title={{AlphaEvolve}: A coding agent for scientific and algorithmic discovery},
  author={Novikov, Alexander and V{\~u}, Ng{\^a}n and Eisenberger, Marvin and Dupont, Emilien and Huang, Po-Sen and Wagner, Adam Zsolt and Shirobokov, Sergey and Kozlovskii, Borislav and Ruiz, Francisco J. R. and Mehrabian, Abbas and Kumar, M. Pawan and See, Abigail and Chaudhuri, Swarat and Holland, George and Davies, Alex and Nowozin, Sebastian and Kohli, Pushmeet and Balog, Matej},
  journal={arXiv preprint arXiv:2506.13131},
  year={2025}
}

@inproceedings{ouyang2025reasoningbank,
  title     = {{ReasoningBank}: Scaling Agent Self-Evolving with Reasoning Memory},
  author    = {Ouyang, Siru and Yan, Jun and Hsu, I-Hung and Chen, Yanfei and Jiang, Ke and Wang, Zifeng and Han, Rujun and Le, Long T. and Daruki, Samira and Tang, Xiangru and Tirumalashetty, Vishy and Lee, George and Rofouei, Mahsan and Lin, Hangfei and Han, Jiawei and Lee, Chen-Yu and Pfister, Tomas},
  booktitle = {The Fourteenth International Conference on Learning Representations (ICLR)},
  year      = {2026},
  note      = {arXiv:2509.25140}
}

@inproceedings{penedo2024fineweb,
  title     = {The {FineWeb} Datasets: Decanting the Web for the Finest Text Data at Scale},
  author    = {Penedo, Guilherme and Kydl{\'i}{\v{c}}ek, Hynek and Ben Allal, Loubna and Lozhkov, Anton and Mitchell, Margaret and Raffel, Colin and Von Werra, Leandro and Wolf, Thomas},
  booktitle = {Advances in Neural Information Processing Systems},
  volume    = {37},
  pages     = {30811--30849},
  publisher = {Curran Associates, Inc.},
  year      = {2024},
  doi       = {10.52202/079017-0970}
}

@article{qi2024cleanagent,
  title   = {{CleanAgent}: Automating Data Standardization with {LLM}-based Agents},
  author  = {Qi, Danrui and Miao, Zhengjie and Wang, Jiannan},
  journal = {arXiv preprint arXiv:2403.08291},
  year    = {2024},
  note    = {Also appeared in the VLDB 2025 Workshop on Data Driven AI (DATAI)}
}

@article{ren2026selfimprove,
  title   = {Self-Improvements in Modern Agentic Systems: A Survey},
  author  = {Ren, Zhe and Chen, Yimeng and Guo, Dandan and Rong, Guowei and Li, Tonghui and Xiong, R. B. and Lan, Qingfeng and Wang, Wenyi and Nanbo, Li and Yang, Yibo and Zhuge, Mingchen and Schmidhuber, J{\"u}rgen},
  journal = {arXiv preprint arXiv:2607.13104},
  year    = {2026}
}

@article{riosgarcia2026aiscientists,
  title   = {{AI} Scientists Produce Results without Reasoning Scientifically},
  author  = {R{\'i}os-Garc{\'i}a, Marti{\~n}o and Alampara, Nawaf and Gupta, Chandan and Mandal, Indrajeet and Mannan, Sajid and Aghajani, Ali Asghar and Krishnan, N. M. Anoop and Jablonka, Kevin Maik},
  journal = {arXiv preprint arXiv:2604.18805},
  year    = {2026}
}

@inproceedings{schmidgall2025agentlab,
  title     = {Agent Laboratory: Using {LLM} Agents as Research Assistants},
  author    = {Schmidgall, Samuel and Su, Yusheng and Wang, Ze and Sun, Ximeng and Wu, Jialian and Yu, Xiaodong and Liu, Jiang and Moor, Michael and Liu, Zicheng and Barsoum, Emad},
  booktitle = {Findings of the Association for Computational Linguistics: EMNLP 2025},
  year      = {2025},
  address   = {Suzhou, China},
  publisher = {Association for Computational Linguistics},
  pages     = {5977--6043},
  doi       = {10.18653/v1/2025.findings-emnlp.320}
}

@article{seedat2025datasavvy,
  title   = {What's the next frontier for Data-centric {AI}? {D}ata Savvy Agents},
  author  = {Seedat, Nabeel and Liu, Jiashuo and van der Schaar, Mihaela},
  journal = {arXiv preprint arXiv:2511.01015},
  year    = {2025}
}

@inproceedings{snoek2012bayesopt,
  title={Practical {Bayesian} Optimization of Machine Learning Algorithms},
  author={Snoek, Jasper and Larochelle, Hugo and Adams, Ryan P.},
  booktitle={Advances in Neural Information Processing Systems},
  volume={25},
  year={2012}
}

@article{sutton1991dyna,
  title={{Dyna}, an Integrated Architecture for Learning, Planning, and Reacting},
  author={Sutton, Richard S.},
  journal={ACM SIGART Bulletin},
  volume={2},
  number={4},
  pages={160--163},
  year={1991},
  doi={10.1145/122344.122377}
}

@article{takahara2026hep,
  title   = {Toward Auditable {AI} Scientists: A Hypothesis Evolution Protocol for {LLM} Agents},
  author  = {Takahara, Izumi and Mizoguchi, Teruyasu},
  journal = {arXiv preprint arXiv:2607.09195},
  year    = {2026}
}

@inproceedings{tian2023calibration,
  title={Just Ask for Calibration: Strategies for Eliciting Calibrated Confidence Scores from Language Models Fine-Tuned with Human Feedback},
  author={Tian, Katherine and Mitchell, Eric and Zhou, Allan and Sharma, Archit and Rafailov, Rafael and Yao, Huaxiu and Finn, Chelsea and Manning, Christopher D.},
  booktitle={Proceedings of the 2023 Conference on Empirical Methods in Natural Language Processing},
  pages={5433--5442},
  publisher={Association for Computational Linguistics},
  year={2023}
}

@article{wainrib2026labloop,
  title   = {Can {AI} Scientist Agents Learn from Lab-in-the-Loop Feedback? {E}vidence from Iterative Perturbation Discovery},
  author  = {Wainrib, Gilles and Bodinier, Barbara and Dakhli, Haitem and Monserrat, Josep and Espin Perez, Almudena and Carpentier, Sabrina and Codato, Roberta and Klein, John},
  journal = {arXiv preprint arXiv:2603.26177},
  year    = {2026}
}

@article{wang2023voyager,
  title={{Voyager}: An Open-Ended Embodied Agent with Large Language Models},
  author={Wang, Guanzhi and Xie, Yuqi and Jiang, Yunfan and Mandlekar, Ajay and Xiao, Chaowei and Zhu, Yuke and Fan, Linxi and Anandkumar, Anima},
  journal={Transactions on Machine Learning Research},
  year={2024},
  note={arXiv:2305.16291}
}

@inproceedings{wang2024awm,
  title={Agent Workflow Memory},
  author={Wang, Zora Zhiruo and Mao, Jiayuan and Fried, Daniel and Neubig, Graham},
  booktitle={Proceedings of the 42nd International Conference on Machine Learning},
  series={Proceedings of Machine Learning Research},
  volume={267},
  pages={63897--63911},
  publisher={PMLR},
  year={2025}
}

@inproceedings{wijk2025rebench,
  title     = {{RE-Bench}: Evaluating Frontier {AI} {R\&D} Capabilities of Language Model Agents against Human Experts},
  author    = {Wijk, Hjalmar and Lin, Tao Roa and Becker, Joel and Jawhar, Sami and Parikh, Neev and Broadley, Thomas and Chan, Lawrence and Chen, Michael and Clymer, Joshua M. and Dhyani, Jai and Ericheva, Elena and Garcia, Katharyn and Goodrich, Brian and Jurkovic, Nikola and Kinniment, Megan and Lajko, Aron and Nix, Seraphina and Sato, Lucas Jun Koba and Saunders, William and Taran, Maksym and West, Ben and Barnes, Elizabeth},
  booktitle = {Proceedings of the 42nd International Conference on Machine Learning},
  series    = {Proceedings of Machine Learning Research},
  volume    = {267},
  pages     = {66772--66832},
  publisher = {PMLR},
  year      = {2025}
}

@inproceedings{xie2023dsir,
  title     = {Data Selection for Language Models via Importance Resampling},
  author    = {Xie, Sang Michael and Santurkar, Shibani and Ma, Tengyu and Liang, Percy},
  booktitle = {Advances in Neural Information Processing Systems},
  volume    = {36},
  pages     = {34201--34227},
  publisher = {Curran Associates, Inc.},
  year      = {2023},
  doi       = {10.52202/075280-1482}
}

@inproceedings{xiong2024uncertainty,
  title={Can {LLMs} Express Their Uncertainty? An Empirical Evaluation of Confidence Elicitation in {LLMs}},
  author={Xiong, Miao and Hu, Zhiyuan and Lu, Xinyang and Li, Yifei and Fu, Jie and He, Junxian and Hooi, Bryan},
  booktitle={The Twelfth International Conference on Learning Representations},
  year={2024}
}

@inproceedings{xu2025amem,
  title     = {{A-MEM}: Agentic Memory for {LLM} Agents},
  author    = {Xu, Wujiang and Liang, Zujie and Mei, Kai and Gao, Hang and Tan, Juntao and Zhang, Yongfeng},
  booktitle = {Advances in Neural Information Processing Systems},
  volume    = {38},
  year      = {2025},
  note      = {arXiv:2502.12110}
}

@article{yamada2025aiscientist2,
  title   = {The {AI} {Scientist}-v2: Workshop-Level Automated Scientific Discovery via Agentic Tree Search},
  author  = {Yamada, Yutaro and Lange, Robert Tjarko and Lu, Cong and Hu, Shengran and Lu, Chris and Foerster, Jakob and Clune, Jeff and Ha, David},
  journal = {arXiv preprint arXiv:2504.08066},
  year    = {2025}
}

@inproceedings{yang2024opro,
  title={Large Language Models as Optimizers},
  author={Yang, Chengrun and Wang, Xuezhi and Lu, Yifeng and Liu, Hanxiao and Le, Quoc V. and Zhou, Denny and Chen, Xinyun},
  booktitle={International Conference on Learning Representations (ICLR)},
  year={2024}
}

@article{yang2025rdagent,
  title   = {{R\&D-Agent}: An {LLM}-Agent Framework Towards Autonomous Data Science},
  author  = {Yang, Xu and Yang, Xiao and Fang, Shikai and Zhang, Yifei and Wang, Jian and Xian, Bowen and Li, Qizheng and Li, Jingyuan and Xu, Minrui and Li, Yuante and Pan, Haoran and Zhang, Yuge and Liu, Weiqing and Shen, Yelong and Chen, Weizhu and Bian, Jiang},
  journal = {arXiv preprint arXiv:2505.14738},
  year    = {2025}
}

@inproceedings{ye2025mixinglaws,
  title     = {Data Mixing Laws: Optimizing Data Mixtures by Predicting Language Modeling Performance},
  author    = {Ye, Jiasheng and Liu, Peiju and Sun, Tianxiang and Zhan, Jun and Zhou, Yunhua and Qiu, Xipeng},
  booktitle = {The Thirteenth International Conference on Learning Representations (ICLR)},
  year      = {2025}
}

@inproceedings{yu2025dynamind,
  title={{Dyna-Mind}: Learning to Simulate from Experience for Better {AI} Agents},
  author={Yu, Xiao and Peng, Baolin and Galley, Michel and Cheng, Hao and Wu, Qianhui and Kulkarni, Janardhan and Nath, Suman and Yu, Zhou and Gao, Jianfeng},
  booktitle={International Conference on Learning Representations (ICLR)},
  year={2026},
  note={arXiv:2510.09577}
}

@article{yu2025dynathink,
  title={{Dyna-Think}: Synergizing Reasoning, Acting, and World Model Simulation in {AI} Agents},
  author={Yu, Xiao and Peng, Baolin and Xu, Ruize and Galley, Michel and Cheng, Hao and Nath, Suman and Gao, Jianfeng and Yu, Zhou},
  journal={arXiv preprint arXiv:2506.00320},
  year={2025}
}

@article{yuksekgonul2025textgrad,
  title={Optimizing generative {AI} by backpropagating language model feedback},
  author={Yuksekgonul, Mert and Bianchi, Federico and Boen, Joseph and Liu, Sheng and Lu, Pan and Huang, Zhi and Guestrin, Carlos and Zou, James},
  journal={Nature},
  volume={639},
  number={8055},
  pages={609--616},
  year={2025},
  doi={10.1038/s41586-025-08661-4}
}

@inproceedings{zhang2025dgm,
  title={{Darwin} {G{\"o}del} Machine: Open-Ended Evolution of Self-Improving Agents},
  author={Zhang, Jenny and Hu, Shengran and Lu, Cong and Lange, Robert Tjarko and Clune, Jeff},
  booktitle={International Conference on Learning Representations (ICLR)},
  year={2026},
  note={arXiv:2505.22954}
}

@article{lin2023evolutionary,
  author  = {Lin, Zeming and Akin, Halil and Rao, Roshan and Hie, Brian and Zhu, Zhongkai and Lu, Wenting and Smetanin, Nikita and Verkuil, Robert and Kabeli, Ori and Shmueli, Yaniv and dos Santos Costa, Allan and Fazel-Zarandi, Maryam and Sercu, Tom and Candido, Salvatore and Rives, Alexander},
  title   = {Evolutionary-scale prediction of atomic-level protein structure with a language model},
  journal = {Science},
  volume  = {379},
  number  = {6637},
  pages   = {1123--1130},
  year    = {2023},
  doi     = {10.1126/science.ade2574}
}

@article{jarvelin2002cumulated,
  author  = {J{\"a}rvelin, Kalervo and Kek{\"a}l{\"a}inen, Jaana},
  title   = {Cumulated gain-based evaluation of {IR} techniques},
  journal = {ACM Transactions on Information Systems},
  volume  = {20},
  number  = {4},
  pages   = {422--446},
  year    = {2002},
  doi     = {10.1145/582415.582418}
}

@article{matthews1975comparison,
  author  = {Matthews, B. W.},
  title   = {Comparison of the predicted and observed secondary structure of {T4} phage lysozyme},
  journal = {Biochimica et Biophysica Acta (BBA) -- Protein Structure},
  volume  = {405},
  number  = {2},
  pages   = {442--451},
  year    = {1975},
  doi     = {10.1016/0005-2795(75)90109-9}
}

@article{gordon2020lentimpra,
  author  = {Gordon, M. Grace and Inoue, Fumitaka and Martin, Beth and Schubach, Max and Agarwal, Vikram and Whalen, Sean and Feng, Shiyun and Zhao, Jingjing and Ashuach, Tal and Ziffra, Ryan and Kreimer, Anat and Georgakopoulos-Soares, Ilias and Yosef, Nir and Ye, Chun Jimmie and Pollard, Katherine S. and Shendure, Jay and Kircher, Martin and Ahituv, Nadav},
  title   = {{lentiMPRA} and {MPRAflow} for high-throughput functional characterization of gene regulatory elements},
  journal = {Nature Protocols},
  volume  = {15},
  number  = {8},
  pages   = {2387--2412},
  year    = {2020},
  doi     = {10.1038/s41596-020-0333-5}
}

@article{igvf2024deciphering,
  author  = {{IGVF Consortium}},
  title   = {Deciphering the impact of genomic variation on function},
  journal = {Nature},
  volume  = {633},
  number  = {8028},
  pages   = {47--57},
  year    = {2024},
  doi     = {10.1038/s41586-024-07510-0}
}

@article{ji2021dnabert,
  author  = {Ji, Yanrong and Zhou, Zhihan and Liu, Han and Davuluri, Ramana V.},
  title   = {{DNABERT}: pre-trained {Bidirectional Encoder Representations from Transformers} model for {DNA}-language in genome},
  journal = {Bioinformatics},
  volume  = {37},
  number  = {15},
  pages   = {2112--2120},
  year    = {2021},
  doi     = {10.1093/bioinformatics/btab083}
}

@article{brier1950verification,
  author  = {Brier, Glenn W.},
  title   = {Verification of forecasts expressed in terms of probability},
  journal = {Monthly Weather Review},
  volume  = {78},
  number  = {1},
  pages   = {1--3},
  year    = {1950},
  doi     = {10.1175/1520-0493(1950)078<0001:VOFEIT>2.0.CO;2}
}

@incollection{platt2000probabilities,
  author    = {Platt, John C.},
  title     = {Probabilities for {SV} Machines},
  booktitle = {Advances in Large-Margin Classifiers},
  editor    = {Smola, Alexander J. and Bartlett, Peter and Sch{\"o}lkopf, Bernhard and Schuurmans, Dale},
  publisher = {The MIT Press},
  pages     = {61--74},
  year      = {2000},
  isbn      = {9780262283977},
  doi       = {10.7551/mitpress/1113.003.0008}
}

@article{ma2020actinn,
  author  = {Ma, Feiyang and Pellegrini, Matteo},
  title   = {{ACTINN}: automated identification of cell types in single cell {RNA} sequencing},
  journal = {Bioinformatics},
  volume  = {36},
  number  = {2},
  pages   = {533--538},
  year    = {2020},
  doi     = {10.1093/bioinformatics/btz592}
}

@article{xu2021probabilistic,
  author  = {Xu, Chenling and Lopez, Romain and Mehlman, Edouard and Regier, Jeffrey and Jordan, Michael I. and Yosef, Nir},
  title   = {Probabilistic harmonization and annotation of single-cell transcriptomics data with deep generative models},
  journal = {Molecular Systems Biology},
  volume  = {17},
  number  = {1},
  pages   = {e9620},
  year    = {2021},
  doi     = {10.15252/msb.20209620}
}

@article{lopez2018deep,
  author  = {Lopez, Romain and Regier, Jeffrey and Cole, Michael B. and Jordan, Michael I. and Yosef, Nir},
  title   = {Deep generative modeling for single-cell transcriptomics},
  journal = {Nature Methods},
  volume  = {15},
  number  = {12},
  pages   = {1053--1058},
  year    = {2018},
  doi     = {10.1038/s41592-018-0229-2}
}

@article{gretton2012kernel,
  author  = {Gretton, Arthur and Borgwardt, Karsten M. and Rasch, Malte J. and Sch{\"o}lkopf, Bernhard and Smola, Alexander},
  title   = {A Kernel Two-Sample Test},
  journal = {Journal of Machine Learning Research},
  volume  = {13},
  number  = {25},
  pages   = {723--773},
  year    = {2012},
  url     = {http://jmlr.org/papers/v13/gretton12a.html}
}

@article{amodio2019exploring,
  author  = {Amodio, Matthew and van Dijk, David and Srinivasan, Krishnan and Chen, William S. and Mohsen, Hussein and Moon, Kevin R. and Campbell, Allison and Zhao, Yujiao and Wang, Xiaomei and Venkataswamy, Manjunatha and Desai, Anita and Ravi, V. and Kumar, Priti and Montgomery, Ruth and Wolf, Guy and Krishnaswamy, Smita},
  title   = {Exploring single-cell data with deep multitasking neural networks},
  journal = {Nature Methods},
  volume  = {16},
  number  = {11},
  pages   = {1139--1145},
  year    = {2019},
  doi     = {10.1038/s41592-019-0576-7}
}

@article{roohani2024predicting,
  author  = {Roohani, Yusuf and Huang, Kexin and Leskovec, Jure},
  title   = {Predicting transcriptional outcomes of novel multigene perturbations with {GEARS}},
  journal = {Nature Biotechnology},
  volume  = {42},
  number  = {6},
  pages   = {927--935},
  year    = {2024},
  doi     = {10.1038/s41587-023-01905-6}
}

@article{ashburner2000gene,
  author  = {Ashburner, Michael and Ball, Catherine A. and Blake, Judith A. and Botstein, David and Butler, Heather and Cherry, J. Michael and Davis, Allan P. and Dolinski, Kara and Dwight, Selina S. and Eppig, Janan T. and Harris, Midori A. and Hill, David P. and Issel-Tarver, Laurie and Kasarskis, Andrew and Lewis, Suzanna and Matese, John C. and Richardson, Joel E. and Ringwald, Martin and Rubin, Gerald M. and Sherlock, Gavin},
  title   = {{Gene Ontology}: tool for the unification of biology},
  journal = {Nature Genetics},
  volume  = {25},
  number  = {1},
  pages   = {25--29},
  year    = {2000},
  doi     = {10.1038/75556}
}

@inproceedings{kingma2015adam,
  author    = {Kingma, Diederik P. and Ba, Jimmy},
  title     = {{Adam}: A Method for Stochastic Optimization},
  booktitle = {International Conference on Learning Representations},
  year      = {2015},
  eprint    = {1412.6980},
  archivePrefix = {arXiv},
  url       = {https://arxiv.org/abs/1412.6980}
}

@inproceedings{paszke2019pytorch,
  author    = {Paszke, Adam and Gross, Sam and Massa, Francisco and Lerer, Adam and Bradbury, James and Chanan, Gregory and Killeen, Trevor and Lin, Zeming and Gimelshein, Natalia and Antiga, Luca and Desmaison, Alban and Kopf, Andreas and Yang, Edward and DeVito, Zachary and Raison, Martin and Tejani, Alykhan and Chilamkurthy, Sasank and Steiner, Benoit and Fang, Lu and Bai, Junjie and Chintala, Soumith},
  title     = {{PyTorch}: An Imperative Style, High-Performance Deep Learning Library},
  booktitle = {Advances in Neural Information Processing Systems},
  editor    = {Wallach, H. and Larochelle, H. and Beygelzimer, A. and d'Alch{\'e}-Buc, F. and Fox, E. and Garnett, R.},
  volume    = {32},
  publisher = {Curran Associates, Inc.},
  year      = {2019},
  url       = {https://proceedings.neurips.cc/paper_files/paper/2019/file/bdbca288fee7f92f2bfa9f7012727740-Paper.pdf}
}
\bibliographystyle{iclr2027_conference}

\clearpage
\appendix

\clearpage

\appendix

\onecolumn
\begin{appendices}
\startcontents[appendices]
\printcontents[appendices]{}{1}{\setcounter{tocdepth}{3}}

\clearpage

\section{Extended Related Work}
\label{sec:extended_related}

\paragraph{Data selection, mixing, and biomedical pipelines.}
Example-level methods select training data using deduplication~\citep{lee2022dedup}, importance resampling toward a target distribution~\citep{xie2023dsir}, gradient-based influence for instruction or vision-language tuning~\citep{xia2024less,wu2024icons}, representation-based scores~\citep{ivison2025large}, or pruning metrics motivated by data scaling~\citep{sorscher2022beyond}. Mixture methods estimate how performance responds to source proportions, sometimes from smaller proxy runs and sometimes during training~\citep{xie2023doremi,ye2025mixinglaws,liu2025regmix,chen2025aioli,chen2026olmix}. Filtering decisions also interact with available compute: repeating a small high-quality pool can change the relative value of adding less-filtered data~\citep{muennighoff2023scaling,goyal2024scaling}. Benchmarks and integrated pipelines support systematic comparison and deployment of curation methods~\citep{mazumder2023dataperf,gadre2023datacomp,penedo2024fineweb,chen2024datajuicer}; see \citet{albalak2024survey} for a survey.

In computational biology, quality control, normalization, and harmonization are essential to comparable analyses~\citep{luecken2022scib,heumos2023practices,notin2023proteingym,peidli2024scperturb}. \citet{vieth2019pipelines} evaluate thousands of single-cell analysis pipelines, while pipeComp~\citep{germain2020pipecomp} studies preprocessing choices and their interactions. These studies concern analysis pipelines; our controlled recipe grid instead evaluates how training-subset choices transfer across datasets and downstream models while holding subset size fixed. It shows that the tested standard recipes average no better than random subsets.

\paragraph{Agents for machine-learning research and data work.}
Coding agents are evaluated on machine-learning competitions, research-engineering tasks, and language-model post-training~\citep{huang2024mlagentbench,chan2024mlebench,wijk2025rebench,rank2026posttrainbench}. Scaffolds structure their search as trees or populations~\citep{jiang2025aide,liu2025mlmaster,yang2025rdagent,li2024autokaggle}, reuse expert cases~\citep{guo2024dsagent}, or iteratively modify and evaluate training scripts~\citep{karpathy2026autoresearch,cai2026mlautoresearch}. Agents also support broader scientific workflows~\citep{lu2024aiscientist,yamada2025aiscientist2,schmidgall2025agentlab,jansen2025codescientist,gottweis2025coscientist,mitchener2025kosmos} and are tested on experimental design and model revision~\citep{jansen2024discoveryworld,gandhi2025boxinggym,wainrib2026labloop}. On the data side, work addresses dataset-issue detection, table cleaning, data collection, and general data-science tasks~\citep{huang2025dca,qi2024cleanagent,ma2025autodata,zhang2026datascibench,seedat2025datasavvy,luo2026dataagents}. Curation-Bench studies policy execution under a fixed training interface~\citep{kang2026curationbench}; DataMaster combines data discovery and refinement with tree-structured search and shared memory~\citep{du2026datamaster}. Our focus is how an agent interprets its own curation experiments over a long trajectory.

\paragraph{Memory, belief states, and scientific hypotheses.}
Agents preserve experience as verbal reflections~\citep{shinn2023reflexion}, distilled workflows~\citep{zhao2024expel,wang2024awm}, reusable skills~\citep{wang2023voyager}, evolving playbooks~\citep{suzgun2026dynamic,zhang2025ace}, or structured reasoning memories~\citep{xu2025amem,ouyang2025reasoningbank}. Surveys examine memory's role in longer-horizon agent learning~\citep{gao2025selfevolving,huang2026memorysurvey,ren2026selfimprove}. ABBEL makes a natural-language belief state the agent's compact representation of earlier observations and studies errors in updating that state~\citep{lidayan2025abbel}. In scientific discovery, HEP maintains evidence-linked hypotheses and belief probabilities~\citep{takahara2026hep}, while POPPER designs sequential falsification tests for measurable implications of a hypothesis~\citep{huang2025popper}. These works make belief maintenance and evidential support explicit; they also make clear why simply recording an agent's conclusions does not establish that those conclusions apply in a new setting. Behavioral evaluations of scientific agents likewise report failures to use evidence and revise beliefs consistently~\citep{riosgarcia2026aiscientists}.

ForecastCompass provides a particularly relevant comparison: it uses resolved forecasts and retrospective analysis to revise reusable factor and reasoning memories for future forecasting tasks~\citep{chang2026forecastcompass}. \alg operates in a different decision loop. Its forecast concerns the validation score of the curation policy the agent is about to execute; the subsequent error tests the policy-conditioned claims used to propose that experiment. Work on language-model calibration provides further motivation to check stated judgments against outcomes~\citep{kadavath2022know,tian2023calibration,xiong2024uncertainty}. Here, ``calibrated memory'' refers to this prospective check and claim revision; it does not imply a guarantee of calibrated predictive probabilities or formal hypothesis-testing error control.

\paragraph{Harness optimization and model-based prediction.}
Prompt and harness optimizers improve agent behavior across tasks through textual gradients, reflective evolution, architecture search, or self-modification~\citep{khattab2024dspy,yang2024opro,yuksekgonul2025textgrad,agrawal2025gepa,hu2025adas,zhang2025aflow,zhang2025dgm,novikov2025alphaevolve}. \alg holds its scaffold fixed and updates its understanding of one downstream curation task during a run. Model-based agents instead simulate the consequences of candidate actions before acting~\citep{sutton1991dyna,chae2025wma,gu2025webdreamer,yu2025dynathink,yu2025dynamind}. Empirical scaling laws and Bayesian optimization likewise use earlier measurements to predict expensive training outcomes~\citep{kaplan2020scaling,hoffmann2022chinchilla,li2025steplaw,bjorck2024lr,snoek2012bayesopt}. \alg also predicts an outcome, but uses the discrepancy after execution to update the written, conditional claims that determine where the agent searches next.

\section{Experimental Details}
\label{app:setup}
The \textbf{science experiments} cover three biological domains, each with its own data type: 
protein sequence and functional data, genomics, and single-cell transcriptomics. We tested five common tasks in these domains, including protein variant effect prediction~\citep{notin2023proteingym}, non-coding variant effect prediction~\citep{marin2024bend}, cell type annotation~\citep{abdelaal2019comparison}, batch integration~\citep{luecken2022scib} and perturbation response prediction~\citep{wu2025perturbench}. Each task combines several datasets with several models, giving \textbf{55 dataset--model tasks} in total. In each task, the agent selects 50\% of the pool. The \textbf{LLM post-training} experiments include \textit{general instruction tuning}, selecting 10{,}000 rows from the T\"ulu~3 \citep{lambert2024tulu} or LESS mixture ~\citep{xia2024less} to fine-tune OLMo-2-1B or Llama-2-7B, and \textit{code instruction tuning}, selecting 10{,}000 rows from OpenCodeInstruct \citep{ahmad2025opencodeinstruct} to fine-tune Qwen2.5-Coder-1.5B. 

Every task is run under two agentic pipelines, \textit{Basic agent} and \alg . We test three agents with both agentic pipelines: Claude Code driven by Opus~4.8, Sonnet~5, or Haiku~4.5; Codex driven by GPT-5.3-Codex or GPT-5.6~Sol; and OpenHands driven by Qwen3.5-397B-A17B, an open-weight model. The agents are initiated from the same-seeded random subset, counted as the first iteration, and run 15 iterations in biology and 10 in the LLM post-training track. 

Each task is scored by several metrics. For biological tasks, we report the task's primary metric from major benchmarks, averaged over the dataset--model pairs for LLM post-training, we report the mean score over each model's fixed benchmark suite. OLMo-2-1B is scored on MMLU~\citep{hendrycks2020measuring}, GSM8K~\citep{cobbe2021training}, BBH~\citep{suzgun2023challenging}, TruthfulQA~\citep{lin2022truthfulqa}, and IFEval~\citep{zhou2023instruction}. Llama-2-7B is scored on MMLU and BBH. Qwen2.5-Coder-1.5B-Instruct is scored on HumanEval+ and MBPP+~\citep{liu2023your}. The number reported is the best score reached within a run. Datasets, models, and metrics are detailed in Appendix~\ref{app:domains}.

\subsection{Domains, datasets, models, and metrics}
\label{app:domains}

Table~\ref{tab:setup-grid} summarizes the seven tasks, five in biology and two in instruction tuning. Each task is then described in three parts: the datasets and how their held-out splits were made, the models, and the metrics.

\begin{table}[htbp]
\centering
\caption{The benchmark grid, pairing each dataset with the models listed for it. Pool is the number of training rows from which the agent selects, and Budget is the number of rows that each submission must contain. Iter.\ is the number of iterations per run.}
\label{tab:setup-grid}
\resizebox{\ifdim\width>\linewidth\linewidth\else\width\fi}{!}{%
\begin{tabular}{llrrlll}
\toprule
Domain & Dataset & Pool & Budget & Models & Headline metric & Iter. \\
\midrule
\multirow{4}{*}{Protein variant effect} & blat\_ecolx & 2,894 & 1,447 & \multirow{4}{*}{ESM-2 ridge, ESM-2 MLP, AA-property ridge} & \multirow{4}{*}{Spearman} & \multirow{4}{*}{15} \\
 & cp2c9\_human & 3,736 & 1,868 & & & \\
 & tpmt\_human & 2,188 & 1,094 & & & \\
 & pten\_human & 4,387 & 2,194 & & & \\
\midrule
\multirow{3}{*}{Non-coding variant effect} & clinvar\_splice & 11,163 & 5,582 & \multirow{3}{*}{DNABERT logistic, 4-mer logistic} & \multirow{3}{*}{AUPRC} & \multirow{3}{*}{15} \\
 & mpra\_hepg2 & 10,978 & 5,489 & & & \\
 & eqtl\_grasp & 11,431 & 5,716 & & & \\
\midrule
\multirow{3}{*}{Cell-type annotation} & pancreas\_crosstech & 8,549 & 4,274 & \multirow{3}{*}{SVM, logistic, ACTINN, SSVAE, majority} & \multirow{3}{*}{Macro-F1} & \multirow{3}{*}{15} \\
 & pbmc\_crossprotocol & 8,165 & 4,082 & & & \\
 & pbmc\_open\_set & 4,263 & 2,131 & & & \\
\midrule
\multirow{3}{*}{Batch integration} & immune & 23,454 & 11,727 & \multirow{3}{*}{PCA, ComBat, Harmony, cVAE, MMD-AE} & \multirow{3}{*}{scIB-style overall score} & \multirow{3}{*}{15} \\
 & lung & 22,732 & 11,366 & & & \\
 & pancreas & 11,472 & 5,736 & & & \\
\midrule
\multirow{3}{*}{Perturbation response} & norman19 & 62,618 & 31,309 & CPA, latent additive, SAMS-VAE, GEARS & \multirow{2}{*}{Cosine of logFC} & \multirow{2}{*}{15} \\
 & frangieh21 & 151,480 & 75,740 & CPA, latent additive, SAMS-VAE & & \\

\midrule
\multirow{2}{*}{Instruction tuning} & T\"ulu 3 SFT mixture & 939,343 & 10,000 & OLMo-2-1B & Mean of 5 benchmarks & \multirow{2}{*}{10} \\
 & LESS mixture & 270,679 & 10,000 & Llama-2-7B & Mean of MMLU, BBH & \\
\midrule
Code instruction tuning & OpenCodeInstruct & 5,000,000 & 10,000 & Qwen2.5-Coder-1.5B-Instruct & Mean of HumanEval+, MBPP & 10 \\
\bottomrule
\end{tabular}}
\end{table}

\subsubsection{Protein variant effect prediction}
\label{app:dom-proteinvar}

Four deep mutational scanning assays are crossed with three frozen regressors, giving 12 cells. Each row is a single amino-acid substitution, and its assay score is the regression target.

\paragraph{Datasets.}
\begin{itemize}
  \item \texttt{blat\_ecolx}: ampicillin resistance of TEM-1 $\beta$-lactamase.
  \item \texttt{cp2c9\_human}: enzymatic activity of CYP2C9, read out by labeling with an activity-based probe.
  \item \texttt{tpmt\_human}: steady-state abundance of TPMT, measured by VAMP-seq.
  \item \texttt{pten\_human}: lipid-phosphatase activity of PTEN.
  \item Source and split: all four assays are taken unchanged from ProteinGym \citep{notin2023proteingym}, together with its position-based cross-validation folds. Three folds form the training pool, one is the evaluation split, and one is a test split. No residue position occurs in both the training pool and the evaluation split, so the task is transfer to positions that carry no labeled variant.
  \item Composition floors: a submission must retain at least 30 distinct positions and at least 10 distinct substituted amino acids.
\end{itemize}

\paragraph{Models.}
Two of the regressors share one frozen protein language model representation and differ only in the prediction head. The third replaces that representation with hand-crafted substitution features. All hyperparameters are fixed in advance, none is tuned per dataset, and training uses a fixed seed.
\begin{itemize}
  \item ESM-2 ridge: ridge regression on mean-pooled final-layer embeddings of the mutated sequence from a frozen 35M-parameter ESM-2 \citep{lin2023evolutionary}.
  \item ESM-2 MLP: a multilayer perceptron with two hidden layers on the same embeddings.
  \item AA-property ridge: ridge regression on 57 substitution features, namely one-hot wild-type and mutant residues, BLOSUM62 exchangeability \citep{henikoff1992blosum}, five physicochemical scales as wild-type, mutant, and difference values, and the relative position of the substituted residue along the chain.
\end{itemize}

\paragraph{Metrics.}
All metrics are computed on the evaluation split.
\begin{itemize}
  \item Spearman rank correlation between predicted and measured assay scores (headline).
  \item NDCG \citep{jarvelin2002cumulated}, with the assay score shifted to be non-negative as the gain, and recall of the measured top decile within the predicted top decile.
  \item ROC-AUC and the Matthews correlation coefficient \citep{matthews1975comparison} against ProteinGym's per-assay binarization. The Matthews correlation coefficient is computed on rank-based calls that label the top half of the predictions positive.
  \item Pearson correlation and RMSE.
  \item Guard: the number of evaluation positions present in the curated training set, which must be zero. A non-zero count sets the tracked score to zero.
  \item Validity flag: false when the predictions are constant or non-finite. A false flag sets the tracked score to zero.
\end{itemize}

\subsubsection{Non-coding variant effect prediction}
\label{app:dom-ncvar}

Three datasets, each based on a different type of evidence that a non-coding variant is functional, are crossed with two frozen classifiers, giving six cells. Each row is a single-nucleotide substitution on GRCh38 with a binary label, represented by the 501-bp window of the reference genome centered on the variant, with the alternative allele substituted at the central position.

\paragraph{Datasets.}
\begin{itemize}
  \item \texttt{clinvar\_splice}: variants from the non-coding ClinVar set distributed with BEND \citep{marin2024bend}, restricted to splice-proximal consequences. Positives are classified in ClinVar as pathogenic or likely pathogenic, and negatives as benign or likely benign.
  \item \texttt{mpra\_hepg2}: variants tested in a lentiviral massively parallel reporter assay (lentiMPRA) \citep{gordon2020lentimpra} and released by the IGVF Consortium \citep{igvf2024deciphering} (accession IGVFFI4378PZYI). Positives are expression-modulating variants at FDR $<0.05$, and negatives are assayed variants that do not reach this threshold. Despite the dataset identifier, the assay was performed in HEK293T cells.
  \item \texttt{eqtl\_grasp}: variants from the set assembled for DeepSEA and redistributed with BEND. Positives are variants reported as eQTLs in GRASP, and negatives are 1000 Genomes variants not reported as eQTLs.
  \item Subsampling: each dataset is randomly subsampled to about 15{,}000 variants, stratified by split and label, which preserves the split proportions and the positive rate of the source.
  \item Split: chromosome-held-out. Chromosomes 5--7 form the evaluation split, chromosomes 8--10 a test split, and the remaining chromosomes the training pool, so that a model cannot score well by memorizing a locus. Because \texttt{eqtl\_grasp} spans only two chromosomes, its training pool is chromosome 1, and its evaluation and test splits are the chromosome-10 variants at or below and above the median variant position, respectively.
  \item Composition floors: a submission must retain at least 100 positives, 100 negatives, and variants from five chromosomes (one for \texttt{eqtl\_grasp}, whose training pool is a single chromosome).
\end{itemize}

\paragraph{Models.}
All hyperparameters are fixed in advance, and training uses a fixed seed.
\begin{itemize}
  \item DNABERT logistic: logistic regression with balanced class weights on mean-pooled final-layer embeddings from a frozen DNABERT (6-mer) \citep{ji2021dnabert}, which tokenizes the window into overlapping 6-mers.
  \item 4-mer logistic: logistic regression on the frequencies of the 136 canonical 4-mers in the window (a 4-mer and its reverse complement are counted as one feature), which represent sequence composition only.
\end{itemize}

\paragraph{Metrics.}
All metrics are computed on the evaluation split.
\begin{itemize}
  \item AUPRC (headline): the area under the precision--recall curve, computed as average precision. Its chance level is close to the positive rate of the evaluation split rather than 0.5.
  \item AUPRC lift, the ratio of AUPRC to that positive rate, and ROC-AUC.
  \item Precision among the 100 top-ranked variants, and recall within the top 10\% of the ranking.
  \item The Matthews correlation coefficient \citep{matthews1975comparison}, computed on rank-based calls that label the $k$ top-ranked variants positive, where $k$ is the number of positives in the evaluation split, and the Brier score \citep{brier1950verification}.
  \item Guard: the number of chromosomes shared between the curated training set and the evaluation split, which must be zero. A non-zero count sets the tracked score to zero.
  \item Validity flag: false when the scores are constant or non-finite. A false flag sets the tracked score to zero.
\end{itemize}

\subsubsection{Single-cell cell-type annotation}
\label{app:dom-sccelltype}

Three annotation tasks are crossed with five frozen classifiers, giving 15 cells. Each row is one cell of a harmonized single-cell dataset. The classifier is trained on the curated cells and scored on the held-out cells of the evaluation split.

\paragraph{Datasets.}
\begin{itemize}
  \item \texttt{pancreas\_crosstech}: cross-technology annotation of human pancreatic cells, built from the human pancreas atlas of the scIB benchmark \citep{luecken2022scib} and restricted to 12 of its 14 cell types. The inDrop cells of one study form the training pool, CEL-Seq2 and Fluidigm C1 cells from two other studies form the evaluation split, and CEL-Seq, SMART-Seq2, and SMARTer cells from three further studies form a test split.
  \item \texttt{pbmc\_crossprotocol}: cross-protocol annotation of peripheral blood mononuclear cells (PBMCs) from the single-laboratory protocol comparison of \citet{ding2020systematic}, using five of its protocols and restricted to the five cell-type labels shared by all five. Cells profiled with 10x Chromium v2 and v3 form the training pool, inDrops cells form the evaluation split, and Drop-seq and Seq-Well cells form a test split. 
  \item \texttt{pbmc\_open\_set}: open-set annotation with rejection on the Zheng 68K PBMC data, using the cells and held-out populations of the unseen-population experiment of \citet{abdelaal2019comparison}. Cells of the seven known populations, all from a single study, are split 70/15/15 into the training pool, the evaluation split, and the test split, stratified by label. The four CD4$^{+}$ T-cell populations are excluded from the training pool and the evaluation split and appear only in the test split, where they form a single unknown class. The evaluation split therefore contains known cells only.
  \item Composition floor: a submission must retain at least three cells of every class.
\end{itemize}

\paragraph{Models.}
All classifiers except the majority control take as input $\log(1+\mathrm{CP10k})$-normalized expression of a panel of 2{,}000 highly variable genes selected on the training pool. The two linear models use balanced class weights. All hyperparameters are fixed, and training runs on CPU with a fixed seed. On the open-set task, a cell is assigned to the unknown class when its maximum class posterior probability falls below 0.70, the rejection threshold used by \citet{abdelaal2019comparison}.
\begin{itemize}
  \item SVM: a linear support vector machine with sigmoid (Platt) calibration \citep{platt2000probabilities}.
  \item Logistic: a multinomial logistic regression.
  \item ACTINN \citep{ma2020actinn}: a multilayer perceptron with three hidden layers.
  \item SSVAE: a semi-supervised variational autoencoder in the style of scANVI \citep{xu2021probabilistic}.
  \item Majority: a control that predicts the training class prior for every cell.
  \item Gene mask: in addition to the row subset, the evaluation harness accepts a boolean mask over the full gene axis. The mask must select at least one gene and replaces the default panel. The seed script sets the mask to this panel, the plain protocol declares the gene axis immutable, and the \alg protocol lists the mask as a second lever.
\end{itemize}

\paragraph{Metrics.}
All metrics are computed on the evaluation split.
\begin{itemize}
  \item Macro-F1 over the scored classes (headline), restricted to the known classes on the open-set task.
  \item Accuracy and balanced accuracy.
  \item Minimum per-class recall, which reveals a rare class that is missed despite a high average score.
  \item Coverage, the fraction of cells that receive a call.
  \item Validity flag: on the two closed-set tasks, false when coverage falls below 0.95. On the open-set task, where a rejection counts as a call, false when fewer than half of the known cells receive a known-class call, so that a classifier cannot gain precision by abstaining. A false flag sets the tracked score to zero.
\end{itemize}

\subsubsection{Single-cell batch integration}
\label{app:dom-scbatch}

Three atlases are crossed with five frozen integration methods, giving 15 cells. Each row is one cell. Every method is fitted on the curated training cells as a reference, and the evaluation cells are then mapped onto it as queries. The methods receive only expression profiles and batch labels, and cell-type labels are used only by the scorer.

\paragraph{Datasets.}
The three atlases are the immune, lung, and pancreas datasets of the scIB benchmark \citep{luecken2022scib}, used as published.
\begin{itemize}
  \item \texttt{immune}: 10 batches of human peripheral blood and bone marrow cells from five studies, annotated with 16 cell types.
  \item \texttt{lung}: 16 batches from three cohorts, annotated with 17 cell types.
  \item \texttt{pancreas}: nine batches spanning six sequencing technologies (one inDrop study contributes four donor batches), annotated with 14 cell types.
  \item Composition floors: a submission must retain at least 10 cells of every batch and five cells of every cell type.
\end{itemize}

\paragraph{Models.}
Each method outputs an embedding of the evaluation cells. All five are self-contained implementations with fixed hyperparameters, run on CPU with a fixed seed, and take as input a panel of 2{,}000 highly variable genes selected on the training pool.
\begin{itemize}
  \item PCA: an uncorrected 50-component PCA, which serves as the negative control.
  \item ComBat \citep{johnson2007combat}: an empirical Bayes location and scale adjustment whose per-batch parameters are estimated on the reference and applied to the evaluation cells, followed by projection onto a 50-component PCA of the corrected reference.
  \item Harmony \citep{korsunsky2019harmony}: fitted on the reference, with the evaluation cells mapped into the corrected space.
  \item cVAE: a conditional variational autoencoder in the style of scVI \citep{lopez2018deep}, with a Gaussian likelihood and the batch as a decoder covariate.
  \item MMD-AE: an autoencoder with a maximum mean discrepancy (MMD) penalty \citep{gretton2012kernel} between the latent distributions of different batches, in the style of SAUCIE \citep{amodio2019exploring}.
  \item Gene mask: as in cell-type annotation, the evaluation harness accepts a boolean mask over the full gene axis in addition to the row subset. Here the mask must select 3 to 2{,}000 genes. The seed script sets the mask to the default panel, the plain protocol declares the gene axis immutable, and the \alg protocol lists the mask as a second lever.
\end{itemize}

\paragraph{Metrics.}
All metrics are computed on the evaluation split.
\begin{itemize}
  \item scIB-style overall score (headline): 0.6 times a biological-conservation score plus 0.4 times a batch-removal score, the weighting used by scIB. It is computed without scIB's per-metric min--max scaling across methods and is therefore not directly comparable with published scIB values.
  \item Biological conservation: the mean of the adjusted Rand index (ARI) and normalized mutual information (NMI) between $k$-means clusters and cell-type labels, the cell-type average silhouette width (ASW), the cell-type local inverse Simpson's index (cLISI), the isolated-label F1 score, and the isolated-label ASW.
  \item Batch removal: the mean of the batch ASW, the integration local inverse Simpson's index (iLISI), graph connectivity, and the principal component regression (PCR) comparison, defined as the relative reduction in the variance explained by batch compared with an unintegrated PCA embedding of the same cells.
  \item Guards: the collapse guard measures the cell-type structure of the embedding, rescaled against that of a random embedding. The batch-structure guard is the ARI between a clustering of the embedding and the batch labels, which is high when the embedding remains organized by batch.
  \item Validity flag: false when fewer than 80\% of the 10 component metrics are finite or when the embedding is degenerate (zero total variance). A false flag sets the tracked score to zero.
\end{itemize}

\subsubsection{Single-cell perturbation-response prediction}
\label{app:dom-scpert}

Two single-cell CRISPR screens are crossed with up to four frozen predictors, all four on \texttt{norman19} and three on \texttt{frangieh21}, giving 7 dataset--model pairs. Each dataset is the processed AnnData object distributed with PerturBench \citep{wu2025perturbench}. Each row is one cell, labeled with its perturbation and covariates, and expression is log-normalized by scaling the counts of each cell to the median library size of the dataset and then applying $\log(1+x)$.

\paragraph{Datasets.}
\begin{itemize}
  \item \texttt{norman19}: a CRISPR activation (CRISPRa) screen in K562 cells with single-gene and two-gene perturbations. The test split contains 98 two-gene combinations and the validation split nine others. Every constituent gene is also perturbed individually in the training pool, so the task is to predict the effects of unseen combinations.
  \item \texttt{frangieh21}: a Perturb-CITE-seq CRISPR knockout screen in a melanoma cell line under three conditions (control, interferon-$\gamma$ stimulation, and co-culture with T cells).
  
  \item Split of \texttt{frangieh21}: the validation and test splits hold out perturbation--context pairs that do not occur in the training pool.
\end{itemize}

\paragraph{Models.}
All four models are native PyTorch \citep{paszke2019pytorch} reimplementations. CPA, the latent-additive model, and SAMS-VAE are ported from PerturBench \citep{wu2025perturbench}, and GEARS \citep{roohani2024predicting} is ported from its authors' code. Each model uses the per-dataset hyperparameters tuned in PerturBench, fixed in advance (for GEARS, those tuned on \texttt{norman19}).
Each model is trained on a single GPU.
\begin{itemize}
  \item CPA: a variational autoencoder whose latent representation is the sum of a basal cell state, a perturbation embedding, and covariate embeddings, trained with an adversarial objective that removes perturbation and covariate information from the basal state. For counterfactual prediction, a control cell is encoded into its basal state, the target perturbation and covariate embeddings are added, and the result is decoded.
  \item Latent additive: an encoder maps a control cell to a latent representation, a learned perturbation embedding is added to it, and a decoder predicts the perturbed expression profile. The model is trained with a mean-squared-error loss.
  \item SAMS-VAE: a sparse additive mechanism shift variational autoencoder, in which a learned sparse mask selects the latent dimensions affected by each perturbation, the masked perturbation effects are summed into a latent shift for each cell, and a basal encoder models the remaining variation.
  \item GEARS, on \texttt{norman19} only: a graph neural network that takes the expression profile of a control cell and a binary perturbation indicator for each gene and predicts the post-perturbation expression profile. It uses a gene co-expression graph built from the training control cells and, for the perturbation embeddings, a gene-similarity graph derived from Gene Ontology \citep{ashburner2000gene} annotations.
  \item Training recipe: CPA, the latent-additive model, and SAMS-VAE share one recipe, which uses the Adam optimizer \citep{kingma2015adam}. Learning rates and maximum numbers of epochs are set per dataset. GEARS keeps its own recipe, which trains for a fixed number of epochs and retains the weights from the epoch with the lowest loss on the same validation split.
\end{itemize}

\paragraph{Metrics.}
For each held-out perturbation, or each held-out perturbation--context pair on \texttt{frangieh21}, the evaluation compares the predicted with the measured pseudobulk log fold-change relative to control cells of the same context. Because the held-out cells of \texttt{norman19} include no control cells, its fold-changes are computed relative to the training controls. Each metric is averaged over the held-out perturbations (or perturbation--context pairs).
\begin{itemize}
  \item Cosine similarity between the predicted and measured fold-change vectors (headline).
  \item Pearson correlation between the predicted and measured fold-changes, computed over all genes and over the 20 genes with the largest absolute measured fold-change.
  \item RMSE and $R^2$ of the predicted fold-change.
  \item Recall of these 20 genes among the 20 genes with the largest absolute predicted fold-change.
  \item Rank metric of PerturBench, computed with RMSE as the distance: for each measured fold-change, the fraction of the other held-out predictions that lie at least as close to it as its matched prediction. It is zero for a perfect predictor, about one half for an uninformative one, and approaches one under mode collapse.
\end{itemize}

\subsubsection{Instruction tuning}
\label{app:dom-it}

Two setups each pair one pool of instruction-tuning examples with one base language model, giving 2 cells. A row is one example, and the budget is 10{,}000 rows. Iteration~0 evaluates the untrained base model, iteration~1 fine-tunes on a seeded random sample of the budget size, and the agent's own selections follow.

\paragraph{Datasets.}
\begin{itemize}
  \item T\"ulu 3 SFT mixture: the supervised fine-tuning mixture released with T\"ulu 3, 939{,}343 rows.
  \item LESS mixture: the 270{,}679-row pool assembled by LESS from FLAN v2 (100{,}000 rows), CoT (100{,}000), Dolly (15{,}011), and Open Assistant (55{,}668).
\end{itemize}

\paragraph{Models and training recipe.}
Every iteration runs one epoch of full-parameter supervised fine-tuning of the base model on the selected rows, in bf16 with an effective batch of 16 sequences and 3\% warm-up.
\begin{itemize}
  \item OLMo-2-1B on the T\"ulu 3 pool: peak learning rate $5\times10^{-6}$, linear schedule, maximum sequence length 4{,}096 tokens.
  \item Llama-2-7B on the LESS pool: peak learning rate $2\times10^{-5}$, linear schedule, maximum sequence length 2{,}048 tokens.
\end{itemize}

\paragraph{Metrics.}
Each setup is scored by a fixed benchmark suite with greedy decoding; the score is the unweighted mean of the benchmark accuracies.
\begin{itemize}
  \item T\"ulu 3 setup: MMLU (zero-shot), GSM8K (8-shot), BBH (3-shot), TruthfulQA (MC2), and IFEval (prompt-level loose accuracy).
  \item LESS setup: MMLU (5-shot) and BBH (3-shot).
\end{itemize}

\subsubsection{Code instruction tuning}
\label{app:dom-code}

One setup pairs a pool of code instruction examples with one code model, giving 1 cell. A row is one example, and the budget is 10{,}000 rows. As in instruction tuning, iteration~0 evaluates the untrained base model and iteration~1 fine-tunes on a seeded random sample of the budget size.

\paragraph{Datasets.}
\begin{itemize}
  \item OpenCodeInstruct: 5{,}000{,}000 instruction-response pairs for code.
\end{itemize}

\paragraph{Model and training recipe.}
\begin{itemize}
  \item Qwen2.5-Coder-1.5B-Instruct: one epoch of full-parameter supervised fine-tuning in bf16 with an effective batch of 16 sequences, peak learning rate $5\times10^{-6}$ with a cosine schedule and 3\% warm-up, maximum sequence length 2{,}048 tokens, and deterministic kernels.
\end{itemize}

\paragraph{Metrics.}
Pass@1 with greedy decoding; the score is the unweighted mean of the two benchmarks.
\begin{itemize}
  \item HumanEval+. HumanEval itself is omitted because it shares its 164 problems with HumanEval+.
  \item MBPP+.
\end{itemize}

\subsection{Data processing baselines for biology tasks}
\label{app:bio-baselines}

\begin{figure}[h]
\centering
\includegraphics[width=\linewidth]{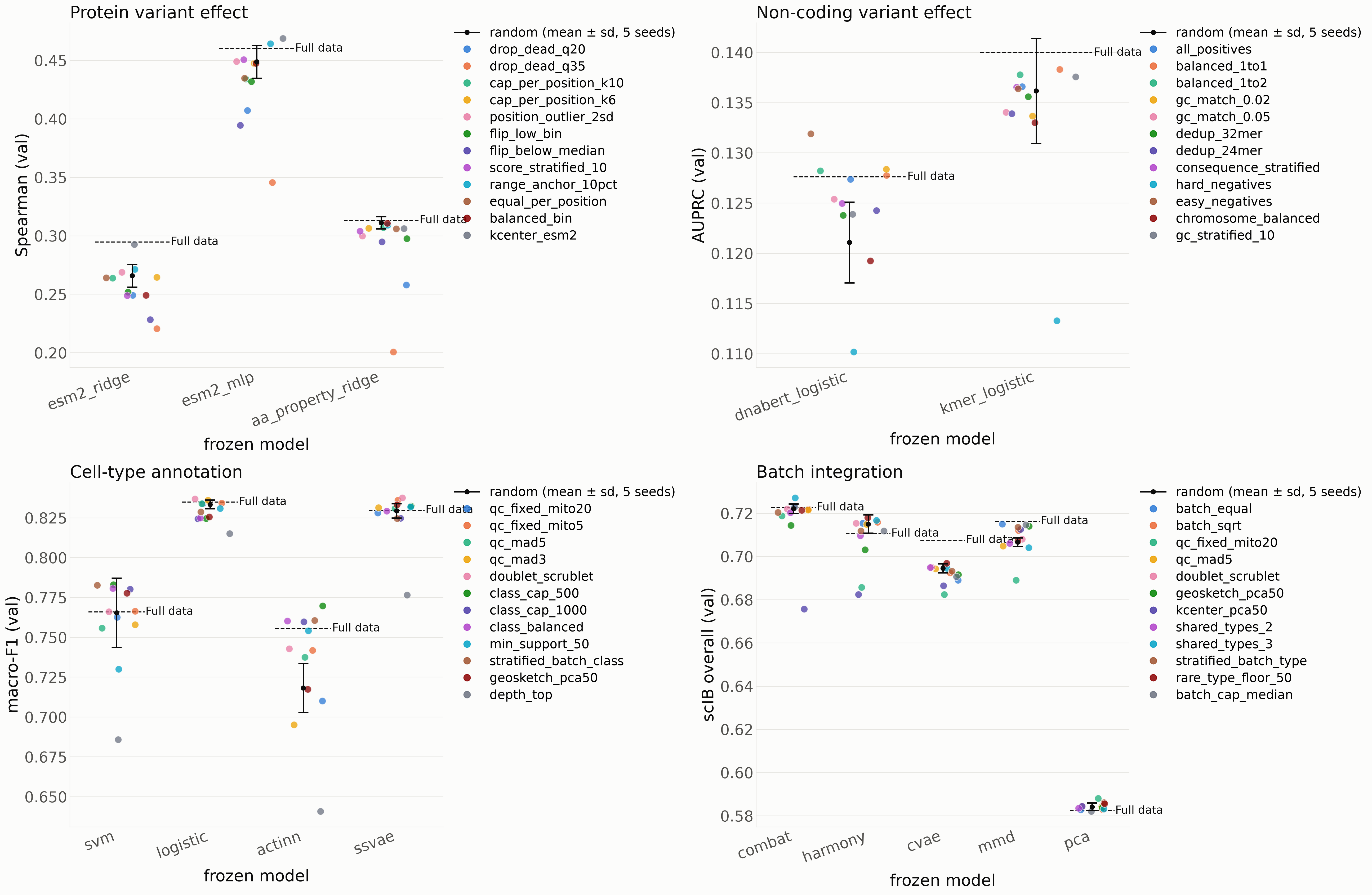}\vspace{-0.0em}
\caption{Selecting 50\% data using standard, human-designed data processing pipelines for biology tasks. Biology tasks typically apply standard, human-designed processing pipelines on the datasets, and apply the sanitized datasets for training different ML models in downstream applications. Yet, different ML models and training pipelines carry different assumption for the data. The fixed data cleaning protocol may yield substantially different outcomes in different use cases. As depicted in the figure, in every case, there are more processing pipelines underperforming training on unprocessed data.}\vspace{-0.5em}
\label{fig:baseline-methods}
\end{figure}

This appendix lists the standard curation recipes evaluated as human-designed baselines
(Figure~\ref{fig:baseline-methods}, Table~\ref{tab:main-bio}), the practice or publication each one
follows, and how each was implemented under the benchmark's contract. Every recipe is applied to
the training pool of a task and must return a training set of exactly the budget
(50\% of the pool) that respects the package's composition floors; Section~\ref{app:recipes:protocol}
gives the common protocol, Sections~\ref{app:recipes:pv}--\ref{app:recipes:sb} the recipes per
domain.

\subsubsection{Common protocol}
\label{app:recipes:protocol}

\paragraph{Budget and floors.}
The harness accepts only submissions of exactly \texttt{target\_rows} rows, half the training pool
rounded to an integer (Table~\ref{tab:setup-grid}), and the package validator rejects at
evaluation time any subset that violates the domain's composition floors. The protein floors are
at least 30 distinct residue positions and 10 distinct substituted amino acids. The non-coding
floors are at least 100 positives, 100 negatives and 5 chromosomes, with the chromosome floor
capped at the number of chromosomes in the training pool (one for \texttt{eqtl\_grasp}).
Annotation requires at least 3 cells per cell type, and integration at least 10 cells per batch
and 5 per cell type. Every recipe therefore first reserves a set of rows that satisfies the floors
and then fills the rest of the budget according to its rule.

\paragraph{Filter, allocation and ranking recipes.}
Apart from \texttt{random}, each recipe is one of three kinds. A \emph{filter} marks rows as
eligible (for example by a QC threshold, a GC-matching rule or a de-duplication rule), and the
budget is drawn uniformly from the eligible rows. If fewer eligible rows than the budget remain,
the shortfall is drawn uniformly from the excluded rows (\emph{filter-then-fill}), and the numbers
of eligible and filled rows are recorded with each artifact. An \emph{allocation} distributes the
budget over strata (positions, classes, batches or geometric-sketch boxes) by water-filling. Each
stratum receives an equal or weighted share, capped by its size and by any recipe-specific cap,
and the remainder is redistributed until the budget is met. A \emph{ranking} orders rows by a
score and keeps the highest-ranked rows up to the budget. The fill seed determines which rows are
drawn within a stratum or an eligible set and breaks ties between equal scores.

\paragraph{Seeds and scoring.}
Every recipe is materialized with five fill seeds, and the random subset with ten further seeds
(fifteen draws in total) to provide the random-search reference. Each artifact is scored with the
frozen model on the evaluation split through the same entry point that the agents use.
As in the main results, a submission that fails a validity guard (for example the open-set guard
requiring that at least half of the cells of known types are not rejected as unknown) is scored
0. Recipe statistics are five-seed means, and the ``best recipe per subtask'' entry in
Table~\ref{tab:main-bio} is the best five-seed mean among the twelve recipes of each
dataset--model pair.

\paragraph{Derived quantities.}
Recipes that need a representation use the same representation for every frozen model in a
domain. For the variant-effect domains these are the mean-pooled embeddings of ESM-2 (35M
parameters) and DNABERT (6-mer) that the embedding-based models use, read from each package's
per-sequence embedding cache. For single-cell data they are the first 50 principal components of
the scaled expression of the 2,000 highly variable genes flagged in each object. QC covariates are
the per-cell columns shipped with each object. The integration pancreas object ships none, so
library size, detected genes and mitochondrial fraction are computed from its count-like
\texttt{counts} layer.

\begin{table}[h]
\centering
\small
\caption{Protein variant-effect recipes (four assays, three frozen regressors). Kind: F filter,
A allocation, R ranking.}
\label{tab:recipes-pv}
\resizebox{\ifdim\width>\linewidth\linewidth\else\width\fi}{!}{
\begin{tabular}{@{}l l p{5.2cm} l p{3.6cm}@{}}
\toprule
Recipe & Kind & Rule & Default & Practice / source \\
\midrule
\texttt{random} & -- & uniform subset, floors reserved & -- & seeded baseline; random training subsets in training-size studies \citep{gelman2021neural,hsu2022learning} \\
\texttt{drop\_dead\_q20}, \texttt{\_q35} & F & drop variants below the $q$-quantile of the assay score & $q=0.20,\,0.35$ & removal of low-fitness ``holes'' \citep{wittmann2021informed} \\
\texttt{cap\_per\_position\_k10}, \texttt{\_k6} & A & at most $k$ variants per residue position & $k=10,\,6$ & position breadth over depth \citep{sruthi2020deep2full} \\
\texttt{position\_outlier\_2sd} & F & drop variants $>2$\,sd from their position's median & 2 sd & proxy for measurement-error and read-count filtering \citep{rubin2017enrich2,faure2020dimsum} \\
\texttt{flip\_low\_bin} & F & keep only variants in ProteinGym's non-functional bin & bin $=0$ & FLIP low-versus-high split \citep{dallago2021flip} \\
\texttt{flip\_below\_median} & F & keep only variants below the median score & median & median analog of the same split \citep{dallago2021flip} \\
\texttt{score\_stratified\_10} & A & equal quota per score decile & 10 bins & per-bin coverage of the score range \citep{sruthi2020deep2full} \\
\texttt{range\_anchor\_10pct} & R & keep both 10\% tails of the score, fill the rest at random & 10\% & score-range anchoring, the opposite of holes removal \citep{wittmann2021informed} \\
\texttt{equal\_per\_position} & A & equal quota per residue position & -- & position-uniform design \citep{sruthi2020deep2full} \\
\texttt{balanced\_bin} & A & equal quota per binary label bin & 1:1 & ProteinGym binarization \citep{notin2023proteingym} \\
\texttt{kcenter\_esm2} & R & farthest-first traversal of the ESM-2 embedding space & $k=$ budget & core-set selection \citep{sener2018coreset,gonzalez1985clustering} \\
\bottomrule
\end{tabular}}
\end{table}

\begin{table}[h]
\centering
\small
\caption{Non-coding variant-effect recipes (three datasets, two frozen classifiers).}
\label{tab:recipes-nc}
\resizebox{\ifdim\width>\linewidth\linewidth\else\width\fi}{!}{
\begin{tabular}{@{}l l p{5.2cm} l p{3.6cm}@{}}
\toprule
Recipe & Kind & Rule & Default & Practice / source \\
\midrule
\texttt{random} & -- & uniform subset, floors reserved & -- & seeded baseline \\
\texttt{all\_positives} & R & keep every positive, fill with random negatives & -- & majority-class undersampling \citep{whalen2022pitfalls} \\
\texttt{balanced\_1to1}, \texttt{\_1to2} & A & target positive:negative ratio & 1:1, 1:2 & class rebalancing \citep{kircher2014cadd,whalen2022pitfalls} \\
\texttt{gc\_match\_0.02}, \texttt{\_0.05} & F & at most as many negatives as positives per GC bin & bin width 0.02, 0.05 & GC-matched null sets \citep{ghandi2014gkmsvm,lee2015deltasvm} \\
\texttt{dedup\_32mer}, \texttt{\_24mer} & F & drop windows sharing any exact $k$-mer with a kept window & $k=32,\,24$ & redundancy / LD pruning \citep{purcell2007plink,whalen2022pitfalls} \\
\texttt{consequence\_stratified} & A & equal quota per consequence class & -- & consequence annotations \citep{landrum2018clinvar,marin2024bend} \\
\texttt{hard\_negatives} & R & negatives ranked by out-of-fold positive-class probability, hardest first & top half & hard-example mining \citep{shrivastava2016ohem} \\
\texttt{easy\_negatives} & R & as above, easiest first & bottom half & contrast to hard-negative mining \\
\texttt{chromosome\_balanced} & A & equal quota per chromosome & -- & chromosome-aware design \citep{whalen2022pitfalls,zhou2015deepsea} \\
\texttt{gc\_stratified\_10} & A & equal quota per GC decile & 10 bins & GC-content control \citep{ghandi2014gkmsvm} \\
\bottomrule
\end{tabular}}
\end{table}

\begin{table}[h]
\centering
\small
\caption{Cell-type annotation recipes (three datasets, five frozen classifiers).}
\label{tab:recipes-ct}
\resizebox{\ifdim\width>\linewidth\linewidth\else\width\fi}{!}{
\begin{tabular}{@{}l l p{5.2cm} l p{3.6cm}@{}}
\toprule
Recipe & Kind & Rule & Default & Practice / source \\
\midrule
\texttt{random} & -- & uniform subset, floors reserved & -- & seeded baseline \\
\texttt{qc\_fixed\_mito20}, \texttt{\_mito5} & F & genes $\ge 200$, counts $\ge 500$, mitochondrial fraction below threshold & 20\%, 5\% & tutorial thresholds \citep{wolf2018scanpy,hao2021seurat,luecken2019practices} \\
\texttt{qc\_mad5}, \texttt{qc\_mad3} & F & per-batch MAD outliers on log counts and log genes (both tails) and on mitochondrial fraction (upper tail) & 5, 3 MADs & MAD-based QC \citep{lun2016stepbystep,germain2020pipecomp,heumos2023practices} \\
\texttt{doublet\_scrublet} & F & remove doublets called by Scrublet within each batch & default threshold & Scrublet \citep{wolock2019scrublet}; scDblFinder \citep{germain2021scdblfinder} \\
\texttt{class\_cap\_500}, \texttt{\_1000} & A & at most $N$ cells per class, remainder water-filled & $N=500,\,1000$ & down-weighting abundant types \citep{dominguezconde2022crosstissue} \\
\texttt{class\_balanced} & A & equal quota per class & -- & type-balanced sampling \citep{dominguezconde2022crosstissue}; errors on small populations \citep{abdelaal2019comparison} \\
\texttt{min\_support\_50} & F & reserve 50 cells per class (all if fewer), then random & 50 & minimum class size \citep{abdelaal2019comparison} \\
\texttt{stratified\_batch\_class} & A & equal quota per (batch, class) stratum & -- & multi-batch reference building \citep{lotfollahi2022scarches,kang2021symphony} \\
\texttt{geosketch\_pca50} & A & plaid covering of PCA-50 space, equal quota per occupied box & boxes $\ge$ budget/2 & geometric sketching \citep{hie2019geosketch} \\
\texttt{depth\_top} & R & keep the cells with the most detected genes & top half & detected genes as quality proxy \citep{ilicic2016lowquality,ding2020systematic} \\
\bottomrule
\end{tabular}}
\end{table}

\begin{table}[h]
\centering
\small
\caption{Batch integration recipes}
\label{tab:recipes-bi}
\resizebox{\ifdim\width>\linewidth\linewidth\else\width\fi}{!}{
\begin{tabular}{@{}l l p{5.2cm} l p{3.6cm}@{}}
\toprule
Recipe & Kind & Rule & Default & Practice / source \\
\midrule
\texttt{random} & -- & uniform subset, floors reserved & -- & seeded baseline \\
\texttt{batch\_equal} & A & equal quota per batch & -- & unequal batch sizes \citep{johnson2007combat,korsunsky2019harmony,tran2020benchmark} \\
\texttt{batch\_sqrt} & A & quota proportional to $\sqrt{\text{batch size}}$ & -- & square-root allocation \\
\texttt{qc\_fixed\_mito20} & F & genes $\ge 200$, counts $\ge 500$, mitochondrial fraction $<20\%$ & 20\% & threshold-based QC \citep{luecken2019practices} \\
\texttt{qc\_mad5} & F & per-batch outliers beyond 5 MADs & 5 MADs & MAD-based QC \citep{lun2016stepbystep,heumos2023practices} \\
\texttt{doublet\_scrublet} & F & remove Scrublet-called doublets per batch & default threshold & \citet{wolock2019scrublet} \\
\texttt{geosketch\_pca50} & A & plaid covering of PCA-50 space & boxes $\ge$ budget/2 & geometric sketching \citep{hie2019geosketch} \\
\texttt{kcenter\_pca50} & R & farthest-first traversal of PCA-50 space & $k=$ budget & Hopper \citep{demeo2020hopper}; \citet{gonzalez1985clustering} \\
\texttt{shared\_types\_2}, \texttt{\_3} & F & keep cell types with $\ge 10$ cells in at least $k$ batches & $k=2,\,3$ & populations shared across batches \citep{haghverdi2018mnn,stuart2019comprehensive,welch2019liger} \\
\texttt{stratified\_batch\_type} & A & equal quota per (batch, cell type) stratum & -- & stratified sampling \\
\texttt{rare\_type\_floor\_50} & F & reserve at least 50 cells per type, then random & 50 & rare cell-type retention \citep{luecken2022scib,hie2019geosketch} \\
\texttt{batch\_cap\_median} & A & cap each batch at the median batch size & median & downsampling of large batches \\
\bottomrule
\end{tabular}}
\end{table}

\subsubsection{Protein variant effect}
\label{app:recipes:pv}

\paragraph{Methods.}
Deep mutational scanning tables are published after enrichment scoring, wild-type normalization
and quality filtering \citep{rubin2017enrich2,faure2020dimsum}. ProteinGym
\citep{notin2023proteingym} redistributes the assay scores and the cross-validation folds but no
read counts, replicate values or standard errors, so neither the standard-error nor the read-count
filter can be reproduced here, and \texttt{position\_outlier\_2sd} substitutes within-position
dispersion for measurement error. In machine-learning-assisted directed evolution, reducing the
inclusion of ``holes'' (variants with zero or extremely low fitness, which are minimally
informative) was the most important of the design choices evaluated for reaching the fitness
maximum of a combinatorial library \citep{wittmann2021informed}, and at a fixed measurement budget
a training set that represents every position predicts better than one that measures all
substitutions at a few sites \citep{sruthi2020deep2full}. FLIP \citep{dallago2021flip} evaluates
transfer from one fitness regime to another. Its low-versus-high split trains on the variants at or
below wild-type fitness and tests on the remainder. A subset that covers the feature space is the
core-set criterion \citep{sener2018coreset}, applied here as farthest-first traversal, a greedy
approximation to the $k$-center problem \citep{gonzalez1985clustering}.

\paragraph{Implementation.}
Floors are reserved as one variant per position and one per substituted amino acid. Quantile
recipes use the empirical quantile of \texttt{DMS\_score} over the pool, and the position-outlier
rule leaves positions with fewer than three variants intact. Caps and stratifications are
water-filled, and \texttt{range\_anchor\_10pct} ranks both tails above the interior and fills the
remaining budget from the interior at random. The non-functional bin is smaller than the budget on
\texttt{pten\_human}, \texttt{tpmt\_human} and \texttt{cp2c9\_human}, so \texttt{flip\_low\_bin} is
completed there by filter-then-fill, whereas the below-median half matches the budget by
construction, up to rounding. \texttt{kcenter\_esm2} standardizes the cached 480-dimensional
mean-pooled ESM-2 embeddings per dimension and runs farthest-first traversal from a seed-dependent
start point. The same space is used for all three regressors, including the amino-acid-property
ridge, which is fitted on 57 substitution-chemistry features and never sees it.

\subsubsection{Non-coding variant effect}
\label{app:recipes:nc}

\paragraph{Methods.}
CADD trains a variant classifier on equal numbers of high-frequency human-derived alleles and
simulated variants \citep{kircher2014cadd}, and the DeepSEA functional-variant classifiers contrast
curated positives with 1000 Genomes SNPs of controlled minor allele frequency, balancing the classes
with sample weights \citep{zhou2015deepsea}. The gkm-SVM and deltaSVM classifiers are trained against an
equal number of random genomic sequences matched to the positives in length, GC content and repeat fraction
\citep{ghandi2014gkmsvm,lee2015deltasvm}. Variants in linkage disequilibrium are commonly pruned to
an approximately independent subset \citep{purcell2007plink} and repeated sequences removed by
de-duplication, and grouping examples by chromosome is standard guidance against leakage between
dependent examples \citep{whalen2022pitfalls}. Hard-negative mining, which trains on the negatives
with the highest loss under the current model, is standard in training object detectors
\citep{shrivastava2016ohem}, and \texttt{easy\_negatives} reverses the ranking as a contrast.

\paragraph{Implementation.}
Floors reserve 100 positives, 100 negatives and one variant from each training chromosome (the
five-chromosome floor is capped at one for \texttt{eqtl\_grasp}, whose training pool is
chromosome 1). Ratio recipes set the number of positives to
$\min(n_{\mathrm{pos}}, \lfloor \text{budget} \times \text{fraction} \rceil)$ and take the rest as
random negatives. At a budget of half the pool, positives are too few to reach either ratio on
\texttt{eqtl\_grasp} and \texttt{mpra\_hepg2}, or 1:1 on \texttt{clinvar\_splice}, so these
recipes keep every positive, as \texttt{all\_positives} does. GC matching bins
\texttt{gc\_content} at the stated width and keeps every positive and, in each bin, at most as many
negatives as positives. No bin holds more positives than negatives, so at both widths the eligible
set is all positives plus an equal number of GC-matched negatives (1,552 rows on
\texttt{mpra\_hepg2}, 1,968 on \texttt{eqtl\_grasp} and 4,568 on \texttt{clinvar\_splice}), and
unmatched negatives from the fill make up about 72\%, 66\% and 18\% of the budget, respectively.
De-duplication visits the pool in a seed-dependent order and keeps a window only if it shares no
$k$-mer with a window already kept. \texttt{consequence\_stratified} and
\texttt{chromosome\_balanced} reduce to random subsets on \texttt{mpra\_hepg2} (one consequence
label, \texttt{mpra\_assayed}) and \texttt{eqtl\_grasp} (one training chromosome), respectively.
Hardness scores are out-of-fold probabilities from a five-fold, class-weighted logistic regression
($C=0.1$) on standardized, mean-pooled DNABERT (6-mer) embeddings and are used for both frozen
classifiers. Positives are kept at the training pool's positive rate, and negatives are taken from
the top (hard) or bottom (easy) of the ranking.

\subsubsection{Cell-type annotation}
\label{app:recipes:ct}

\paragraph{Methods.}
Standard single-cell workflows remove low-quality cells on the basis of detected genes, library size
and mitochondrial fraction, either with fixed thresholds such as those of the Scanpy and Seurat
tutorials \citep{wolf2018scanpy,hao2021seurat} or with per-batch outlier rules based on the median
absolute deviation (MAD) \citep{lun2016stepbystep,germain2020pipecomp,heumos2023practices}. They
remove doublets with detectors that score each cell by its similarity to simulated doublets
\citep{wolock2019scrublet,germain2021scdblfinder,xi2021doublet}. CellTypist can balance cell types
within training mini-batches by sampling each cell with a probability inversely proportional to the
size of its cell type \citep{dominguezconde2022crosstissue}, and classifier accuracy decreases on
deeply annotated datasets that contain many small populations \citep{abdelaal2019comparison}.
Reference atlases for query mapping are built by integrating multiple datasets or batches
\citep{lotfollahi2022scarches,kang2021symphony}, and geometric sketching preserves rare cell states
when a dataset is downsampled \citep{hie2019geosketch}.

\paragraph{Implementation.}
Floors reserve three cells per cell type. QC covariates are the shipped per-cell columns
\texttt{ncounts} (library size), \texttt{ngenes} (detected genes) and \texttt{percent\_mito}, which is
stored as a percentage and divided by 100. The pancreas atlas contains no mitochondrial genes, so its
mitochondrial fraction is zero for every cell and the mitochondrial rules have no effect there. MAD
rules are applied within each batch (inDrop1--4 for pancreas, 10x Chromium v2 and v3 for
cross-protocol and a single batch for open-set) with the scaled MAD ($1.4826\times$). They are
two-sided on $\log(1+\text{counts})$ and $\log(1+\text{genes})$ and upper-tailed on the mitochondrial
fraction. Scrublet runs per batch on the raw-count layer (\texttt{counts} for the PBMC datasets and
\texttt{source\_expression} for pancreas, whose training cells have integer counts) with its default
automatic threshold. If automatic thresholding fails, the cells with
the top 5\% of doublet scores are called doublets. Class caps and stratified quotas are water-filled,
so a cap that cannot fill the budget, as with the 500-cell cap on the pancreas and cross-protocol
pools, is raised uniformly. The open-set dataset has a single batch, so
\texttt{stratified\_batch\_class} reduces to \texttt{class\_balanced} there. The plaid sketch divides
each of the 50 principal components into $B$ equal-width bins, increases $B$ until the number of
occupied boxes reaches half the budget, and then allocates the budget equally over the occupied
boxes. On the pancreas pool, which the atlas authors quality-controlled before release
\citep{luecken2022scib}, the fixed thresholds and the 5-MAD rule exclude no cells, so those artifacts
coincide with the random subset. On the open-set pool, from which cells with few detected genes were
removed before release \citep{abdelaal2019comparison}, the 20\% threshold likewise excludes no cells,
whereas the 5\% threshold excludes 10 and the 5-MAD rule 17 of the 4,263 training cells, all
upper-tail outliers. On the cross-protocol pool, whose QC columns are computed over the 2,191 genes
shared by all five protocols, \texttt{qc\_fixed\_mito5} retains only 3,286 of the 8,165 training cells
(including 5 of the 2,730 10x Chromium v3 cells), fewer than the 4,082-cell budget, so the shortfall is drawn
from the excluded cells.

\subsubsection{Batch integration}
\label{app:recipes:bi}
\label{app:recipes:sb}

\paragraph{Methods.}
Integration methods estimate batch-specific parameters from the reference, so the number of cells
in each batch affects the fit. ComBat estimates location and scale batch effects for each gene with
empirical Bayes shrinkage \citep{johnson2007combat}, and Harmony penalizes clusters whose batch
composition departs from the overall batch proportions \citep{korsunsky2019harmony}. Integration
benchmarks include tasks in which batches differ in cell number and cell-type composition
\citep{tran2020benchmark,luecken2022scib}. Mutual-nearest-neighbor and anchor-based methods assume
that at least one cell population is shared between batches
\citep{haghverdi2018mnn,stuart2019comprehensive}, and LIGER aligns datasets through shared factors
and joint clusters \citep{welch2019liger}, which motivates the shared-type filters. Geometric
sketching has been used to accelerate integration by applying the integration method to the sketch
and then transforming the full dataset \citep{hie2019geosketch}, and Hopper builds sketches by
farthest-first traversal, a greedy approximation to the $k$-center problem
\citep{demeo2020hopper,gonzalez1985clustering}.

\paragraph{Implementation.}
Floors reserve ten cells per batch and five per cell type. Batch allocations water-fill equal or
$\sqrt{\text{size}}$-weighted quotas, and \texttt{batch\_cap\_median} caps every batch at the median
batch size of the training pool (1,207, 1,444 and 1,752 cells for pancreas, lung and immune)
before water-filling. The capped pool still exceeds the budget on every atlas. Shared-type filters
keep cell types with at least ten cells in at least $k$ batches. On immune every cell type meets
this criterion at $k=3$, so neither filter excludes any cell. The QC and doublet recipes are
implemented as in the annotation domain (Section~\ref{app:recipes:ct}). The scIB pancreas object
ships no per-cell QC columns (lung and immune do), so library size and detected genes are computed
from its count-like layer. Its gene set contains no \texttt{MT-} genes, so the mitochondrial
fraction is zero and the mitochondrial criteria have no effect. The fixed thresholds exclude no
pancreas training cells. On lung they exclude 2,867 training cells, all from the four Drop-seq
batches and all with fewer than 500 counts. The $k$-center and geometric sketches use the PCA-50
coordinates of the common protocol (Section~\ref{app:recipes:protocol}).

\subsection{Fine-tuning recipes of the instruction-tuning setups}
\label{app:llm-recipe}

The harness fixes the entire fine-tuning stage, so that the only variable across an agent's iterations is the curated subset. Given a submitted subset, it fine-tunes the setup's base model on a single NVIDIA A100 GPU for one epoch under a fixed recipe; no training hyperparameter is exposed to the agent, and every iteration starts from the same base checkpoint. Iteration~0 evaluates the untrained base model. Each base model has its own recipe: OLMo-2-1B on the T\"ulu~3 SFT mixture (Table~\ref{tab:recipe-tulu3}), Llama-2-7B on the LESS mixture (Table~\ref{tab:recipe-less}), and Qwen2.5-Coder-1.5B-Instruct on OpenCodeInstruct (Table~\ref{tab:recipe-code}); all three update every parameter of the model.

\begin{table}[htbp]
\centering
\small
\caption{Fine-tuning settings for OLMo-2-1B on the T\"ulu~3 SFT mixture.}
\label{tab:recipe-tulu3}
\begin{tabular}{ll}
\toprule
\textbf{Component} & \textbf{Value} \\
\midrule
Initial model & OLMo-2-0425-1B (base) \\
Training hardware & 1 NVIDIA A100 GPU \\
Training epochs & 1 \\
Trainable modules & All parameters (full fine-tuning) \\
Per-device batch size & 1 \\
Gradient accumulation & 16 \\
Learning rate & $5\times10^{-6}$ \\
Learning-rate schedule & Linear \\
Warmup ratio & 0.03 \\
Weight decay & 0.0 \\
Max sequence length & 4{,}096 \\
Precision & bfloat16 \\
\bottomrule
\end{tabular}
\end{table}

\begin{table}[htbp]
\centering
\small
\caption{Fine-tuning settings for Llama-2-7B on the LESS mixture.}
\label{tab:recipe-less}
\begin{tabular}{ll}
\toprule
\textbf{Component} & \textbf{Value} \\
\midrule
Initial model & Llama-2-7B (base) \\
Training hardware & 1 NVIDIA A100 GPU \\
Training epochs & 1 \\
Trainable modules & All parameters (full fine-tuning) \\
Per-device batch size & 2 \\
Gradient accumulation & 8 \\
Learning rate & $2\times10^{-5}$ \\
Learning-rate schedule & Linear \\
Warmup ratio & 0.03 \\
Weight decay & 0.0 \\
Max sequence length & 2{,}048 \\
Precision & bfloat16 \\
\bottomrule
\end{tabular}
\end{table}

\begin{table}[htbp]
\centering
\small
\caption{Fine-tuning settings for Qwen2.5-Coder-1.5B-Instruct on OpenCodeInstruct.}
\label{tab:recipe-code}
\begin{tabular}{ll}
\toprule
\textbf{Component} & \textbf{Value} \\
\midrule
Initial model & Qwen2.5-Coder-1.5B-Instruct \\
Training hardware & 1 NVIDIA A100 GPU \\
Training epochs & 1 \\
Trainable modules & All parameters (full fine-tuning) \\
Per-device batch size & 4 \\
Gradient accumulation & 4 \\
Learning rate & $5\times10^{-6}$ \\
Learning-rate schedule & Cosine \\
Warmup ratio & 0.03 \\
Weight decay & 0.0 \\
Max sequence length & 2{,}048 \\
Precision & bfloat16 \\
Deterministic algorithms & On \\
\bottomrule
\end{tabular}
\end{table}

\providecolor{protoaccent}{RGB}{176,112,80}   
\providecolor{prototint}{RGB}{253,247,241}    
\tcbset{protobox/.style={enhanced, breakable, sharp corners,
  colback=prototint, colframe=protoaccent, boxrule=0.3pt, leftrule=2.5pt,
  colbacktitle=prototint, coltitle=protoaccent, fonttitle=\small\scshape,
  titlerule=0pt, toptitle=3pt, bottomtitle=1pt,
  left=6pt, right=6pt, top=3pt, bottom=4pt}}

\section{Examples of Agent Trajectory and Memory Files}

\subsection{Basic Agent Trajectory}

\providecolor{baselineRow}{HTML}{F2F2F2}
\providecolor{discardRow}{HTML}{FBEAEA}
\providecolor{bestRow}{HTML}{E6F4EA}
\begin{table}[htbp]
\centering
\small
\caption{\small{
Complete trajectory of the basic agent on single-cell perturbation-response prediction (Norman19,
GEARS), all 15 iterations; Table~\ref{tab:basic-trace} shows an excerpt. Scores are validation cosine
similarity between predicted and observed log fold changes ($\times$100). A perturbation is the set of
cells that received the same genetic perturbation, and control cells are unperturbed. Each row lists
what the agent changed from the previous iteration and a paraphrase of the note it wrote after seeing
the score. At iteration 10 the agent intended to keep the 1{,}000 control cells with the highest
counts, but a column-name bug left the policy identical to iteration 6.
}}
\label{tab:basic-trace-full}
\vspace{-0.5em}
\resizebox{\linewidth}{!}{%
\renewcommand{\arraystretch}{1.15}
\begin{tabular}{>{\centering\arraybackslash}p{1.3cm}>{\centering\arraybackslash}p{1.0cm}>{\centering\arraybackslash}p{1.1cm}p{5.0cm}p{6.6cm}}
\toprule
\textbf{Iteration} & \textbf{Score} & \textbf{Status} & \textbf{What changed from the previous iteration}
& \textbf{Agent's note (paraphrased)} \\
\midrule
\rowcolor{baselineRow}
1 & 61.30 & keep & Random 50\% subset (baseline) & The score to beat. \\
\rowcolor{discardRow}
2 & 57.58 & discard & All $\sim$9{,}700 control cells; the same number of cells for every perturbation;
within each perturbation, the cells with the highest counts
& Worse than baseline. At least one of the three changes hurt, most likely picking the
highest-count cells, which may be doublets. \\
\rowcolor{discardRow}
3 & 60.83 & discard & Controls back to $\sim$4{,}900; random cells within each perturbation again
& Still slightly below baseline. Picking the highest-count cells was the problem; the same number
per perturbation is also slightly worse than splitting by size. \\
4 & 63.63 & keep & Give each perturbation at least 100 cells (all of its cells if it has fewer);
split the rest by perturbation size
& New best: small perturbations now keep all their cells, and large ones keep their share. \\
5 & 63.83 & keep & Control cells: $\sim$4{,}900 $\to$ 2{,}000
& New best: cutting controls frees cells for perturbations at little cost. \\
6 & 67.53 & keep & Control cells: 2{,}000 $\to$ 1{,}000
& Large gain: more perturbation cells matter more than more controls. \\
\rowcolor{discardRow}
7 & 66.60 & discard & Control cells: 1{,}000 $\to$ 500
& Worse: 500 controls are too few, so 1{,}000 is the lower limit. Next, keep 1{,}000 controls and
try a higher per-perturbation minimum. \\
\rowcolor{discardRow}
8 & 65.32 & discard & Control cells back to 1{,}000; per-perturbation minimum: 100 $\to$ 150
& Worse: a higher minimum takes cells from large perturbations, so 100 is the right minimum. \\
\rowcolor{discardRow}
9 & 65.30 & discard & Minimum back to 100; split the rest by the square root of perturbation size
instead of by size
& Worse: this helps mid-sized perturbations but hurts large ones, so splitting by size is optimal. \\
\rowcolor{discardRow}
10 & 67.53 & discard & Split by size again; intended to keep the 1{,}000 control cells with the highest
counts, but a bug left the policy identical to iteration 6
& Identical to iteration 6 because of the bug, so no new information. \\
\rowcolor{bestRow}
11 & \textbf{77.23} & keep & Keep the 1{,}000 control cells with the most detected genes instead of
random ones
& Large gain: controls with more detected genes give a richer gene co-expression graph. \\
\rowcolor{discardRow}
12 & 75.14 & discard & Control cells: 1{,}000 $\to$ 2{,}000, still those with the most detected genes
& Worse: more controls take cells from perturbations, so 1{,}000 is the sweet spot. \\
\rowcolor{discardRow}
13 & 75.08 & discard & Controls back to 1{,}000; also pick the perturbation cells with the most detected
genes
& Worse: picking perturbation cells by quality hurts; random selection is better. \\
\rowcolor{discardRow}
14 & 77.11 & discard & Random perturbation cells again; control cells: 1{,}000 $\to$ 750
& Slightly worse: 1{,}000 controls is the sweet spot, not 750. \\
\rowcolor{discardRow}
15 & 76.57 & discard & Controls back to 1{,}000; fill the smallest perturbations first instead of the
at-least-100 rule
& Worse: at least 100 per perturbation, with the rest split by size, remains the best allocation. \\
\bottomrule
\end{tabular}%
}
\vspace{-1em}
\end{table}

\subsection{Agent Trajectory with \alg}
\providecolor{baselineRow}{HTML}{F2F2F2}
\providecolor{discardRow}{HTML}{FBEAEA}
\providecolor{bestRow}{HTML}{E6F4EA}
\begin{table}[htbp]
\centering
\small
\caption{\small{
Complete trajectory of \alg (Claude Opus~4.8) on single-cell cell type annotation (PBMC
cross-protocol, SVM), all 15 iterations. Scores are validation macro-F1 ($\times$100), the F1 score
averaged over five cell types: cytotoxic T cells, CD4$^+$ T cells, CD14$^+$ monocytes, B cells, and
megakaryocytes. The training pool has 8{,}165 cells from two 10x protocols, and the
validation cells come from a third protocol. Each iteration keeps 4{,}082 cells and can also choose
the genes the model uses; the default is a panel of 2{,}000 highly variable genes. The SVM already
weights each class inversely to its frequency. Each row lists what the agent changed from the
previous iteration and a paraphrase of the note it wrote after seeing the score. Iterations 5 and 10 (bold)
change the gene panel in opposite directions, and the agent's memory joins the two into one claim:
the 2{,}000-gene panel is a sweet spot.
}}
\label{tab:memory-trace-full}
\vspace{-0.5em}
\resizebox{\linewidth}{!}{%
\renewcommand{\arraystretch}{1.15}
\begin{tabular}{>{\centering\arraybackslash}p{1.3cm}>{\centering\arraybackslash}p{1.0cm}>{\centering\arraybackslash}p{1.1cm}p{5.0cm}p{6.6cm}}
\toprule
\textbf{Iteration} & \textbf{Score} & \textbf{Status} & \textbf{What changed from the previous iteration}
& \textbf{Agent's note (paraphrased)} \\
\midrule
\rowcolor{baselineRow}
1 & 84.92 & keep & Random 50\% subset (baseline)
& The score to beat. One cell type has a recall of only 0.52 and drags the average down. \\
\rowcolor{discardRow}
2 & 84.42 & discard & The same number of cells for every cell type, capping the large ones
& Worse: the weakest cell type improves, but the large ones lose cells. The SVM already weights
classes by frequency, so balancing the counts is redundant. \\
3 & 89.84 & keep & No count balancing; keep the 4{,}082 cells whose labels a cross-validated
logistic regression predicts most confidently
& Large gain: dropping ambiguous or mislabeled cells sharpens the linear boundary, and the weakest
cell type's recall rises from 0.52 to 0.79. \\
\rowcolor{discardRow}
4 & 88.35 & discard & Score cells with a classifier trained on the other 10x protocol instead of
cross-validation
& Worse: a scorer trained on one protocol is less reliable than cross-validation over the whole
pool. \\
\rowcolor{discardRow}
\textbf{5} & \textbf{88.96} & \textbf{discard} & \textbf{Back to the iteration-3 cells; drop the
sparsest genes (2{,}000 $\to$ 1{,}684 genes)}
& \textbf{Worse: removing genes hurts; the SVM needs the full 2{,}000-gene panel.} \\
\rowcolor{discardRow}
6 & 89.17 & discard & All 2{,}000 genes again; score cells with the evaluation model itself (the
SVM) instead of logistic regression
& Worse: the SVM scorer removes too many B cells and megakaryocytes; the gentler
logistic-regression scorer stays best. \\
\rowcolor{discardRow}
7 & 89.50 & discard & Logistic-regression scorer again, averaged over five random seeds
& No gain: averaging over seeds does not improve the selection, so iteration 3 stays best. \\
\rowcolor{discardRow}
8 & 78.33 & discard & One seed again; keep every monocyte, B cell and megakaryocyte, and fill the
rest with the most confident T cells
& Collapse: only 1{,}423 T cells are left, and the T cells near the class boundary were
informative. The $\sim$2{,}100 T cells of iteration 3 are the right amount. \\
\rowcolor{discardRow}
9 & 58.50 & discard & Back to ranking all cells together; score each cell by how many of its 30
nearest neighbors share its label
& Collapse: the two T-cell types overlap, so cytotoxic T cells score low and only the required
minimum of 3 is kept; the model never predicts them. \\
\rowcolor{discardRow}
\textbf{10} & \textbf{85.85} & \textbf{discard} & \textbf{Back to the iteration-3 cells; use all
2{,}191 genes instead of the 2{,}000-gene panel}
& \textbf{Worse: the 191 extra genes vary little across cells and mostly carry protocol
differences. With iteration 5, both fewer and more genes hurt, so the 2{,}000-gene panel is the
sweet spot.} \\
\rowcolor{discardRow}
11 & 89.77 & discard & 2{,}000-gene panel again; a more strongly regularized logistic-regression
scorer ($C$: 1 $\to$ 0.3)
& About the same as iteration 3: the scorer's regularization makes little difference. \\
\rowcolor{discardRow}
12 & 85.23 & discard & Scorer back to $C=1$; within each cell type, cluster the cells and keep the
most confident cell of each cluster, at iteration 3's counts
& Worse: spreading the subset out brings back ambiguous boundary cells, which blur the linear
boundary. Keeping the cleanest cells wins. \\
13 & 90.45 & keep & Back to ranking all cells together; score in two rounds: refit the scorer on the 75\% most
trusted cells, then rescore all cells
& New best: the second round ranks the borderline T cells better, and the weakest cell type
improves again. \\
14 & 90.81 & keep & Second-round scorer refit on the 60\% most trusted cells instead of 75\%
& New best: a cleaner training set for the second round helps further. \\
\rowcolor{bestRow}
15 & \textbf{91.27} & keep & Second-round scorer refit on the 45\% most trusted cells instead of
60\%
& New best: the score keeps rising as the second round trusts fewer, cleaner cells (90.45, 90.81,
91.27). \\
\bottomrule
\end{tabular}%
}
\vspace{-1em}
\end{table}

\subsection{Memory Files}\label{app:memfiles}

Because the agent's raw memory is accumulated incrementally and can be fragmented, we lightly reorganize it for readability while preserving its content. We also manually separate the memory into \emph{Evidence} and \emph{Claims} to make the distinction explicit.

\begin{tcolorbox}[protobox, title={Agent memory before iteration 12: cell type annotation (PBMC cross-protocol, SVM)}]
\small
\textbf{Evidence: facts.}
Task: cross-protocol PBMC. Train = 10x v2 (5435) + v3 (2730). Val = inDrops (hidden).\\
5 classes: Cytotoxic T 3090, CD4+ T 2416, CD14+ monocyte 1365, B 1022, Megakaryocyte 272.\\
Model = SVM on log1p(CP10k), features = curated mask (default = 2000 HVG). Headline macro\_f1.

\textbf{Evidence: experiments.}
\begin{itemize}\setlength{\itemsep}{0pt}\setlength{\parskip}{0pt}
\item iter1 $\cdot$ seed random 4082 $\cdot$ macro\_f1 0.8492, min\_recall 0.520 $\cdot$ baseline.
\item iter2 $\cdot$ water-fill class balance $\cdot$ macro\_f1 0.8442 REVERT (harmful; model already class\_weight=balanced).
\item iter3 $\cdot$ clean coreset (top-conf 4082) $\cdot$ macro\_f1 0.8984 KEEP (held, big jump).
\item iter4 $\cdot$ cross-protocol scorer $\cdot$ macro\_f1 0.8835 REVERT (worse than within-sample).
\item iter5 $\cdot$ coreset + prevalence mask (1684g) $\cdot$ macro\_f1 0.8896 REVERT (feature reduction hurts).
\item iter6 $\cdot$ coreset SVM-matched scorer $\cdot$ macro\_f1 0.8917 REVERT (over-trims easy classes).
\item iter7 $\cdot$ multi-seed avg logreg OOF $\cdot$ macro\_f1 0.8950 REVERT (no gain).
\item iter8 $\cdot$ support-aware forced allocation $\cdot$ macro\_f1 0.7833 REVERT (collapse; over-thinned T).
\item iter9 $\cdot$ kNN local purity scorer $\cdot$ macro\_f1 0.5850 REVERT (collapse; kNN purity class-confounded by T-overlap, Cyto$\to$3).
\item iter10 $\cdot$ full 2191-gene mask $\cdot$ macro\_f1 0.8585 REVERT (feature expansion hurts; HVG optimal).
\item iter11 $\cdot$ logreg scorer C=0.3 $\cdot$ macro\_f1 0.8977 REVERT (flat/tied).
\end{itemize}

\textbf{Claims.}
MECHANISM MAP (COMPLETE): coreset cleanliness (logreg-OOF global top-conf) = the ONLY lever, +0.049 over random. All other axes mapped and closed:
\begin{itemize}\setlength{\itemsep}{0pt}\setlength{\parskip}{0pt}
\item class balance --- redundant w/ SVM class\_weight=balanced $\to$ harmful (iter2)
\item scorer: logreg-OOF best; cross-protocol worse (iter4), calibrated-LinearSVC worse (iter6), kNN class-confounded$\to$collapse (iter9), C=0.3 flat (iter11). Principle: LINEAR, class-aware, comparably-scaled posterior matched to the linear eval SVM.
\item allocation: GLOBAL threshold optimal; multi-seed flat (iter7), forced per-class collapses (iter8). T-support $\sim$2133 is a sharp optimum.
\item \textbf{feature mask: HVG(2000) is a sweet spot --- reduction hurts (iter5 $-0.009$), expansion hurts (iter10 $-0.040$).}
\end{itemize}

\end{tcolorbox}

\section{Additional Empirical Results}\label{app:res}

Table~\ref{tab:main-bio} presents the main results on the biology domains.
Each entry is the mean, over the dataset--model pairs of the domain,
of the best headline score reached within the 15-iteration budget, scaled to 0--100.
Random is the seeded subset every run starts from,
Full data trains on the entire pool without curation,
and the fixed-pipeline rows are standard curation recipes of each domain:
the mean $\pm$ sd over the 12 recipes, and the best recipe chosen separately for every subtask.
Across all five domains, the agents outperform Full data and the fixed curation pipelines.
\alg further improves over the corresponding Basic agent in 29 of 30 model--domain comparisons.

\begin{table*}[htbp]
\centering
\caption{Main results on the biology domains.
Each entry is the mean, over the dataset--model pairs of the domain,
of the best headline score reached within the 15-iteration budget, scaled to 0--100.
Random is the seeded subset every run starts from,
Full data trains on the entire pool without curation,
and the fixed-pipeline rows are standard curation recipes of each domain:
the mean $\pm$ sd over the 12 recipes, and the best recipe chosen separately for every subtask.
Percentages (\%) are relative change against Full data; entries more than 5\% above Full data
are in \textcolor{tabGain}{green} and entries more than 5\% below it in \textcolor{tabLoss}{red}.
Within each block, the best entry in each column is in \textbf{bold}.
}
\label{tab:main-bio}

\vspace{2pt}
\footnotesize
\setlength{\tabcolsep}{4pt}
\renewcommand{\arraystretch}{1.18}

\resizebox{\ifdim\width>\textwidth\textwidth\else\width\fi}{!}{%
\begin{NiceTabular}{@{}l l ccccc@{}}

\toprule

\rowcolor{tabHeader}
\textbf{Model}
&
\textbf{Agent variant}
&
\Block{}{Protein\\(Spearman)}
&
\Block{}{Non-coding\\(AUPRC)}
&
\Block{}{Cell type\\(macro-F1)}
&
\Block{}{Batch\\(scIB overall)}
&
\Block{}{Perturbation\\(cosine)}
\\

\midrule

\rowcolor{tabRef}
\Block[l]{1-2}{Full data \textit{(baseline)}}
& -
&
35.60
&
13.38
&
64.65
&
68.79
&
46.04
\\

\Block[l]{1-2}{Random 50\%}
& -
&
34.20\,{\tiny$\pm$\,0.65}
&
12.86\,{\tiny$\pm$\,0.30}
&
63.86\,{\tiny$\pm$\,0.81}
&
68.45\,{\tiny$\pm$\,0.15}
&
45.95\,{\tiny$\pm$\,0.41}
\\

\rowcolor{tabRef}
\Block[l]{1-2}{Fixed pipeline 50\%\\{\scriptsize\itshape \textbf{avg.}\ of 12 methods}}
& -
&
\tabentry{32.64}{2.68}{-}{8.3}
&
\tabentry{12.93}{0.59}{-}{3.4}
&
\tabentry{63.82}{1.60}{-}{1.3}
&
\tabentry{68.21}{0.56}{-}{0.8}
&
\tabentry{44.64}{3.45}{-}{3.0}
\\

\rowcolor{tabRef}
\Block[l]{1-2}{Fixed pipeline 50\%\\{\scriptsize\itshape \textbf{best} method per subtask}}
& -
&
\textbf{\tabentry{36.79}{}{+}{3.3}}
&
\textbf{\tabentry{13.73}{}{+}{2.6}}
&
\textbf{\tabentry{66.18}{}{+}{2.4}}
&
\textbf{\tabentry{69.25}{}{+}{0.7}}
&
\textbf{\tabentry{48.27}{}{+}{4.8}}
\\

\midrule

\rowcolor{tabClaude}
\Block[l]{1-7}{\textit{Claude Code - select 50\%}}
&
&
&
&
&
&
\\

\rowcolor{tabClaude}
\Block[l]{2-1}{Opus 4.8}
&
Basic agent
&
\tabentry{38.61}{}{+}{8.5}
&
\tabentry{14.29}{}{+}{6.8}
&
\tabentry{69.46}{}{+}{7.4}
&
\tabentry{69.53}{}{+}{1.1}
&
\tabentry{49.67}{}{+}{7.9}
\\

\rowcolor{tabClaude}
&
\alg
&
\textbf{\tabentry{39.59}{}{+}{11.2}}
&
\textbf{\tabentry{15.41}{}{+}{15.2}}
&
\textbf{\tabentry{72.30}{}{+}{11.8}}
&
\textbf{\tabentry{72.89}{}{+}{6.0}}
&
\tabentry{51.00}{}{+}{10.8}
\\

\rowcolor{tabClaude}
\Block[l]{2-1}{Sonnet 5}
&
Basic agent
&
\tabentry{37.52}{}{+}{5.4}
&
\tabentry{14.36}{}{+}{7.3}
&
\tabentry{68.39}{}{+}{5.8}
&
\tabentry{69.67}{}{+}{1.3}
&
\tabentry{49.24}{}{+}{7.0}
\\

\rowcolor{tabClaude}
&
\alg
&
\tabentry{38.75}{}{+}{8.8}
&
\tabentry{15.08}{}{+}{12.7}
&
\tabentry{71.04}{}{+}{9.9}
&
\tabentry{71.55}{}{+}{4.0}
&
\textbf{\tabentry{51.07}{}{+}{10.9}}
\\

\rowcolor{tabClaude}
\Block[l]{2-1}{Haiku 4.5}
&
Basic agent
&
\tabentry{37.69}{}{+}{5.9}
&
\tabentry{14.17}{}{+}{5.9}
&
\tabentry{67.81}{}{+}{4.9}
&
\tabentry{69.63}{}{+}{1.2}
&
\tabentry{48.86}{}{+}{6.1}
\\

\rowcolor{tabClaude}
&
\alg
&
\tabentry{38.57}{}{+}{8.3}
&
\tabentry{15.05}{}{+}{12.5}
&
\tabentry{67.59}{}{+}{4.5}
&
\tabentry{70.46}{}{+}{2.4}
&
\tabentry{50.34}{}{+}{9.3}
\\

\midrule

\rowcolor{tabCodex}
\Block[l]{1-7}{\textit{Codex}}
&
&
&
&
&
&
\\

\rowcolor{tabCodex}
\Block[l]{2-1}{GPT-5.3-Codex}
&
Basic agent
&
\tabentry{37.44}{}{+}{5.2}
&
\tabentry{14.19}{}{+}{6.1}
&
\tabentry{68.13}{}{+}{5.4}
&
\tabentry{69.25}{}{+}{0.7}
&
\tabentry{49.22}{}{+}{6.9}
\\

\rowcolor{tabCodex}
&
\alg
&
\tabentry{38.81}{}{+}{9.0}
&
\tabentry{15.22}{}{+}{13.8}
&
\tabentry{70.47}{}{+}{9.0}
&
\tabentry{71.21}{}{+}{3.5}
&
\tabentry{50.85}{}{+}{10.4}
\\

\rowcolor{tabCodex}
\Block[l]{2-1}{GPT-5.6 Sol}
&
Basic agent
&
\tabentry{38.39}{}{+}{7.8}
&
\tabentry{14.32}{}{+}{7.0}
&
\tabentry{68.97}{}{+}{6.7}
&
\tabentry{69.49}{}{+}{1.0}
&
\tabentry{49.59}{}{+}{7.7}
\\

\rowcolor{tabCodex}
&
\alg
&
\textbf{\tabentry{39.66}{}{+}{11.4}}
&
\textbf{\tabentry{15.97}{}{+}{19.4}}
&
\textbf{\tabentry{73.02}{}{+}{12.9}}
&
\textbf{\tabentry{72.68}{}{+}{5.7}}
&
\textbf{\tabentry{51.52}{}{+}{11.9}}
\\

\midrule

\rowcolor{tabOpen}
\Block[l]{1-7}{\textit{OpenHands}}
&
&
&
&
&
&
\\

\rowcolor{tabOpen}
\Block[l]{2-1}{Qwen3.5-397B-A17B}
&
Basic agent
&
\tabentry{37.26}{}{+}{4.7}
&
\tabentry{14.48}{}{+}{8.2}
&
\tabentry{66.46}{}{+}{2.8}
&
\tabentry{69.53}{}{+}{1.1}
&
\tabentry{48.78}{}{+}{6.0}
\\

\rowcolor{tabOpen}
&
\alg
&
\textbf{\tabentry{38.37}{}{+}{7.8}}
&
\textbf{\tabentry{15.26}{}{+}{14.1}}
&
\textbf{\tabentry{69.81}{}{+}{8.0}}
&
\textbf{\tabentry{70.28}{}{+}{2.2}}
&
\textbf{\tabentry{50.67}{}{+}{10.1}}
\\

\bottomrule

\end{NiceTabular}
}
\end{table*}

Table~\ref{tab:main-llm}
 presents 
main results on LLM post-training tasks. "Random", "LESS", "DELIFT", and "Agent variant" all select training data of the same size from the same pool. "Random" is a randomly downsampled subset, LESS~\citep{xia2024less} and DELIFT~\citep{agarwal2025delift} are human-designed selection methods, whereas "Entire dataset" trains on the entire pool without curation. Each agent entry is the best mean accuracy over the setup's benchmark suite reached within the 10-iteration budget.  
\begin{table}[t]
\centering
\caption{\small{Main results on LLM post-training tasks. "Random", "LESS", "DELIFT", and "Agent variant" all select training data of the same size from the same pool. "Random" is a randomly downsampled subset, LESS~\citep{xia2024less} and DELIFT~\citep{agarwal2025delift} are human-designed selection methods, whereas "Entire dataset" trains on the entire pool without curation. Each agent entry is the best mean accuracy over the setup's benchmark suite reached within the 10-iteration budget. Percentages (\%) are relative change against Random; entries more than 5\% above Full data
are in \textcolor{tabGain}{green} and entries more than 5\% below it in \textcolor{tabLoss}{red}.
}}
\label{tab:main-llm}

\vspace{2pt}
\footnotesize
\setlength{\tabcolsep}{4pt}
\renewcommand{\arraystretch}{1.18}

\resizebox{0.8\linewidth}{!}{%
\begin{NiceTabular}{@{}l l ccc@{}}

\toprule

\rowcolor{tabHeader}
\textbf{Model} & \textbf{Agent variant}
& \Block{}{OLMo-2-1B\\(select 1.1\%)}
& \Block{}{Llama-2-7B\\(select 3.7\%)}
& \Block{}{Qwen2.5-Coder-1.5B\\(select 0.2\%)} \\

\midrule

\rowcolor{tabRef}
\Block[l]{1-2}{Random \textit{(baseline)}} & -
& 31.00\,{\tiny$\pm$\,0.52}
& 42.44\,{\tiny$\pm$\,0.28}
& 57.72\,{\tiny$\pm$\,0.24} \\

\Block[l]{1-2}{Entire dataset} & -
& 30.35 & \textbf{45.70} & \textbf{62.03} \\

\rowcolor{tabRef}
\Block[l]{1-2}{LESS~\citep{xia2024less}} & -
& \textbf{\tabentry{31.81}{}{+}{2.6}}
& \tabentry{45.05}{}{+}{6.1}
& \tabentry{58.38}{}{+}{1.1} \\

\rowcolor{tabRef}
\Block[l]{1-2}{DELIFT~\citep{agarwal2025delift}} & -
& \tabentry{31.54}{}{+}{1.7}
& \tabentry{43.69}{}{+}{2.9}
& \tabentry{58.57}{}{+}{1.5} \\
\midrule

\rowcolor{tabClaude}
\Block[l]{1-5}{\textit{Claude Code}} & & & & \\

\rowcolor{tabClaude}
\Block[l]{2-1}{Opus 4.8} & Basic agent
& \tabentry{31.92}{}{+}{3.0}
& \tabentry{44.51}{}{+}{4.9}
& \tabentry{58.15}{}{+}{0.7} \\

\rowcolor{tabClaude}
& \alg
& \textbf{\tabentry{32.86}{}{+}{6.0}}
& \textbf{\tabentry{46.07}{}{+}{8.6}}
& \textbf{\tabentry{59.72}{}{+}{3.5}} \\

\rowcolor{tabClaude}
\Block[l]{2-1}{Sonnet 5} & Basic agent
& \tabentry{31.84}{}{+}{2.7}
& \tabentry{44.42}{}{+}{4.7}
& \tabentry{58.13}{}{+}{0.7} \\

\rowcolor{tabClaude}
& \alg
& \textcolor{tabGain}{\tabentry{32.56}{}{+}{5.0}}
& \tabentry{45.78}{}{+}{7.9}
& \tabentry{59.14}{}{+}{2.5} \\

\rowcolor{tabClaude}
\Block[l]{2-1}{Haiku 4.5} & Basic agent
& \tabentry{31.15}{}{+}{0.5}
& \tabentry{44.86}{}{+}{5.7}
& \tabentry{58.24}{}{+}{0.9} \\

\rowcolor{tabClaude}
& \alg
& \tabentry{31.82}{}{+}{2.6}
& \tabentry{45.12}{}{+}{6.3}
& \tabentry{59.23}{}{+}{2.6} \\

\midrule

\rowcolor{tabCodex}
\Block[l]{1-5}{\textit{Codex}} & & & & \\

\rowcolor{tabCodex}
\Block[l]{2-1}{GPT-5.6 Sol} & Basic agent
& \tabentry{32.17}{}{+}{3.8}
& \tabentry{44.72}{}{+}{5.4}
& \tabentry{58.57}{}{+}{1.5} \\

\rowcolor{tabCodex}
& \alg
& \textbf{\tabentry{33.14}{}{+}{6.9}}
& \textbf{\tabentry{45.85}{}{+}{8.0}}
& \textbf{\tabentry{59.71}{}{+}{3.4}} \\

\rowcolor{tabCodex}
\Block[l]{2-1}{GPT-5.3-Codex} & Basic agent
& \tabentry{31.20}{}{+}{0.6}
& \tabentry{44.60}{}{+}{5.1}
& \tabentry{58.11}{}{+}{0.7} \\

\rowcolor{tabCodex}
& \alg
& \tabentry{32.04}{}{+}{3.4}
& \tabentry{45.31}{}{+}{6.8}
& \tabentry{58.89}{}{+}{2.0} \\

\midrule

\rowcolor{tabOpen}
\Block[l]{1-5}{\textit{OpenHands}} & & & & \\

\rowcolor{tabOpen}
\Block[l]{2-1}{Qwen3.5-397B-A17B} & Basic agent
& \tabentry{30.77}{}{-}{0.7}
& \tabentry{44.13}{}{+}{4.0}
& \tabentry{58.12}{}{+}{0.7} \\

\rowcolor{tabOpen}
& \alg
& \textbf{\tabentry{31.48}{}{+}{1.5}}
& \textbf{\tabentry{45.54}{}{+}{7.3}}
& \textbf{\tabentry{59.05}{}{+}{2.3}} \\

\bottomrule

\end{NiceTabular}
}
\end{table}

\begin{figure}[htbp]
\centering
\includegraphics[width=\linewidth]{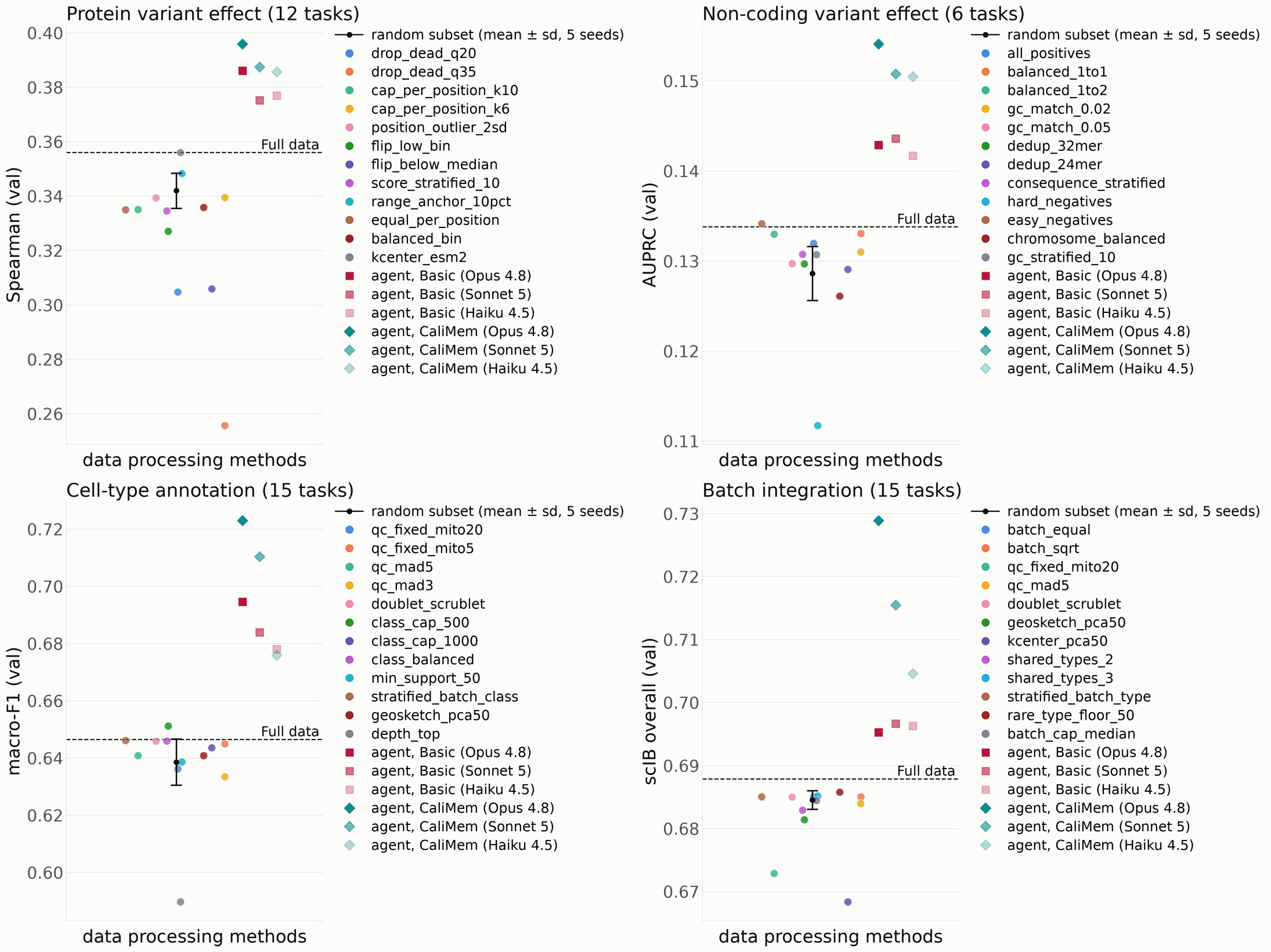}\vspace{-0.0em}
\caption{\small{Selecting 50\% of the training data for machine-learning models in biology tasks. In biology, datasets are typically cleaned with standard, human-designed processing pipelines, and the processed data are then used to train different downstream models, which often make different assumptions about their data. A fixed cleaning protocol can therefore have very different effects across use cases: in every task, the pipelines that underperform training on unprocessed data (a random 50\% subset) outnumber those that outperform it. Agentic data curation instead optimizes the training data directly for the target model pipeline, giving large performance gains through test-time scaling and outperforming fixed pipelines. \alg further strengthens the data agent and consistently outperforms the default agent. }}\vspace{-0.5em}
\label{fig:5}
\end{figure}

\begin{figure}[htbp]
  \vspace{-12pt}
  \centering
  \includegraphics[width=0.6\linewidth]{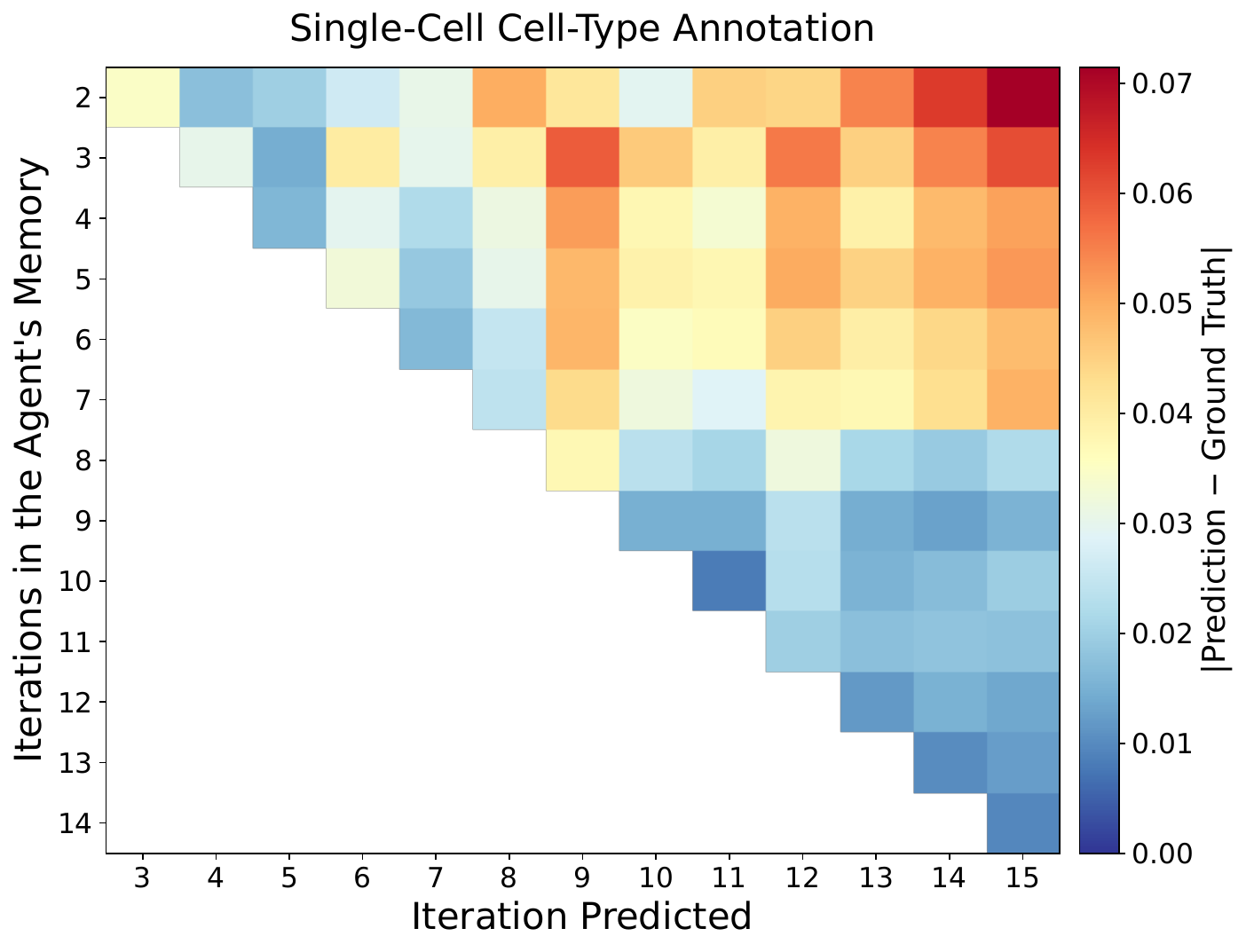}
   \caption{\small{\textbf{The memory becomes more predictive as evidence accumulates.}
  Absolute error when a fresh agent uses the \alg memory frozen after iteration $k$
  (rows) to predict a later, unseen iteration (columns).
  Results are averaged over single-cell cell-type annotation tasks from Claude
  Sonnet~5 runs; lighter cells indicate smaller error.
  A fixed memory becomes less accurate as it extrapolates further ahead, whereas
  predictions of the same target iteration improve when the memory is fitted on more
  preceding experiments.}}
  \label{fig:matrix}
\end{figure}

\section{Forecasting held-out iterations from the \alg}
\label{app:exam}

\subsection{Extended Experiment Results}
\label{app:exam-more}
\begin{figure}[htbp]
  \centering
  \begin{subfigure}[t]{0.32\linewidth}
    \centering
    \includegraphics[width=\linewidth]{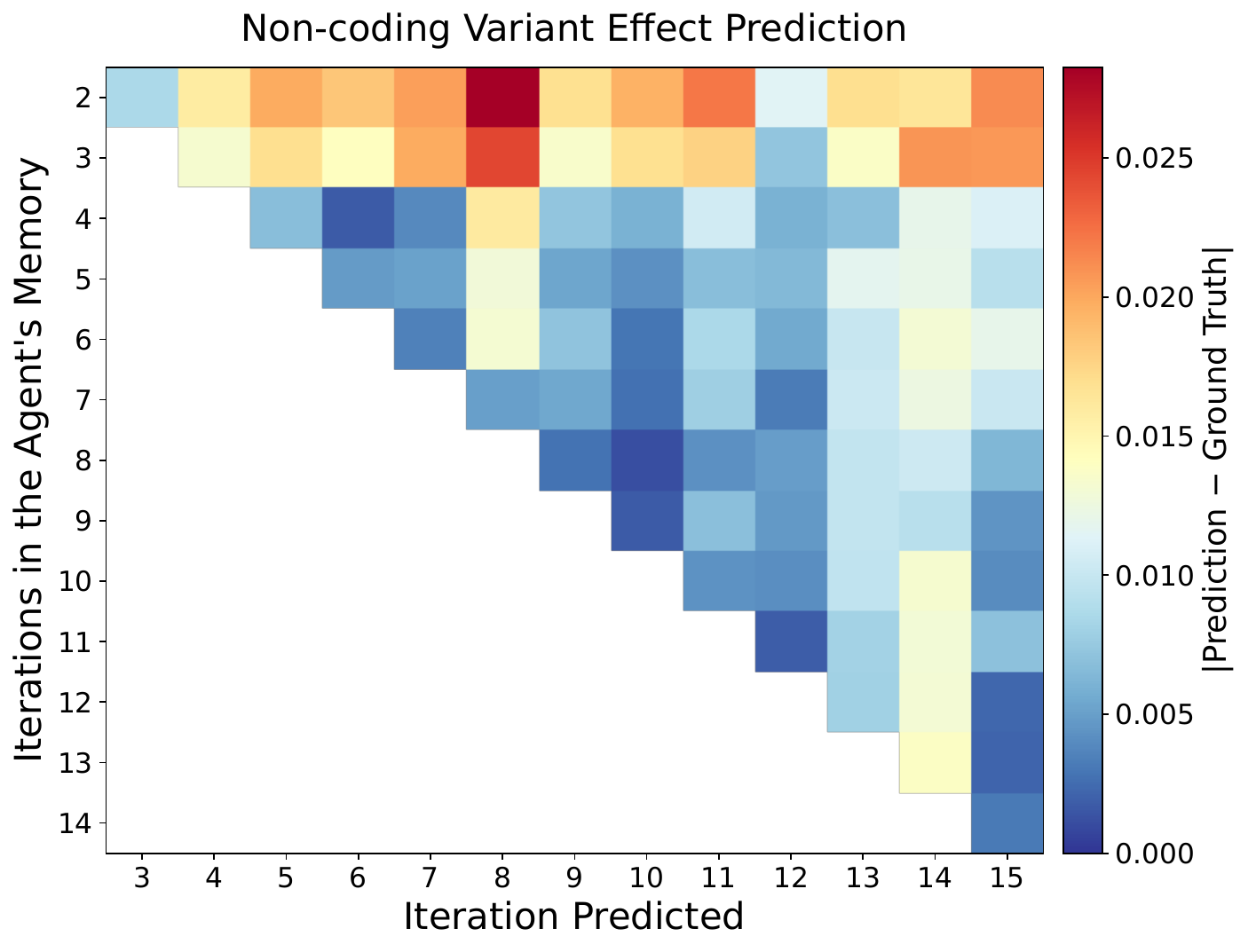}
    \caption{Non-coding variant effect, Claude Opus~4.8 runs.}
    \label{fig:matrix-opus-ncvar}
  \end{subfigure}\hfill
  \begin{subfigure}[t]{0.32\linewidth}
    \centering
    \includegraphics[width=\linewidth]{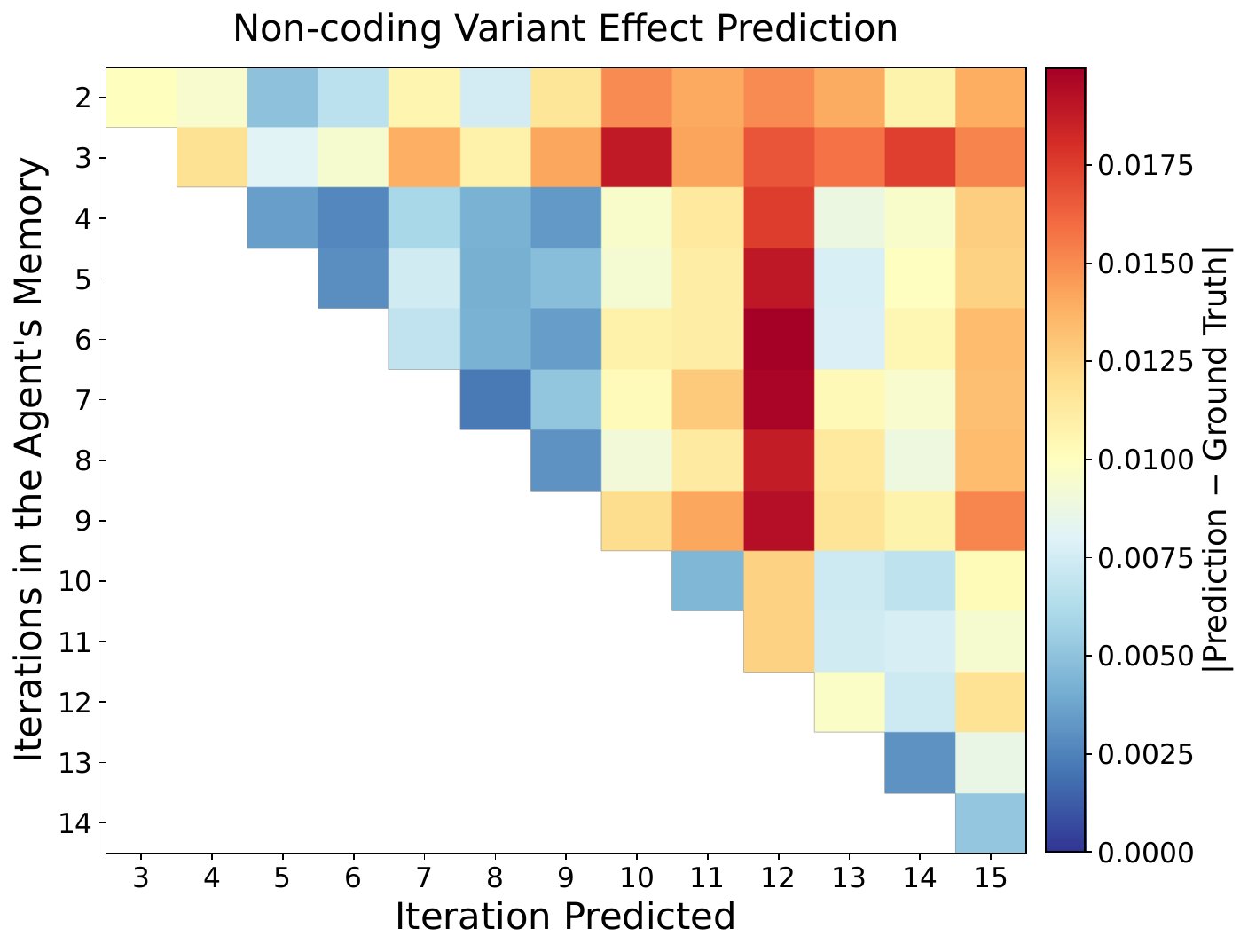}
    \caption{Non-coding variant effect, Claude Sonnet~5 runs.}
    \label{fig:matrix-sonnet-ncvar}
  \end{subfigure}\hfill
  \begin{subfigure}[t]{0.32\linewidth}
    \centering
    \includegraphics[width=\linewidth]{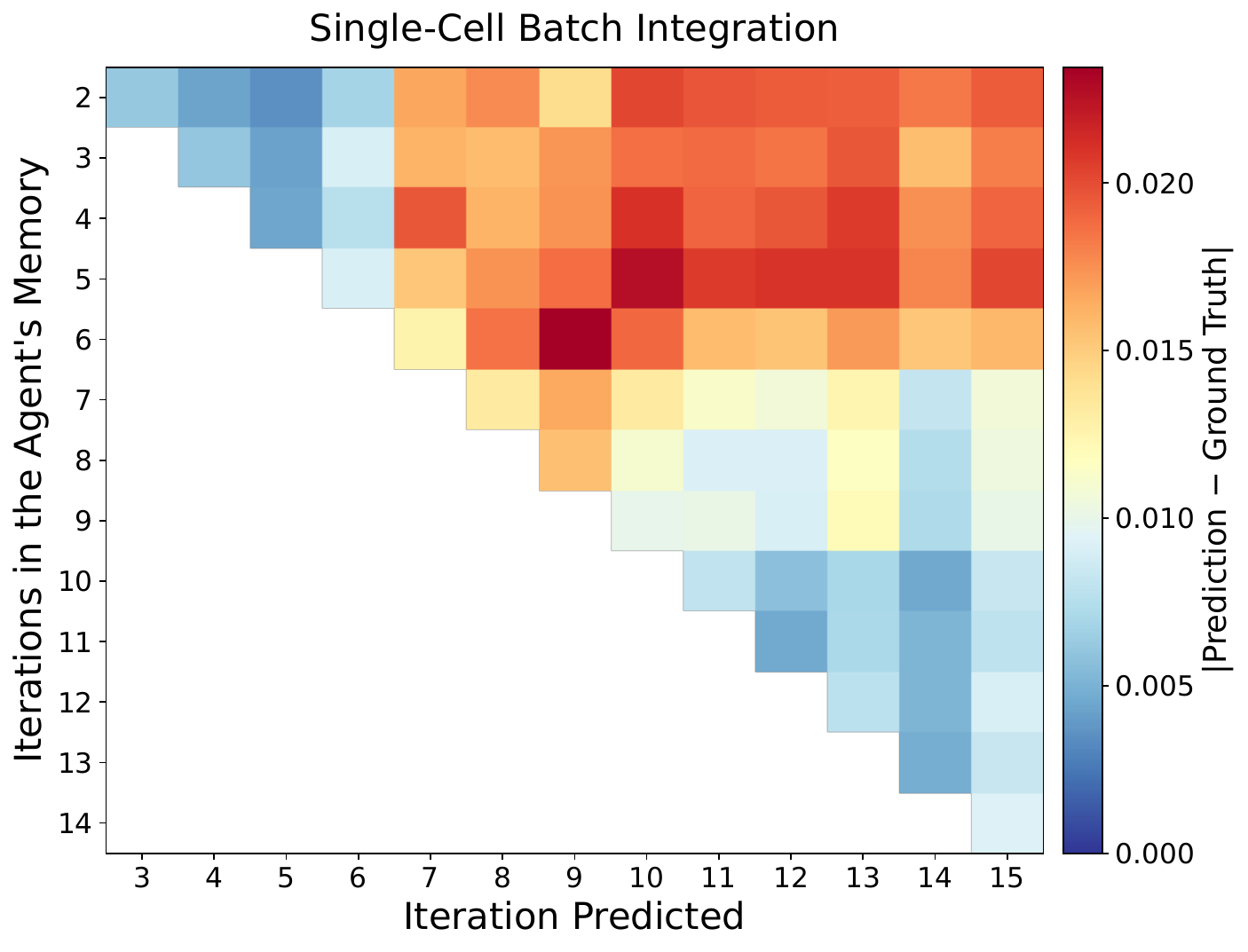}
    \caption{Batch integration, Claude Sonnet~5 runs.}
    \label{fig:matrix-sonnet-scbatch}
  \end{subfigure}
  \caption{Forecasting error of a \alg fitted on the first $k$ iterations (rows) when a fresh agent uses it to predict the score of every later iteration (columns), for the three domains not shown in the main text. Each entry is the absolute error in the domain's headline metric, averaged over all tasks of the domain; lighter is smaller error.}
  \label{fig:matrix-more}
\end{figure}

The forecasting analysis of Section~\ref{sec:forecast} is conducted in four experimental settings.
Figure~\ref{fig:matrix} shows single-cell cell-type annotation using trajectories generated by
Claude Sonnet~5, while Figure~\ref{fig:matrix-more} reports the remaining three settings:
non-coding variant-effect prediction using trajectories generated by Claude Opus~4.8,
non-coding variant-effect prediction using trajectories generated by Claude Sonnet~5,
and single-cell batch integration using trajectories generated by Claude Sonnet~5.
All four settings follow the same evaluation protocol: after iteration $k$, we freeze the
\alg{} memory and ask a fresh agent to predict the scores of policies executed later in the
original trajectory without observing their outcomes.
Each matrix contains one row per cut point $k$ and one column per target iteration, and each
entry is the absolute prediction error averaged over the dataset--model pairs in that setting.
The forecasting agent is the same across all settings
(Appendix~\ref{app:exam-prompt}).

Batch integration exhibits the same two patterns as cell-type annotation.
For a fixed memory, prediction error generally increases with extrapolation distance along
a row; for a fixed target iteration, the error generally decreases down a column as the
memory incorporates more preceding experiments.
The two non-coding variant-effect settings are noisier because they average over fewer
dataset--model pairs, making individual volatile trajectories more influential.
Nevertheless, the increase in error with extrapolation distance remains visible, while the
improvement from longer memory prefixes is weaker.


\subsection{Prompt}
\label{app:exam-prompt}

The agent taking the exam is Claude Opus~4.8 for every matrix, including those built from Claude Sonnet~5 runs, so a memory written by one model is read by another. It answers in a single turn, one sample per prompt. Besides the memory and the strategies, the prompt gives the task goal, a few fixed facts about the cell, and the score of iteration 1 as a scale anchor; a strategy that refers to an earlier iteration by number has that iteration's strategy appended in brackets. The template is condensed below, with the parts filled in per run and per cut point in italics.

\begin{tcblisting}{protobox, listing only, title={Prompt template, cut point $k$},
  listing options={basicstyle=\ttfamily\footnotesize, breaklines=true, columns=fullflexible, keepspaces=true, escapeinside={(*}{*)}}}
Predict what an automated data-curation experiment will score. You did not design these changes; you only forecast them.

(*\textit{task goal}*)

Headline metric `(*\textit{metric}*)`, higher is better; an iteration below the best so far is normally reverted.

Facts:
(*\textit{dataset, frozen model, budget, and the score of iteration 1}*)

## The agent's notes after (*$k$*) scored iterations
This is the only record of the run you get: no results table, no code.
(*\textit{the law after the first $k$ iterations}*)

## Predict these iterations
The iterations below were run one after another, in this order. Predict the score of each one.
### Iteration (*$k{+}1$*)
The one-line description the agent wrote for it: (*\textit{strategy of iteration $k{+}1$}*)
(*\textit{\dots{} one block per remaining iteration \dots}*)

Answer with ONE JSON object, nothing else:
{"reasoning": "<2-4 sentences>", "bet": {"iter_(*$k{+}1$*)": <number>, ..., "iter_15": <number>}}
\end{tcblisting}

\section{Discovered Curation Strategies}
\label{app:discovered}

Within our experiments, the agent has discovered multiple strong policies by itself that significantly improved baseline performance. In this section, we illustrate some of the strongest policies discovered by the agent as examples.

\subsection{Biology}

The first case is a \alg run of Claude Opus~4.8 on the CYP2C9 deep mutational scanning (DMS) assay (\texttt{cp2c9\_human}). The agent selects 1{,}868 of the 3{,}736 single amino-acid substitution variants in the training pool. The downstream model is a ridge regressor with fixed hyperparameters on frozen ESM-2 embeddings, trained on the selected variants and evaluated on the validation split.

\begin{tcolorbox}[protobox, title={Position-stratified selection spanning substitution severity (CYP2C9)}]
\small
\textbf{Starting rule.} A DMS assay measures the functional effect of every single amino-acid substitution of a protein~\citep{fowler2014dms}, so training under a fixed budget requires choosing which substitutions to keep. The card's rule is stratified sampling with residue positions as strata: every position is represented instead of a few positions being sampled exhaustively, and within each position the selected substitutions are spaced along an amino-acid distance scale, from conservative (physicochemically similar) to radical (dissimilar) replacements. The training set therefore shows how strongly the effect at each position depends on how chemically different the replacement is (substitution severity). The budget is split across positions in proportion to their number of measured variants (proportional allocation). The card does not specify which distance to use.

\textbf{What the agent found.} The agent identified two issues that the card's description had left open. First, the choice of distance matters: spacing substitutions along BLOSUM62 substitution scores~\citep{henikoff1992blosum} or Miyata distance~\citep{miyata1979two} performed at the level of a random 50\% subset, whereas Grantham distance~\citep{grantham1974amino}, which combines composition, polarity and volume, improved on it. Second, proportional allocation left 40 of the 287 positions with a single selected substitution; a position with one substitution provides no within-position contrast, so the rationale of the rule does not apply there. The agent also found that the spacing must be computed within each position's own severity range: using severity thresholds shared across all positions brought performance back to the random-subset level.

\textbf{Final policy.} The policy uses Grantham distance, spaces substitutions within each position's own severity range, and enforces a minimum allocation of two substitutions (one conservative, one radical) at every position with at least two measured variants. The budget for these reserved slots is taken from the positions with the largest allocations, and the remaining budget is allocated slightly super-proportionally to the number of measured variants per position. The number of positions with a single selected substitution fell from 40 to 8, and each of those 8 has only one variant in the pool.

\textbf{Results.} Spearman's $\rho$ on the validation split, with the regressor trained on the 1{,}868 selected variants:
\begin{center}
\renewcommand{\arraystretch}{1.1}
\begin{tabular}{@{}lc@{}}
\toprule
Full training pool (no curation) & 0.430 \\
Random 50\% subset & 0.391 \\
Severity spacing, BLOSUM62 & 0.391 \\
Severity spacing, Grantham & 0.448 \\
\quad + super-proportional allocation & 0.450 \\
Agent's final policy & \textbf{0.462} \\
\bottomrule
\end{tabular}
\end{center}
Each setting was also evaluated at values on both sides of the one the agent chose. A minimum of two substitutions per position was best: reserving three took substitutions away from the most densely measured positions and lowered $\rho$ to 0.446.
\end{tcolorbox}


The second case is a run of the same model on the PTEN DMS assay (\texttt{pten\_human}), with an MLP regressor on the same frozen ESM-2 embeddings; the agent selects 2{,}194 of 4{,}387 variants. 

\begin{tcolorbox}[protobox, title={Noise filtering before coverage-based selection (PTEN)}]
\small
\textbf{Starting rule.} Coverage-based selection chooses a subset that spans the model's input space, a criterion related to core-set selection~\citep{sener2018coreset}. The pool is partitioned into regions of the ESM-2 embedding space and the budget is filled round-robin across regions, so the training set covers the space the model must generalize over instead of concentrating where the assay sampled most densely.

\textbf{What the agent found.} On its own, this rule performed below random selection. Round-robin filling over-samples sparsely populated regions of the embedding space, and in this assay those regions contained the least reliable measurements rather than the most informative substitutions. Coverage selection cannot distinguish a variant that is atypical because it is chemically distinctive from one that is atypical because its measurement is noisy, so a rule meant to broaden the training set spent much of the budget on noisy labels.

\textbf{Final policy.} The agent filtered noisy variants before applying coverage selection, in the spirit of cross-validated filtering of mislabeled training data~\citep{brodley1999identifying}. Each variant was scored by its prediction error under a model fitted on variants at other residue positions (an out-of-fold residual); within each bin of assay score, the 25\% of variants with the largest errors were removed; and the same coverage selection was then applied to the remaining variants. Neither step was modified; only their order was new. Coverage selection improved performance only when applied after filtering.

\textbf{Results.} Spearman's $\rho$ on the validation split:
\begin{center}
\renewcommand{\arraystretch}{1.1}
\begin{tabular}{@{}lc@{}}
\toprule
Full training pool (no curation) & 0.359 \\
Random 50\% subset & 0.338 \\
Coverage selection only & 0.319 \\
Residual-based filtering only & 0.390 \\
Agent's final policy (filter, then cover) & \textbf{0.403} \\
\bottomrule
\end{tabular}
\end{center}
The filtering fraction mattered sharply rather than gradually: removing the worst 15\% of each bin instead of 25\% lowered $\rho$ to 0.345, consistent with coverage selection concentrating on the noisy variants that remained.
\end{tcolorbox}


The third case is single-cell cell-type annotation across sequencing protocols (\texttt{pbmc\_crossprotocol}). A protocol is the experimental method used to capture cells and prepare sequencing libraries. In a \alg run of Claude Sonnet~5, the agent selects 4{,}082 of 8{,}165 cells to train a linear classifier with fixed hyperparameters. The training pool contains 10x Chromium v2 and v3 cells, and the classifier is evaluated on cells from a protocol absent from the pool (inDrops).

\begin{tcolorbox}[protobox, title={Cross-protocol margin scoring within stratified quotas (PBMC)}]
\small
\textbf{Starting rule.} Margin-based selection ranks each cell by its classification margin, the classifier's score for the cell's annotated label minus its highest score among the other labels, computed by a classifier fitted on the training pool, and fills the budget from one end of the ranking. Keeping high-margin cells gives lower-variance estimates of where each cell type lies in expression space, at the cost of the ambiguous cells near the boundaries between types.

\textbf{What the agent found.} When the margin comes from a classifier fitted on the same pool it ranks, it reflects how well a cell fits its own sequencing protocol. A cell can receive a high margin from protocol-specific technical signal rather than from its cell type (datasets from different technologies carry batch-specific systematic variation~\citep{tran2020benchmark}), and that signal does not transfer to an unseen protocol, so the ranking favors the cells least likely to remain informative at evaluation. In a comparison within the training pool, margins from classifiers fitted on a different protocol gave a macro-F1 of 0.929, against 0.918 for margins computed within the same protocol.

\textbf{Final policy.} Each cell was scored by classifiers fitted without its protocol (leave-one-protocol-out scoring), and each protocol~$\times$~cell-type quota was filled with its highest-scoring cells. The final score averages two such scores, each standardized (z-scored) within protocol: one derived from the calibrated class posterior and one from distances to the class centroids, both from the same held-out-protocol fits. The number of cells per protocol and per cell type was left unchanged, so the policy changed only which cells fill each quota.

\textbf{Results.} Macro-F1 on the validation split:
\begin{center}
\renewcommand{\arraystretch}{1.1}
\begin{tabular}{@{}lc@{}}
\toprule
Full training pool (no curation) & 0.837 \\
Random 50\% subset & 0.849 \\
Lowest-margin cells & 0.826 \\
Cross-protocol margin, posterior only & 0.889 \\
Agent's final policy & \textbf{0.911} \\
\bottomrule
\end{tabular}
\end{center}
An independent run on the same task with a neural-network classifier adopted the same criterion and improved macro-F1 from 0.784 to 0.900, against 0.794 with the full training pool.
\end{tcolorbox}

\subsection{LLM}
\label{app:casellm}

The strongest policy produced in the language-model track came from a \alg run of Claude Sonnet~5 on the LESS mixture, in which the agent selects 10{,}000 of 270{,}679 rows to fine-tune Llama-2-7B and is scored on the mean of MMLU and BBH. The agent implemented RDS+ \citep{ivison2025large} and mapped the method's degrees of freedom over several iterations, and then identified a defect in the method itself and wrote the variant below, which is the run's best policy.

\begin{tcolorbox}[protobox, title={A variant of RDS+ written by the agent}]
\small
\textbf{The published method.} RDS+ selects by cosine similarity in a cached embedding space. Every pool row, and every query point drawn from the target evaluations and written in the format those evaluations use, is encoded as a position-weighted mean of the last hidden layer of the model about to be fine-tuned. The budget is then filled by round-robin over the query points, each taking its nearest unselected row. The cost is one forward pass over the pool, reusable across budgets.

\textbf{The problem the agent found.} The similarity that ranks candidates is a similarity of surface form rather than of what a row teaches. Query points formatted as the benchmarks format them, and in particular the multi-step reasoning traces used for BBH, are closest to the pool's chain-of-thought rows, which consequently dominate nearly every ranking and absorb most of the budget. The two human-written, open-ended sources in the pool are almost entirely excluded: no query point resembles an open-ended exchange, so similarity cannot surface those rows however much the model needs them to follow an instruction and lay out an answer. The agent's diagnosis was that one ranking is being asked to settle two decisions of different kinds, the allocation of the budget across sources and the choice of rows within a source, and that it is informative only about the second.

\textbf{The agent's variant.} The variant separates the two decisions. A fixed share of the budget is reserved for each under-selected source and filled by the same round-robin restricted to that source's candidates, before the unrestricted round proceeds over the remainder. The rows admitted this way are still the nearest available rows of their source to a query point rather than random draws from it, and the embedding, the query set and the ranking are untouched. Similarity continues to order rows within a source; it no longer determines the allocation across sources.

\textbf{Result.} Mean accuracy over MMLU and BBH after fine-tuning Llama-2-7B on the 10{,}000 selected rows:
\begin{center}
\renewcommand{\arraystretch}{1.1}
\begin{tabular}{@{}lc@{}}
\toprule
Base model, no fine-tuning & 42.59 \\
Random selection & 42.64 \\
RDS+ & 45.41 \\
The agent's variant & \textbf{45.78} \\
\bottomrule
\end{tabular}
\end{center}
\end{tcolorbox}

\end{appendices}

\end{document}